\documentclass[10pt]{article}
\usepackage[preprint]{tmlr}
\usepackage{hyperref}
\usepackage{url}
\usepackage{booktabs}
\usepackage{amsfonts}
\usepackage{nicefrac}
\usepackage{microtype}
\usepackage{xcolor}
\usepackage{graphicx}
\usepackage{caption}
\usepackage{subcaption}
\usepackage{amsmath}
\usepackage{mathtools}
\usepackage{upgreek}
\usepackage{relsize}
\usepackage{rotating}

\usepackage{amsmath,amsfonts,bm}

\def\eqref#1{equation~\ref{#1}}

\def\1{\bm{1}}

\DeclareMathAlphabet{\mathsfit}{\encodingdefault}{\sfdefault}{m}{sl}
\SetMathAlphabet{\mathsfit}{bold}{\encodingdefault}{\sfdefault}{bx}{n}

\title{Uncertainty-Aware Deep Learning for Genomics Applications: Insights from an Empirical Study}

\author{\name Sepideh Saran \email sepideh.saran@mdc-berlin.de \\
      \addr The Berlin Institute for Medical Systems Biology,\\
      Max Delbr\"uck Center for Molecular Medicine, Germany\\
      Department of Electrical Engineering and Computer Science,\\
      Technical University of Berlin, Germany
      \AND
      \name Mahsa Ghanbari \email mahsa.ghanbari@mdc-berlin.de \\
      \addr The Berlin Institute for Medical Systems Biology,\\
      Max Delbr\"uck Center for Molecular Medicine, Germany\\
      Department of Biology, Humboldt University of Berlin, Germany
      \AND
      \name Uwe Ohler \email uwe.ohler@mdc-berlin.de \\
      \addr The Berlin Institute for Medical Systems Biology,\\
      Max Delbr\"uck Center for Molecular Medicine, Germany\\
      Department of Biology, Humboldt University of Berlin, Germany\\
      Department of Computer Science, Humboldt University of Berlin, Germany}

\newcommand{\chg}[1]{#1}
\newcommand{\rev}[1]{#1}
\def\month{MM}  
\def\year{YYYY} 
\def\openreview{\url{https://openreview.net/forum?id=XXXX}} 
\newenvironment{ack}{\section*{Acknowledgments}}{}

\begin{document}

\maketitle

\begin{abstract}
Deep learning models have \chg{emerged as the standard computational tool} for a wide range of applications in genomics. Yet, uncertainty quantification  (UQ) --- and more specifically, the reliability of different uncertainty estimates in this domain --- has received little systematic attention. This work presents an empirical analysis of UQ in deep learning models, focusing on genomics applications. In a series of experiments, we contrast Deep Ensembles, Bayesian Neural Networks, and Monte Carlo-dropout methods. We assess their ability to quantify uncertainty in different scenarios, accounting for common dataset characteristics in two genomic application areas and modalities: sequence-to-activity models, and single-cell expression analysis. Our systematic comparison framework provides guidelines for the applicability and reliability of UQ methods in genomics, highlighting their strengths and limitations in different scenarios. \chg{\rev{Our results indicate that Bayesian Neural Networks give the most consistent and directionally specific uncertainty response to class imbalance and out-of-distribution data in the tested settings, and mark the corrupted classes under label noise, at a higher computational and tuning cost. MC-dropout matches them on the sequence data, but is uncertain about every cell on the single-cell data, and deep ensembles respond less specifically.} Moreover, we show how uncertainty scores can be used to select high-quality predictions in protein--RNA interactions.}
\end{abstract}

\section{Introduction}\label{sec1:intro}
 Neural networks have become the state-of-the-art methods for identifying functional elements in the genome \citep{theis_review}. Superior performance, scalability, and the rise of explainable artificial intelligence (XAI) \citep{xai_1,xai_2, xai_3,sundararajan2017axiomatic,koo2021, majdandzic2023} have made deep learning models a popular choice in many genomics applications. 

For the ultimate use of these models in downstream tasks (e.g. predicting genetic variant effects as part of clinical diagnostics \citep{remo}), it is essential to provide a measure of confidence in the model's outputs. Providing uncertainty measurements improve the credibility of a machine learning solution and support downstream decision-making. Moreover, high costs necessitate the transfer of pre-trained models for inference on other datasets, but frequently we lack sufficient information about the characteristics of the original training data (e.g. whether the model was trained on a highly imbalanced/biased dataset). UQ methods can shed light on such shortcomings of the training data.
 
Evaluating UQ in genomics is itself challenging, because of the nature of the data and the diversity of uncertainty sources. Experimental noise, incorrect labels, out-of-distribution samples, and class imbalance are among the major sources of uncertainty for computational models in biology, and they often co-occur with task complexity such as multi-label classification. Critically, reliable ground truth is frequently unavailable --- either because labels are themselves derived from noisy experiments, or because the model is applied to a novel domain that lacks expert annotation. This makes assessing the quality of uncertainty scores \emph{independently of predictive accuracy} essential in genomics.

So far, limited work has leveraged UQ methods in genomic applications and provided guidelines for choosing among them. In computational biology, UQ has previously been approached through calibration \citep{gruber2022proper, maierhein2024metrics}, which aligns confidence scores with the true likelihood of predictions, and through conformal prediction \citep{angelopoulos2021gentle}, which provides statistically rigorous coverage guarantees. Both, however, rely on labeled data, making them difficult to apply when labels are noisy or unavailable --- both common in genomics. Closest to our work, \citet{bajwa2024characterizing} quantify uncertainty in genomic sequence-to-activity models through the consistency of deep ensemble replicates, and \citet{degu} introduce DEGU, which distills an ENS into a single model capturing both epistemic and aleatoric uncertainty across functional genomics prediction tasks, with conformal prediction providing coverage guarantees.

Systematic benchmarks of UQ methods exist for computer vision and other data-rich domains~\citep{ovadia2019can, filos2019systematic, arctique}\rev{, and more recently for molecular property prediction with pre-trained representation models~\citep{muben2024}}, but genomics presents fundamentally different challenges that limit the transferability of these findings. Moreover, existing benchmarks typically assess uncertainty through its relation to predictive accuracy in supervised settings, not suitable for the label-scarce settings common in genomics. To the best of our knowledge, no prior work has systematically compared multiple UQ methods across the dataset conditions that characterize genomics, nor provided guidelines for which method to choose in a given setting.

To address this gap, we present a framework for evaluating UQ methods in deep learning. We examine UQ through deep ensemble (ENS), Bayesian neural networks (BNN), and Monte Carlo-dropout (MC-dropout), in two well-studied classification tasks in genomics: protein binding prediction from DNA/RNA sequence and predicting cell type from single-cell gene expression. Using both simulated and real data, we systematically compared whether these UQ methods capture uncertainty from various sources commonly encountered in genomics. This controlled comparison highlights the strengths and weaknesses of these UQ methods in different scenarios, providing guidelines for their applicability and reliability. To assess UQ in a controlled manner, we begin with simulated sequence data, where the source of uncertainty can be more reliably controlled. We repeat the experiments with real sequence data and demonstrate the practical utility of uncertainty scores to obtain high-quality predictions. Finally, we look at how the findings play out on a different data modality.

The remainder of this paper is organized as follows: In section~\ref{sec_2:background}, we introduce the background on the UQ methods and the genomic tasks in the scope of this work. Section~\ref{sec_3:framework} describes the framework for this analysis and the modeling details. Section~\ref{sec_4:results}, presents highlights of the findings of these empirical evaluations. Finally, section~\ref{sec_5:conclusion} concludes this paper and presents directions for future work.

\section{Background}\label{sec_2:background}
\subsection{Methods for quantifying uncertainty in neural networks}\label{sec_2_1:uqmethods}
To study the predictive uncertainty estimation methods in deep learning, we compare three of the most common UQ methods for neural networks: BNN trained with variational inference, deep ensemble, and MC-dropout.

Deep ensembles, inspired by tree-based ensemble models, are a simple way of acquiring uncertainty scores in neural networks (NNs). A deep ensemble is a group of NNs, each trained separately with a different random initialization. The prediction of the ensemble is given by averaging all models' predictions and the uncertainty score, by taking the standard deviation (SD) or variance of them. Different training strategies have been employed to either train each model in the ensemble on the whole training set or only a subset of it, or to include adversarial training \citep{NIPS2017_9ef2ed4b}. 

Bayesian neural networks are the more theoretically sophisticated alternatives for UQ. BNNs introduce stochasticity to the network by learning a distribution for each weight (and bias) instead of a single point estimate \citep{blundell2015weight}. A prior distribution is selected for each parameter and is represented by its mean and standard deviation, increasing the number of trainable parameters in the Bayesian neural network to twice the size of the equivalent deterministic architecture. Since exact inference is computationally intractable for these networks, Variational Inference \citep{jordan99, Wainwright2008, vi_alex_graves, Blei_2017} is one commonly used way to approximate the posterior on weights and thus train the network \citep{filos2019systematic}. A predictive distribution for the output can also be generated by drawing values for weights from the approximated posterior distribution (i.e., Monte Carlo sampling). Other strategies, such as Laplace approximation and MCMC-based learning, have been used for training BNNs as well \citep{review1}.

Monte Carlo-dropout has been introduced as a probabilistic approach that approximates Bayesian inference in deep Gaussian processes, considering certain assumptions \citep{gal2016dropout}. MC-dropout extends the now-well-established practice of using Dropouts \citep{JMLR:v15:srivastava14a} during the training phase to the test phase, bringing stochasticity to the network at inference time. MC-dropout has the same number of parameters as the deterministic equivalent network and uses the same training procedure, making it computationally more efficient than Bayesian neural networks. MC-dropout has revealed promising results on image datasets \citep{filos2019systematic}, but to the best of our knowledge, it has not been yet tested on other data types such as genomic sequence data.

These three methods represent different trade-offs between theoretical grounding, computational cost, and implementation simplicity. Next, we describe the two genomic use cases in which we compare these methods empirically.

\subsection{Use case I: predicting protein binding from DNA/RNA sequence}\label{sec_2_3:usecase}
Proteins regulate different stages of gene expression through binding to specific locations in the genome (DNA) or the transcriptome (RNA). Proteins that bind to the DNA, Transcription Factors (TFs), can activate or repress transcription, and RNA binding proteins (RBPs) can affect post-transcriptional processes such as splicing, RNA localization, translation, and degradation \citep{Gerstberger1}. 
These proteins are selective of their target binding sites, as they recognize short, local patterns in the sequence (i.e. sequence motifs) and/or the DNA/RNA structure (i.e. structural motifs). Evolutionarily related proteins often share preferred sequence patterns for binding, while those with different protein binding domains have distinct motifs. Identifying proteins' binding preferences opens the door to understanding their impact on downstream biological processes. Following previous work \citet{DeepRipe}, we formulate the problem of protein binding prediction as a multi-label classification task (i.e., to predict which protein(s) bind to a given section of DNA/RNA) and extend our experiments to compare the uncertainty estimation approaches for this problem. 

\subsubsection{Simulated transcription factor dataset}\label{sec_2_3:simdata}
To account for different dataset characteristics of interest in a controlled setting, we generated simulated datasets for predicting protein-binding events for DNA sequences. The input data is a $150$-nucleotide long section of DNA, and \chg{the} classification task is to predict which proteins out of a selection of transcription factors bind to it.  We chose $6$ TFs with widely known sequence motifs, namely, CTCF, HNF4A, JUN, MEF2A, MYC, and TAL1 as our target classes for this task. With this simulation strategy, we can generate scenarios for comparing UQ method performance on clean versus noisy datasets. In each scenario, the noisy dataset has only a single source of uncertainty introduced compared to the clean version. The simulated input sequences are generated so that \chg{the} noisy dataset \chg{either has} missing sequence motifs\chg{,} or \chg{is} partially mislabeled. Such cases are common when working with noisy biological data. 

To this end, we used SimDNA\citep{simdna} to synthesize DNA sequences and plant sequence motifs in them. SimDNA is a tool developed in Python for generating genomic regulatory sequences. It first generates a background sequence and then embeds desired sequence motifs in the background. For our experiments, we used a first-order Markov model to generate DNA background sequences of length $150$ and embedded a single motif sequence for a TF class. We used the Jaspar open-source database\citep{jaspar2020} to obtain sequence motifs of our selected TFs. JASPAR provides manually curated TF binding profiles as position frequency matrices (PFMs). These PFMs are then used in SimDNA to sample and embed a sequence motif of a TF into the background DNA sequence (see suppl. Figure 1). We generated three simulated datasets: a clean single-label balanced dataset as a noise-free reference, a noisy dataset where the sequences have no motifs, and a single-label balanced dataset where a portion of data is mislabeled. 

The simulated dataset consists of $2000$ samples per class\rev{, split into training, validation, and test sets as the real data (Section~\ref{app_a:data:real})} (the entire dataset size is $12000$). For the dataset with no motifs, we simply used the background sequences without implanting any motifs and for the mislabeled dataset, we swapped the labels between the two TF classes JUN and MEF2A for \rev{half of the samples of each ($16.6\%$ of all training samples)}.

\subsubsection{Real RNA binding protein dataset}\label{sec_2_3:realdata}
\begin{figure}
     \centering
     \begin{subfigure}[b]{0.48\textwidth}
         \centering
         \includegraphics[width=\textwidth]{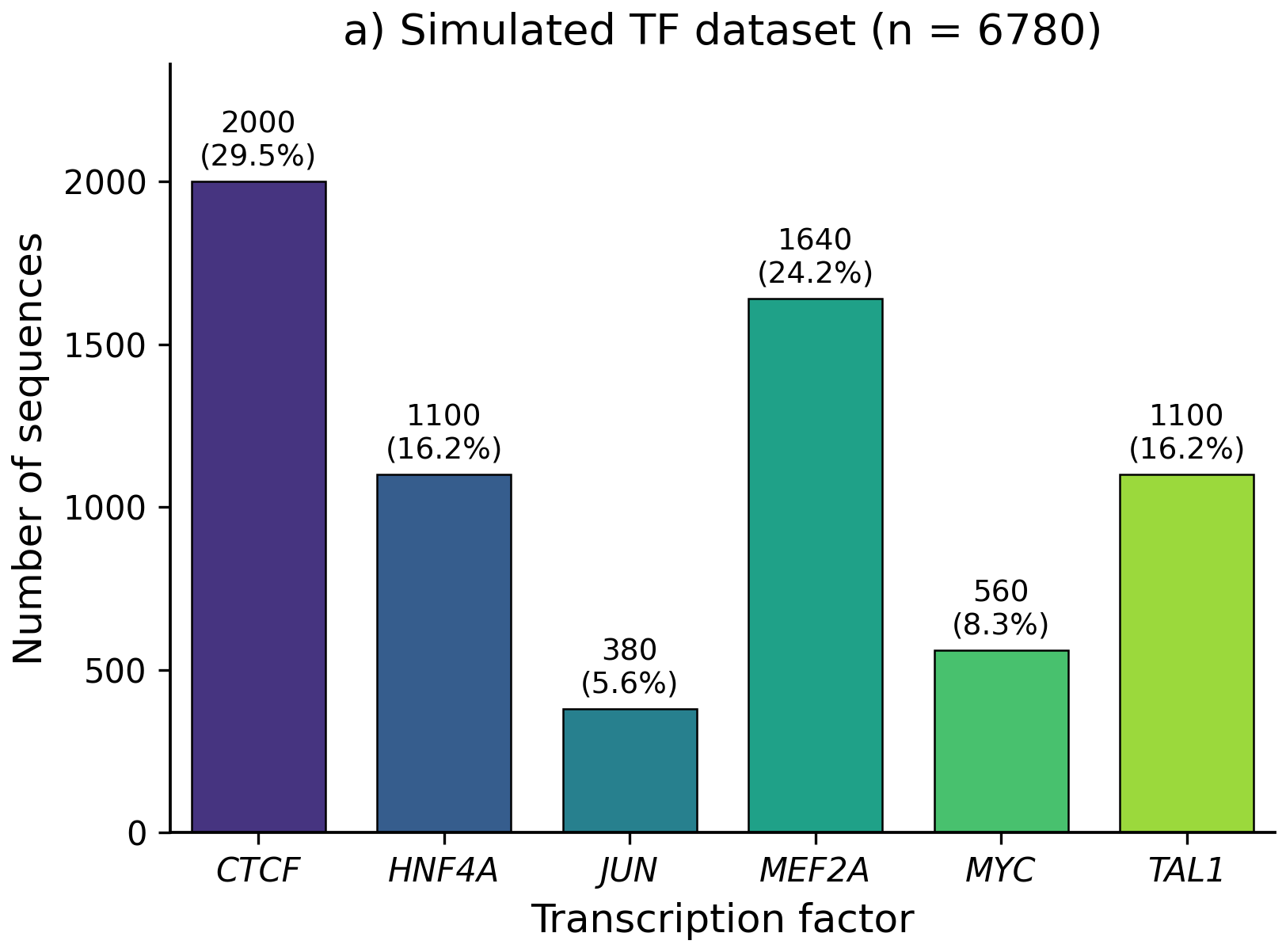}
     \end{subfigure}
     \hfill
     \begin{subfigure}[b]{0.48\textwidth}
         \centering
         \includegraphics[width=\textwidth]{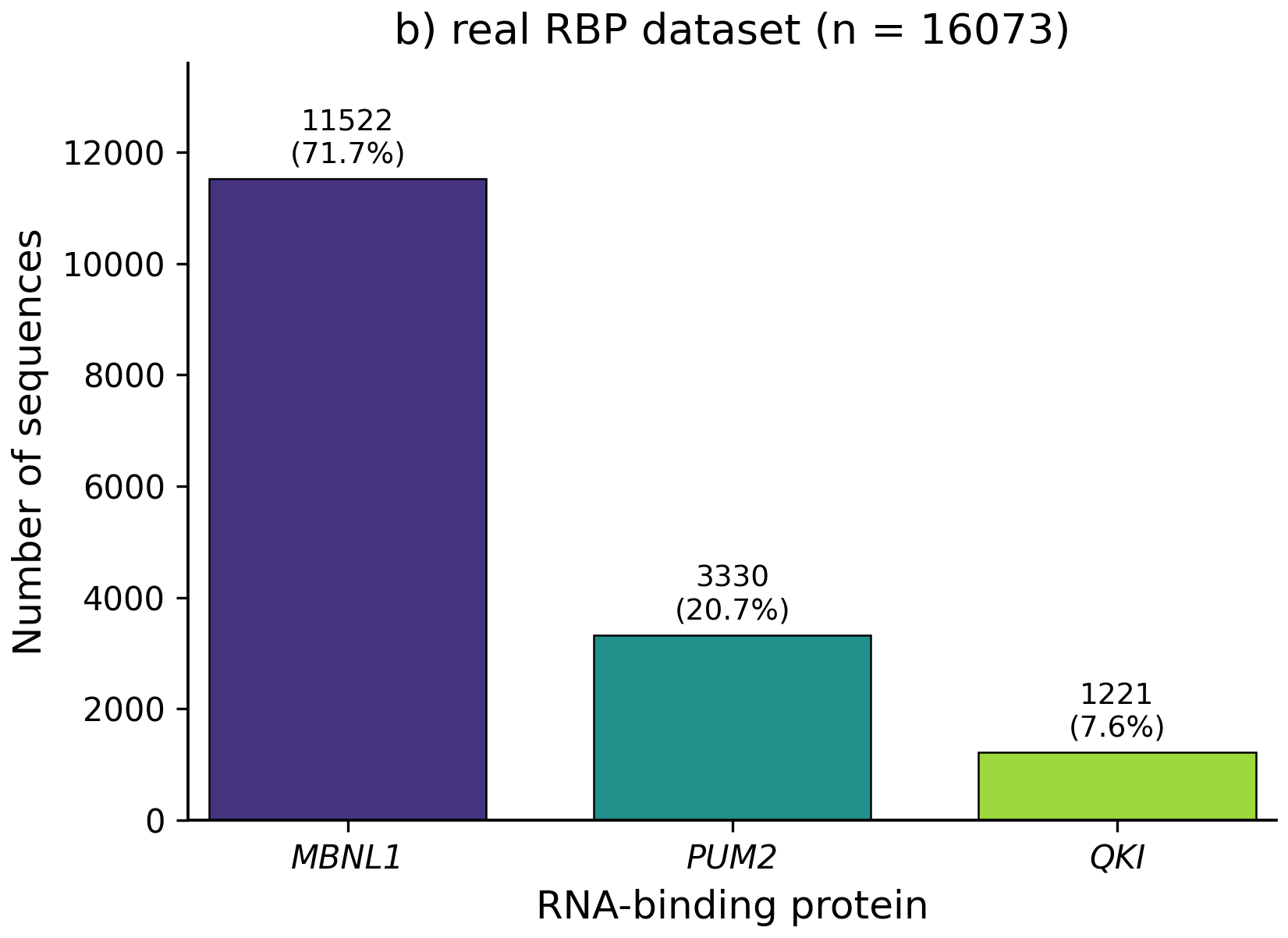}
     \end{subfigure}
     \hfill
        \caption{\textbf{Simulated TF (a) and Real RBP (b) datasets:} class distribution \chg{ of the imbalanced version of the simulated TF dataset (a), and of the real RBP dataset (b)}\rev{. The imbalanced simulated TF dataset ($6{,}780$ sequences) is shown for comparison with the real data; the TF experiments in this paper use the balanced simulated dataset ($2{,}000$ sequences per class, Section~\ref{sec_2_3:simdata}). The imbalanced RBP dataset is used throughout.}}
    \label{fig:class_dist}
\end{figure}

In further experiments, we assess the UQ methods on previously published real protein binding datasets. PAR-CLIP experiments provide high-resolution binding site sequences for the RBP of interest \citep{parclip}. These bindings sites are small RNA fragments of typically a few dozen nucleotides, which contain sequence patterns of typically 4-8 nucleotides in length. We select PAR-CLIP data \citep{neel2019} for three RBPs, namely MBNL1, PUM2, and QKI, for this task. Since PAR-CLIP datasets are generally noisy, we have selected this subset of RBPs as their preferred sequence motifs are known, making it a suitable choice for benchmarking with real data in a more controlled setting. Our dataset is a combination of several published PAR-CLIP datasets obtained in the HEK293 cell-line, which were processed uniformly through a single pipeline \citep{paralyzer}. The dataset contains approximately $16,500$ samples in total. It is imbalanced and includes approximately $475$ multi-label samples, meaning one sequence can have the binding motifs for multiple RBPs. For further analysis, we sub-sample from the larger classes to build a balanced multi-label dataset containing about $4,500$ (including $460$ multi-label) samples (see Figure~\ref{fig:class_dist}) and a balanced single-label version of the same dataset. \chg{Another source of uncertainty we experiment with in this use case is out-of-distribution data. Applications on unseen genomes are exactly the setting where a UQ method matters in practice: experimental labels are expensive or unavailable, as for a newly sequenced virus. As the underlying biology is not yet fully understood, one cannot know in advance how much a model trained on human data will transfer, even between datasets from the same protocol and tissue. We tested this directly by applying the RBP models, without retraining, to eight coronavirus genomes where no RBP binding label exists (Section~\ref{app_b:ood}).}

\subsubsection{Protein binding prediction}\label{sec_2_3:model}
The base model for this use case is a simple deterministic convolutional neural network (CNN) adopted from DeepRipe \citep{DeepRipe}, which has state-of-the-art performance for this classification task. The base model architecture has two convolutional layers, each followed by a max-pooling layer and a final fully-connected layer.  For single-label setting, we use the softmax activation function in the last layer, whereas, in multi-label, sigmoid. 

\subsection{Use case II: predicting cell type from single-cell gene expression}\label{sec_2_2:usecase}

Single-cell RNA-seq (scRNA-seq) experiments measure the degree to which each gene is expressed (i.e. activated) for each of thousands of single cells. A typical scRNA-seq experiment measures the expression of around $15,000$ genes for tens of thousands of cells. These data can be used as input for computational tasks such as cell type prediction, understanding disease states, developmental trajectories, and more. 

Out of technical reasons, data generated by such experiments provides a noisy, large, sparse count matrix (number of cells by  number of genes) with many missing values. To analyze this data and derive meaningful biology, it is necessary to present the data in a lower-dimensional space. Dimensionality reduction approaches for scRNA-seq data include classical methods such as variations of PCA, UMAP~\citep{mcinnes2018umap}, and matrix factorization methods. More recently, the field has adapted variational auto-encoders (VAEs)~\citep{vae_original} as a powerful alternative, capable of learning biologically meaningful low-dimensional representations in an unsupervised manner~\citep{luecken2019current}. The expectation is that successful embeddings represent the cell types in latent space while preserving biologically meaningful variation\citep{liam}.

\subsubsection{Single-cell dataset}\label{sec_2_2:data}
The data for this use case is selected from a dataset of multi-modal single-cell experiments~\citep{luecken2021a}, available from the Gene Expression Omnibus under accession GSE194122. This dataset was originally published for the multimodal single-cell data integration competition in NeurIPS 2021~\citep{pmlr-v176-lance22a}. It contains several single-cell data modalities from a total of $120,000$ bone marrow cells measured on $4$ different sites (i.e. laboratories) from $10$ human donors. This dataset also provides the annotations for the cell types that are the labels in our use case. To analyze the uncertainty scores in a controlled setting without unwanted source of uncertainty, we select a subset of this dataset to simplify the classification task. Specifically, we selected a subset of four cell types from the single-cell RNA-seq data, but included data obtained from across multiple sites and donors to keep the biological variation. \rev{The train, validation and test splits are drawn at random over cells, so a donor can appear on both sides of a split; Section~\ref{app_b:donor} reports accuracy per donor and per site and shows that it does not vary with either.} We generate a balanced (\rev{$18,000$ cells}) and an imbalanced version (\rev{about $25,000$ cells}) of this dataset (see Figure~\ref{supp_fig:gex_class_imba}) to compare the performance of UQ methods when imbalance is introduced to a dataset as a source of uncertainty. \rev{In every perturbed setting of this paper the test set is the one of the reference setting and is held out before the perturbed training and validation sets are drawn, so that a model is never evaluated on a sample it was trained on (Section~\ref{app_a:data}).} 

\subsubsection{Cell type classification}\label{sec_2_2:model}
\begin{figure}
     \centering
     \includegraphics[width=\textwidth]{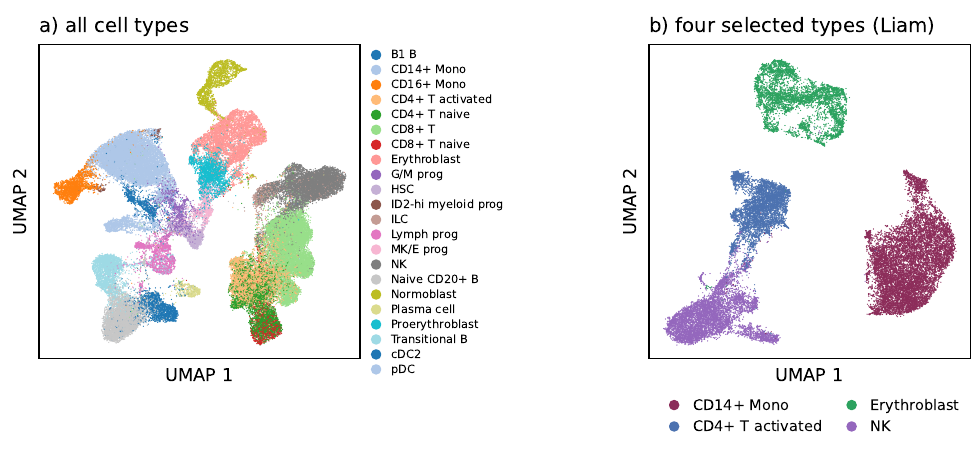}
        \caption{\textbf{Single-cell gene-expression dataset: Low-dimensional embeddings.} \rev{a) UMAP of the full scRNA-seq dataset ($69{,}249$ cells \rev{of the multiome experiment, whose gene-expression modality we use}, $22$ annotated cell types). b) UMAP of the $20$-dimensional Liam embedding~\citep{liam} of the $28{,}214$ cells of the four selected cell types that make up the use case.}}
        \label{supp_fig:liam_embeddings}
\end{figure}

\begin{figure}
     \centering
    \includegraphics[width=0.5\textwidth]{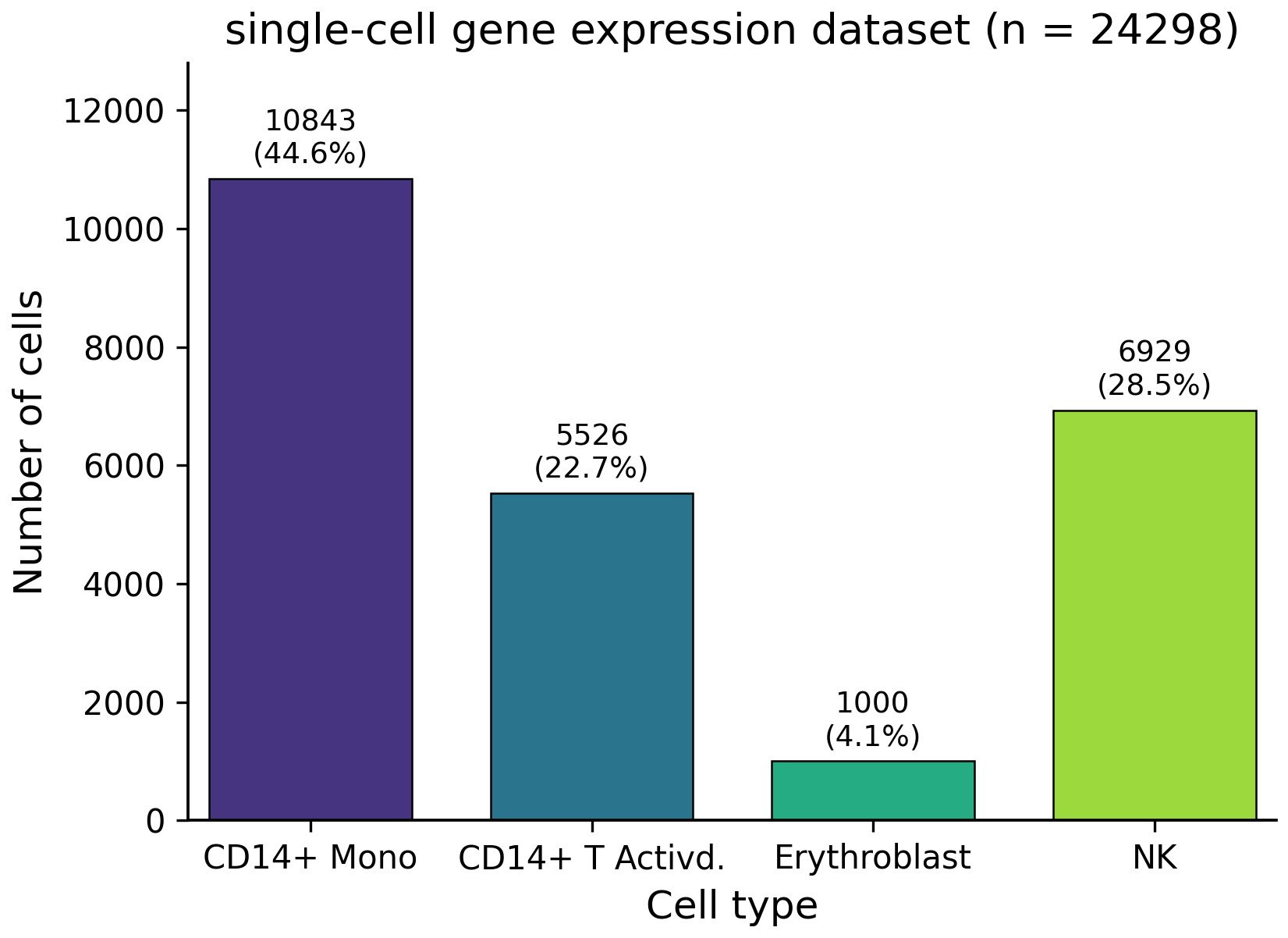}
    \caption{\textbf{Single-cell gene-expression dataset:} Class distribution for the four selected cell types in the imbalanced setting.}
    \label{supp_fig:gex_class_imba}
\end{figure}

As mentioned above, cell type prediction typically starts from dimensionality-reduced data rather than the raw high-dimensional sparse data. \chg{W}e used a published VAE model Liam\citep{liam} for generating $20$-dimensional embeddings from the high-dimensional original gene expression matrix. This model showed state-of-the-art performance in the competition that provided the data for our task\citep{pmlr-v176-lance22a} (Figure~\ref{supp_fig:liam_embeddings}). We then use the embeddings as input to a two-layer fully-connected neural network model to predict cell types. For the UQ comparison, we generate the embeddings once, and we introduce stochasticity only to the \rev{fully-connected classifier}. Therefore, the UQ comparisons here refer to the fully-connected neural network architecture, with $10$ nodes in each dense layer and a drop-out rate of $0.3$.

\section{The empirical analysis framework}\label{sec_3:framework}
The empirical framework outlined in this study intends to provide guidelines for choosing the most suitable UQ method in applications with specific dataset characteristics, such that the calculated uncertainty score can be trusted when using the model in a critical task. To this end, we conduct experiments in each use case under a collection of controlled dataset settings to cover scenarios with a particularly interesting source of uncertainty in biological applications, namely highly imbalanced data, scarce data, label noise, out-of-distribution, and multi-label classification. 

\subsection{Modeling and measuring uncertainty}
The base model for solving each task is a simple deterministic neural network as described in Sections~\ref{sec_2_3:model} and~\ref{sec_2_2:model}. We implemented probabilistic versions of these architectures to build uncertainty-aware models (i.e. BNN, MCD and ENS) for each task, respectively. We evaluate the prediction and uncertainty scores of the three selected approaches with respect to the deterministic variant for each scenario.

In training the deterministic baseline models, we use dropout for regularization. The BNN is built on a similar architecture, but with Gaussian prior and posterior on the trainable parameters and mean-field variational inference \citep{mfvi,blundell2015weight} as the training approach. To speed up the training process, we use the Flipout Monte Carlo estimator that de-correlates the gradients within each mini-batch \citep{wen2018flipout}. The MC-dropout networks have the same architecture as the deterministic models, with the difference of \rev{keeping the dropout layers active at inference time (Section~\ref{app_a:impl}; variants with a single active layer or a higher rate are in Section~\ref{app_b:mcd_zero})}. The ENS is a collection of $10$ deterministic models, each following the base model architecture. 

Training plans are adapted per model and dataset characteristic, as each setting requires specific considerations to converge reliably. Training is done to the point where all models reach a similar accuracy on the same setting, for uncertainty scores and predictions to be comparable. All models except for BNN are trained with a batch size of $128$ for $50$ epochs with early stopping based on the validation loss with the patience set to $5$. However, for the results of the networks to be comparable, we needed to train the BNNs longer to reach similar accuracy to MC-dropout's. For all experiments, we used the Adam optimizer~\citep{kingma2015adam}. 

\begin{figure}
    \centering
    \includegraphics[width=1.0\linewidth]{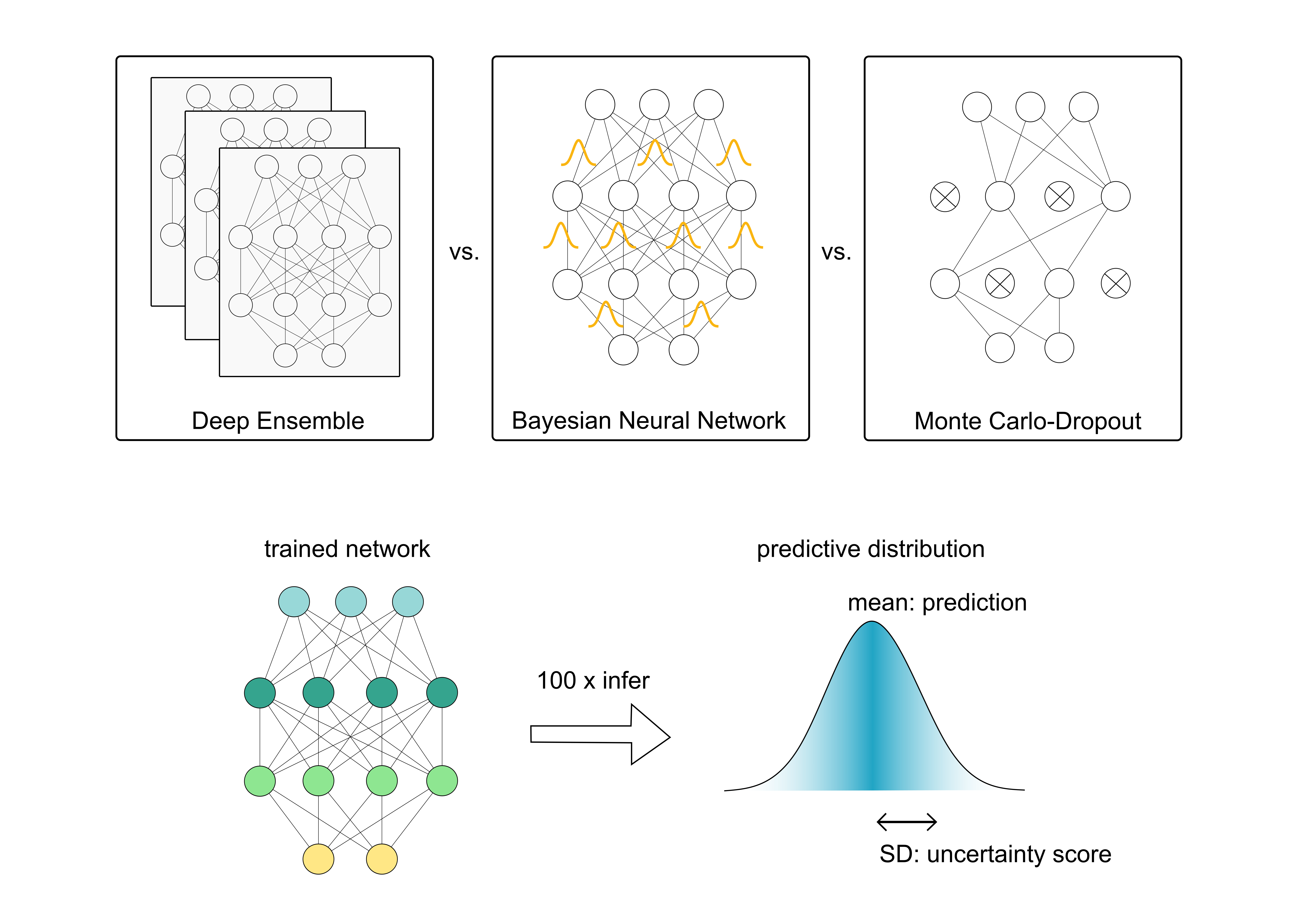}
    \caption{Overview of the UQ assessment framework. We compare three methods for uncertainty quantification in neural networks, namely, deep ensemble, Bayesian neural network, and the Monte Carlo-dropout. For each task and dataset setting, the three probabilistic networks are trained to achieve similar performance on the same training set. For each trained model, inference on the same test set is repeated $100$ times \rev{(BNN and MCD; ENS uses its $10$ members)} to generate a predictive distribution for each test sample. The mean of this distribution is the prediction and the standard deviation is the uncertainty score for the sample. }
    \label{fig:overview}
\end{figure}

After training the probabilistic networks, the inference on the test set is repeated for $100$ iterations \rev{for BNN and MCD}, building an output distribution for each test example through \rev{stochastic forward passes}\rev{; for ENS the distribution consists of the $10$ member predictions}. The mean of the distribution is regarded as the model prediction, and the standard deviation represents the model's uncertainty for the given test sample (see Figure \ref{fig:overview}). We fix the test set across all models for each task and compare the performance and uncertainties of the models.

The experiments are repeated with 5 different random seeds, and the models are implemented in python using Tensorflow \citep{tensorflow2015-whitepaper}. 

\subsection{Evaluating performance and uncertainty scores}

We consider the accuracy of the deterministic model as our baseline performance in each scenario. The probabilistic models are then trained to reach similar performance, to ensure comparability, and to avoid sacrificing accuracy when estimating uncertainty. In addition to the standard metrics for evaluating predictive accuracy on a test set, inspired by related work \citep{NIPS2017_9ef2ed4b}, we used Negative Log Likelihood (NLL) and the Brier score~\citep{brier1950verification} as two appropriate scoring rules to evaluate the predictive uncertainty. Furthermore, we calculate the Pearson correlation and Spearman correlation between the means of the predictive distributions to illustrate the agreements of the methods in their predictions.

The use cases are solved in supervised settings; therefore, we examine the distribution of uncertainty scores \chg{both for all test samples and, separately, for the samples classified correctly by both methods of a pair}. We assess a potential shift in the distribution of the uncertainty scores' distribution when we introduce a single source of uncertainty (e.g. mislabeled samples) to clean data. This controlled approach enables us to judge how well the UQ methods can reflect each source of uncertainty that could be present in the data. 

The range of absolute values for the uncertainty scores generated by the UQ methods can differ for the same data and task across the methods. Moreover, the uncertainty score can be calculated differently on the distribution of the predictions for the same UQ method. Therefore, we are not relying on the absolute values of the uncertainty scores for our evaluation, but we argue that looking at the shift of this range can provide model-and-use-case-specific insights. This exercise helps in choosing a threshold in the downstream use of uncertainty scores to filter for reliable predictions. 

We note that evaluating the uncertainty score beyond its relation to the prediction accuracy is limited to empirical comparisons on carefully constructed use-cases: there is no ground truth value for the uncertainty score, even for simple supervised tasks. To investigate the agreement of uncertainty estimates provided by the UQ methods and identify commonalities and differences, we compare the uncertainty scores across methods in three ways: Firstly, we calculate Kendall's tau \citep{kendall} between SDs of the pairs of models over 5 random seeds. Kendall rank coefficient correlation (Kendall's tau) measures the correspondence between ordinal data. We used this test to measure agreement of the ranking of uncertainty scores among UQ models. A strong agreement between ranked list of uncertainty scores in each scenario and random seed indicates that UQ methods point to the same samples as highly uncertain. 
\chg{Secondly, we perform the Kolmogorov–Smirnov (KS) test on the SD distributions of each method separately, comparing a reference (i.e. clean or balanced) dataset setting against a perturbed one, to measure how strongly each method's uncertainty responds to a given source of uncertainty. The null hypothesis of the test is that, for a given method and class, the predictive standard deviations are drawn from the same distribution in the reference and the perturbed setting. A larger KS statistic therefore means the method's uncertainty scores for that class moved further when the perturbation was introduced; a value near zero means they did not respond to it at all. Since the KS test can only show if the distribution has moved or not, we complement that with plotting the SD distribution shifts to check if the change in distribution is in the expected direction (i.e. larger uncertainty scores for models train on perturbed data).}
Lastly, we select the most uncertain samples (i.e highest standard deviation) as identified by each method on the same data and task. We then calculate the overlap of the top $100$ most uncertain samples across the experiments as a proxy to check if the UQ methods reflect similar approximations of the uncertainty and point to the same underlying source of uncertainty.

\section{Results}\label{sec_4:results}

To assess uncertainty quantification in a controlled manner, we first study sequence-based prediction of protein binding, where simulated transcription factor data lets us control the source of uncertainty exactly. In parallel, we examine how the same behavior plays out on real RNA-binding-protein (RBP) data. We then move to a different data modality, single-cell gene expression (GEX), evaluating the same aspects and referring back to the sequence-based results. Throughout, we report (i) predictive performance, (ii) agreement of uncertainty scores across UQ methods, and (iii) sensitivity of the uncertainty scores to a single, controlled source of uncertainty.\\

\subsection{Predictive performance}\label{sec_4_1:perf}
\begin{figure}[p]
    \centering
    \includegraphics[width=\textwidth]{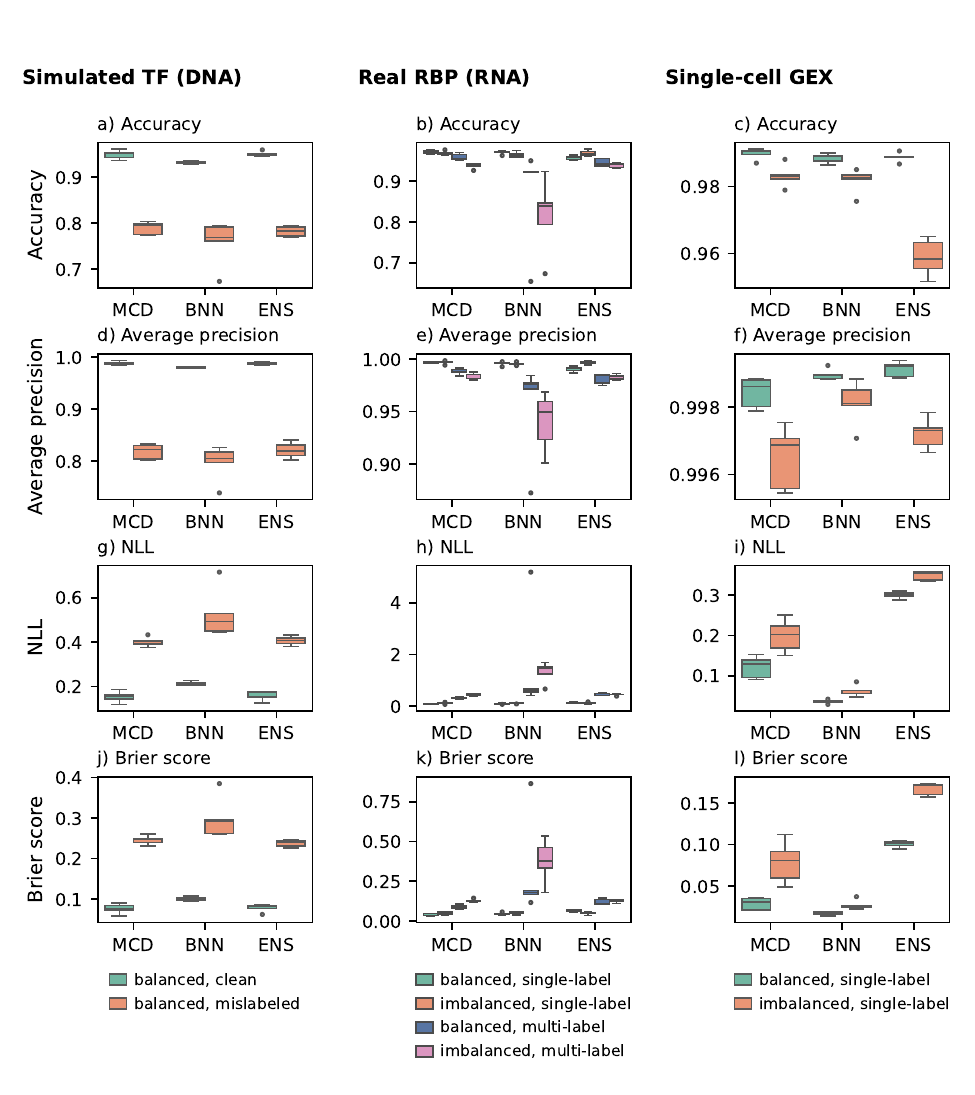}
    \caption{Predictive performance of the three UQ methods across all three use cases. Accuracy (a--c), average precision (d--f), negative log-likelihood (g--i), and Brier score (j--l) on the held-out test set for MCD, BNN and ENS. Columns group the use cases: simulated TF binding (a, d, g, j), real RBP binding (b, e, h, k), and single-cell cell-type prediction (c, f, i, l). Box plots summarize $5$ random-seed data splits and model initializations. \rev{Green is the reference setting of each use case (clean, or balanced single-label) and orange the perturbed one (mislabeled, or imbalanced single-label); for the RBP data, blue is the balanced and pink the imbalanced multi-label setting. The y-axis range differs between columns and is chosen per panel, so values should be compared within a column, not across.} \rev{For the multi-label RBP settings the NLL is the binary cross-entropy summed over classes (Section~\ref{app_a:metrics}); all five seeds are shown, including those in which the BNN reached lower accuracy.}}
    \label{fig:pred_acc}
    \label{fig:gex_pred_acc} 
\end{figure}

As intended, all models reach high and comparable predictive performance in each task, so that differences in their uncertainty scores can be attributed to the UQ method rather than to differences in accuracy (Figure~\ref{fig:pred_acc}\chg{b}).
\chg{On the simulated TF data all three methods likewise reach comparable accuracy, \rev{$0.93$--$0.96$} on the clean balanced dataset and \rev{$0.77$--$0.80$ (BNN $0.67$ in one seed)} once \rev{half of the JUN and MEF2A labels} are swapped (Figure~\ref{fig:pred_acc}a), confirming that the label noise is what degrades performance rather than any difference between the UQ methods.} \rev{With half of the labels swapped the two classes cannot be told apart from their labels, so the best attainable accuracy on the clean test set is $(4 + 2 \times 0.5)/6 = 0.83$, slightly below which all three methods land.}
BNNs require substantially longer training to reach comparable performance. Especially in the multi-label setting, BNN tends towards prediction values closer to zero or one than the deterministic and MCD models\chg{,} \rev{and does not reach comparable performance to the other models in one of the five random seeds on balanced multi-label data ($0.65$ against $0.92$--$0.95$ in the other four) and in two seeds on imbalanced multi-label data ($0.67$ and $0.79$ against $0.84$--$0.92$), while MCD and ENS vary by less than $0.03$ across seeds.} \rev{We report all runs in every figure rather than excluding the seeds in which BNN reached lower accuracy, and we note where this affects a result.}

As for the single-cell gene expression task, all UQ methods reach high and comparable accuracy on cell-type prediction, again ensuring that the uncertainty comparison is not confounded by differences in performance (Figure~\ref{fig:gex_pred_acc}\chg{c}). \rev{On the imbalanced version all three methods stay within $0.95$--$0.99$ in every seed.}

\chg{Average precision, which unlike accuracy does not depend on a decision threshold, gives the same ordering of methods and settings (Figure~\ref{fig:pred_acc}d--f).} \rev{The two proper scoring rules follow accuracy: NLL and Brier score are highest for BNN on the mislabeled TF data and on the multi-label RBP data, where the seeds in which it reached lower accuracy also give the highest NLL (Figure~\ref{fig:pred_acc}h,k), and lowest for BNN on the single-cell data, where ENS has the highest NLL and Brier score in both settings (Figure~\ref{fig:pred_acc}i,l).}

\subsection{Agreement of uncertainty scores}\label{sec_4_2:seq_agree}
We next ask whether the three UQ methods agree on which samples are uncertain. We measure agreement in three complementary ways introduced in Section~\ref{sec_3:framework}: the rank correlation of uncertainty scores (Kendall's tau), the overlap of the top-100 most uncertain samples per class, and the KS test on the SD distributions.

\chg{Across the simulated TF data (Figure~\ref{supp_fig:kendall_noise}a,b) and the single-label real RBP settings (Figure~\ref{fig:kendall}e,f), Kendall's tau is positive in all scenarios, with $p$-values rejecting the null hypothesis of no association.} \rev{On the sequence data the three pairings agree to a similar degree (median $\tau$ between $0.51$ and $0.78$ on the TF and the single-label RBP settings), and agreement decreases as the task becomes more complex, e.g. in the multi-label setting. On the single-cell data (Figure~\ref{fig:kendall}c,d) BNN and ENS remain positively correlated while MCD's ranking is only weakly associated with either, which we return to below.}

\begin{figure}[t]
    \centering
    \includegraphics[width=\textwidth]{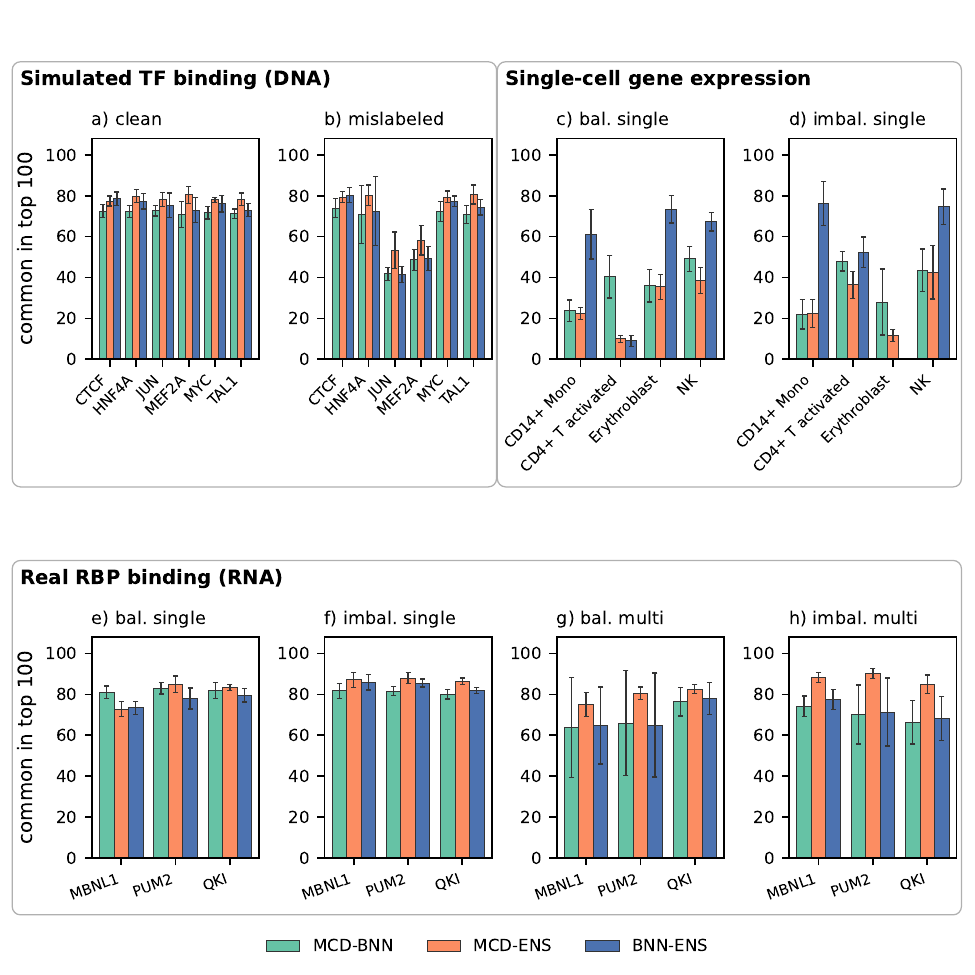}
    \caption{Overlap of the most uncertain samples identified by each pair of UQ methods. For each class, the $100$ test samples with the highest uncertainty score \rev{(standard deviation of the predicted class's probability)} were selected independently per method; bars give the number common to both members of a pair. Panels cover the simulated TF binding task (a, b), the single-cell task (c, d) and the real RBP task (e--h). Bars are the mean over $5$ random seeds, error bars one standard deviation. \rev{Green bars are the MCD--BNN pair, orange MCD--ENS and blue BNN--ENS.}}
    \label{fig:common_uncertain_100}
    \label{supp_fig:gex_most_uncertain}
\end{figure}

\begin{figure}[t]
    \centering
    \includegraphics[width=\textwidth]{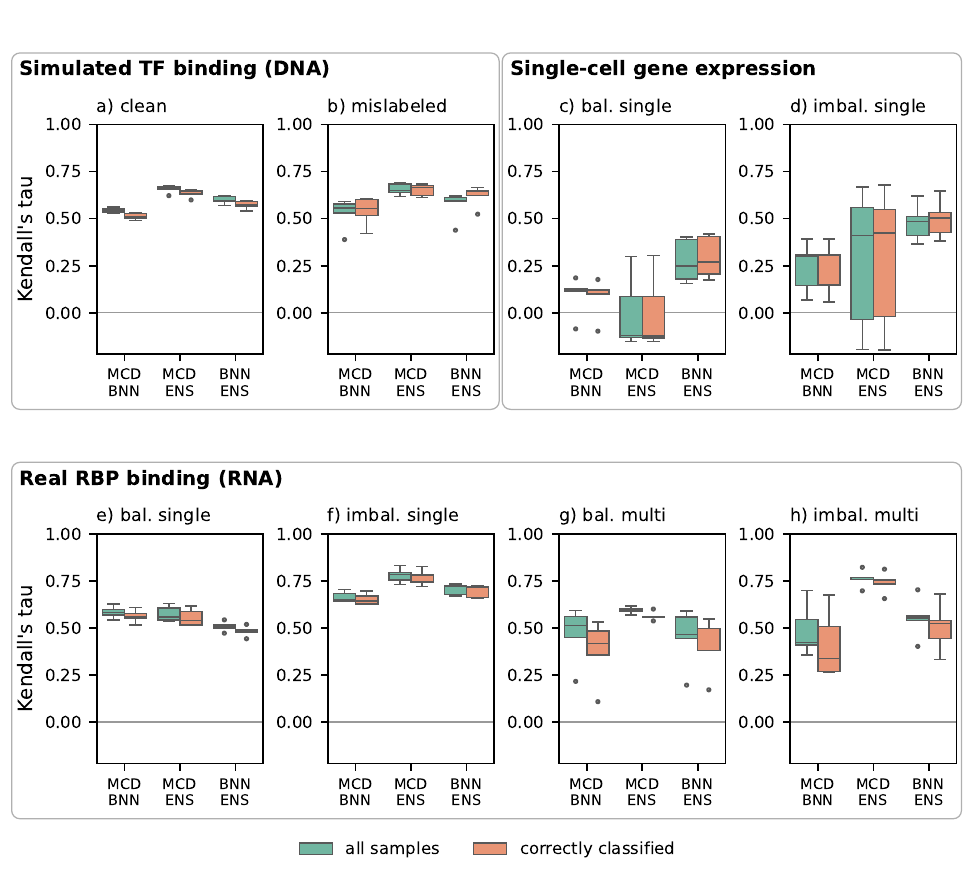}
    \caption{Agreement of the uncertainty scores of each pair of UQ methods, measured by Kendall's tau. Rank correlation between the standard deviations \rev{of the true class's probability} assigned to the same test samples, for the simulated TF binding task (a, b), the single-cell task (c, d) and the real RBP task (e--h). Within each panel, values are shown for all test samples \rev{(green)} and for the subset classified correctly by both methods \rev{(orange)}. Box plots summarize $5$ random seeds; the \chg{gray} line marks $\tau = 0$.}
    \label{fig:kendall}
    \label{supp_fig:kendall_noise}
    \label{supp_fig:gex_kendall_tau}
\end{figure}

\begin{figure}[t]
    \centering
        \includegraphics[width=\textwidth]{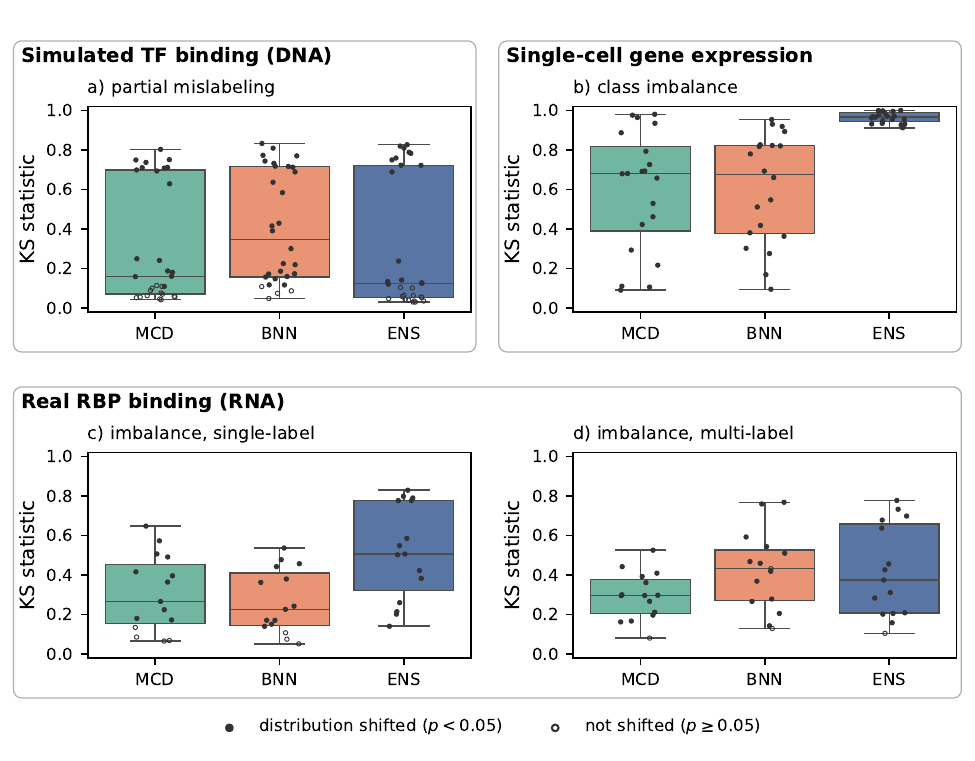}
    \caption{Two-sided Kolmogorov--Smirnov test on the uncertainty-score distributions. For each method, the predictive standard deviations for a given class \rev{(over all test samples)} are compared between a reference setting and a perturbed one: clean balanced versus \rev{JUN/MEF2A-mislabeled} balanced for TF (a), balanced versus imbalanced for the single-cell data (b), and balanced single-label versus imbalanced single-label for RBP (c), and balanced multi-label versus imbalanced multi-label for RBP use case (d). \rev{Boxes are green for MCD, orange for BNN and blue for ENS.} Each point is one class in one of $5$ seeds; filled points reject the null at $p<0.05$\rev{, open points do not}. A larger KS statistic therefore means the method's uncertainty scores for that class moved further when the perturbation was introduced; a value near zero means they did not respond to it at all.}
    \label{fig:ks}
    \label{supp_fig:gex_ks}
\end{figure}

The overlap of the top-100 most uncertain samples on the real RBP data (Figure~\ref{fig:common_uncertain_100}\chg{e--h}) shows the same pattern: \rev{averaged over classes, the three pairings share $69$--$88$ of their top-100 samples, MCD and ENS the most in three of the four settings, and the overlap of the pairings that involve BNN shrinks with task complexity ($77$--$85$ on single-label, $69$--$72$ on multi-label data, with single classes as low as $64$)}. \rev{On the simulated TF data (Figure~\ref{fig:common_uncertain_100}a,b) all three pairings share $71$--$80$ of their top-100 most uncertain sequences per TF on clean data; on mislabeled data the overlap drops to $42$--$58$ for the two swapped classes, for every pairing. Excluding the seeds in which BNN reached lower accuracy (Section~\ref{sec_4_1:perf}) raises the overlap of the BNN pairings on balanced multi-label RBP data from $69$ to $76$--$77$ and their median tau from $0.47$--$0.51$ to $0.51$--$0.54$, and changes nothing on the imbalanced multi-label data.} This is expected, as the methods rest on different posterior approximations and therefore need not agree on the exact ranking of the SDs. \chg{The KS test (Figure~\ref{fig:ks}c,d) shows that every method's uncertainty distribution shifts when class imbalance is introduced, but by very different amounts, which we return to in Section~\ref{sec_4_3:seq_sens}.} Together these results show that the choice of UQ method matters, and that agreement between methods can itself be informative when no ground-truth labels are available.

\chg{\rev{On the GEX data,} the overlap of the top-100 most uncertain samples (Figure~\ref{supp_fig:gex_most_uncertain}c,d), Kendall's tau (Figure~\ref{supp_fig:gex_kendall_tau}c,d) and the KS test (Figure~\ref{supp_fig:gex_ks}b) all show BNN and ENS agreeing more closely with each other than with MCD\rev{, unlike on the sequence data, where the three pairings agree to a similar degree}. \rev{On balanced single-label data MCD's uncertainty ranking is only weakly associated with BNN's (median $\tau=0.12$) and anticorrelated with ENS's ($-0.12$), whereas the corresponding balanced single-label RBP setting gives $\tau=0.58$ and $0.56$; with imbalance the two MCD pairings move to $0.30$ and $0.41$. The reason is the scale of MC-dropout's scores on this task, which we examine next.}}

\rev{On the single-cell data MC-dropout is uncertain about every cell: its median SD at the predicted class is $0.14$--$0.22$ on balanced and $0.14$--$0.15$ on imbalanced data, at an accuracy of $0.98$--$0.99$, and its ranking is only weakly associated with those of BNN and ENS (Section~\ref{app_b:umap_unc} shows that it still concentrates on the misclassified cells). On the sequence data its scores are small (median SD at the predicted class between $10^{-5}$ and $10^{-1}$) but rank the samples consistently with the other two methods (Section~\ref{app_b:mcd_zero}).}

\chg{\rev{A second observation is} that for the minority Erythroblast class the BNN and ENS top-100 sets are disjoint in three of five seeds (Figure~\ref{supp_fig:gex_most_uncertain}d). \rev{Section~\ref{app_b:umap_unc} shows where each method's uncertain cells lie in the embedding: BNN's and MCD's at the CD4+ T activated / NK boundary where the misclassified cells are, ENS's across one cell type.} This is a property of the standard deviation as a score rather than a disagreement about which cells are uncertain, and it affects three of $96$ class-by-method combinations.}

\clearpage

\subsection{Sensitivity to sources of uncertainty}\label{sec_4_3:seq_sens}
We now ask whether a UQ method reflects a specific source of uncertainty more strongly than the others. We take a clean single-label balanced dataset as the reference, introduce one source of uncertainty at a time, and ask whether the distribution of uncertainty scores shifts relative to the clean case. 

\begin{figure}
     \centering
     \includegraphics[width=\textwidth]{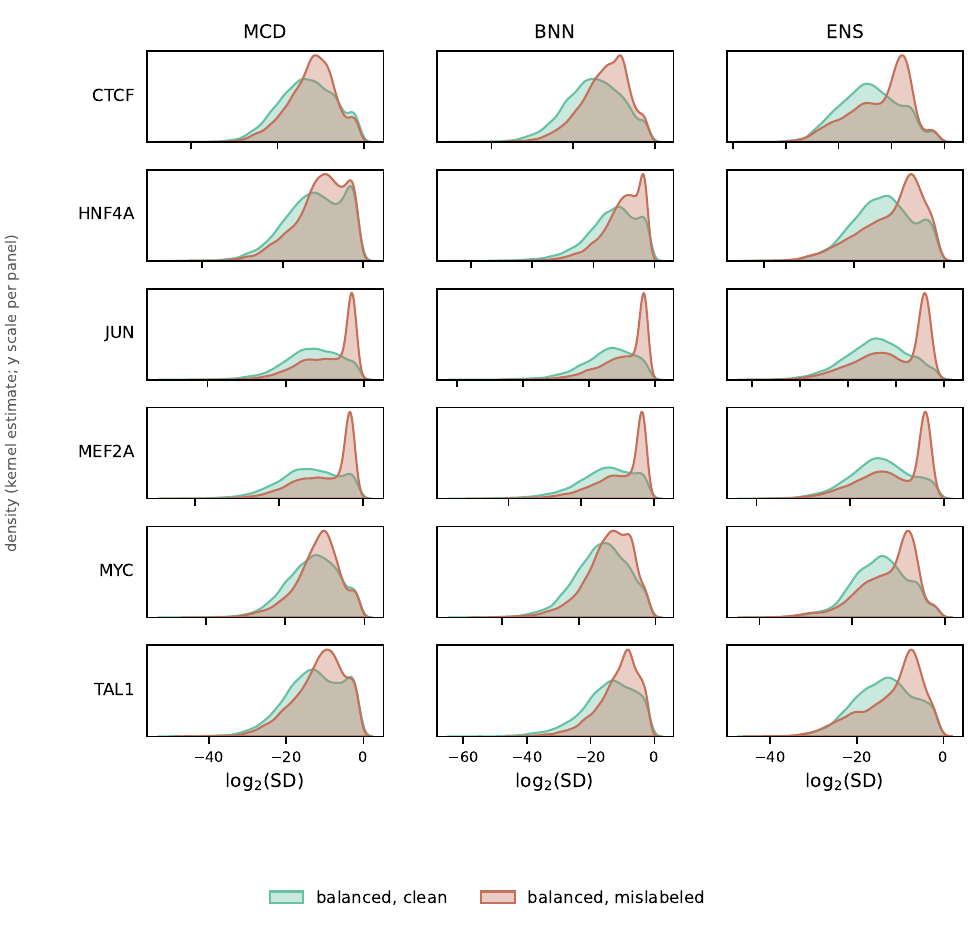}
     \caption{\textbf{Sensitivity of uncertainty scores to label noise on simulated TF data.} Distribution of uncertainty scores when the model is trained and tested on clean balanced single-label data \chg{(green)} versus the same model under partially mislabeled training data \chg{(orange; label swap between JUN and MEF2A)}. \chg{Rows are the six transcription factors; columns} are MCD, BNN, and ENS. \chg{Scores are on a $\log_2$ scale.} \rev{For each class the score is the standard deviation of that class's probability over all test samples; the five random seeds are pooled, and per-seed shifts are given in Table~}\ref{supp_tab:seed_shift}\rev{.}}
     \label{fig:tf_shift}
\end{figure}

\begin{figure}
\centering
    \includegraphics[width=\textwidth]{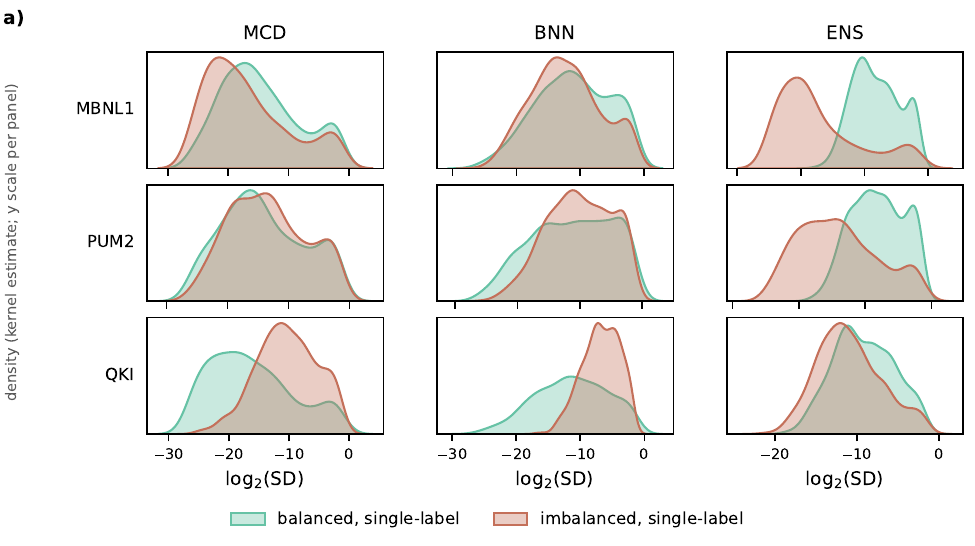}

    \vspace{0.5em}
    \includegraphics[width=\textwidth]{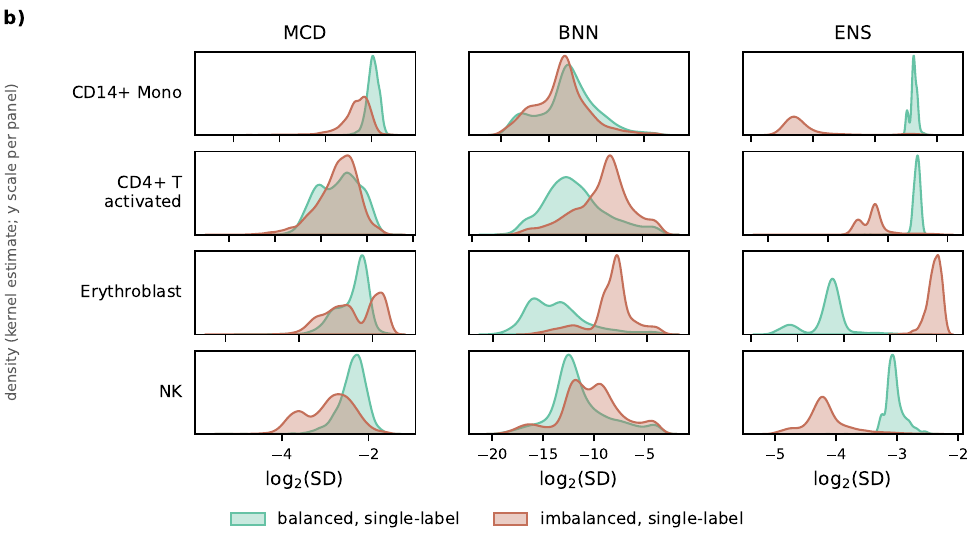}
    \caption{\chg{\textbf{Sensitivity of uncertainty scores to class imbalance for the real RBP binding and single-cell gene expression.} Columns are MCD, BNN, and ENS; green is the model trained on balanced training set, orange is the model trained on the imbalanced one. All models tested on \rev{the same} balanced test set.  a) real RBP binding use case, rows are MBNL1, PUM2 and QKI; b) single-cell gene expression use case, rows are CD14+ Mono, CD4+ T activated, Erythroblast and NK. Scores are on a $\log_2$ scale\rev{. For each class the score is the standard deviation of that class's probability over the test samples the model predicts as that class, pooled over the five random seeds; the per-seed shifts at the predicted class are in Table~\ref{supp_tab:entropy_mi}, and paired on the shared test set in Table~\ref{supp_tab:seed_shift}}.}}
    \label{fig:sd_shift_imbalance}
    \label{supp_fig:rbp_sd_shift}
    \label{supp_fig:gex_sd_shift}  
\end{figure}

The simulated TF data is ideal for this because we control exactly which source is present. Here, we introduced label noise (swapping the labels of \rev{half of the JUN and MEF2A samples}) and missing motifs (background sequences with no planted motif). Figure~\ref{fig:tf_shift} shows the resulting shifts in the SD distribution\chg{, on a $\log_2$ scale. Raw-scale figures are available in supplementary Figure~\ref{supp_fig:tf_shift_raw}, and Figure~\ref{supp_fig:sd_shift_raw_real}. The missing-motif experiment is reported in Section~\ref{app_b:tf_no_motif}}. \rev{All three methods confine the largest increase in uncertainty to the two mislabeled classes. For BNN,} averaged over the five seeds, \rev{the} median $\log_2$(SD) rises by $6.5$ units on JUN and MEF2A but by $3.2$ on the four classes whose labels were untouched, and the corrupted classes shift more than the untouched ones in every seed. ENS shows the same pattern more weakly ($4.8$ against $3.6$, also in every seed), and MCD the most selectively ($6.2$ against $1.7$, in every seed); read at the predicted class instead (Table~\ref{supp_tab:entropy_mi}), the ordering of the methods changes, so we do not rank them on label noise. The untouched classes do shift for all methods, which we attribute to the shared feature layers: label noise on two classes changes the learned filters that all six classes rely on. Per-seed values are given in Table~\ref{supp_tab:seed_shift}, and Section~\ref{app_b:noise_rate} repeats the experiment with $15\%$ and $30\%$ of the labels swapped, with the same outcome. \rev{Swapped labels are aleatoric noise, which a well-specified posterior would express in the mean prediction rather than in disagreement between passes. The mutual information, which measures that disagreement, rises for the swapped classes as the SD does (Table~\ref{supp_tab:entropy_mi}), so the passes disagree more on the corrupted classes, consistent with the models fitting the noisy labels. The rise in SD therefore marks the classes whose labels were corrupted; it is not by itself the correct response to aleatoric noise.} \chg{For} the missing-motif case the methods behave similarly, \chg{every class shifting by a large and roughly equal amount, which is expected when the predictive signal is removed from the input altogether}. We did not find a consistent winner for mislabeled or missing-motif noise \chg{in terms of the size of the shift; the difference between the methods lies in whether the shift is specific to the affected classes}. On the real RBP data, \rev{BNN reflects class imbalance in the direction expected of an epistemic score. The imbalanced setting enriches the two larger classes (MBNL1 $\times 10.2$, PUM2 $\times 2.8$) and leaves QKI unchanged, so that QKI becomes the relatively rare class. Reading the score at the predicted class and evaluating both settings on the same balanced test set (Figure~\ref{supp_fig:rbp_sd_shift}a), BNN becomes less uncertain when it predicts MBNL1 (median $\log_2$(SD) shift $-1.4$, four of five seeds) and more uncertain when it predicts QKI ($+4.5$, all five seeds; paired on the samples the reference model predicts as the class, Table~\ref{supp_tab:seed_shift}, while Figure~\ref{supp_fig:rbp_sd_shift}a pools each setting's own predictions). ENS becomes more confident in every class, including QKI, in every seed ($-13.6$ to $-1.8$), and MCD shows the same direction as BNN, with a larger shift ($-3.9$ for MBNL1 and $+8.2$ for QKI, in all five seeds). Pooled over all test samples rather than the predicted class, BNN's and MCD's uncertainty rises for every class ($+1.0$ to $+2.8$), while ENS still becomes uniformly more confident; both readings are given in Table~\ref{supp_tab:seed_shift}.} \chg{We quantified the shift over all methods, classes and $5$ random seeds with the KS test (Figure~\ref{fig:ks}c,d), which rejected the null hypothesis of no shift in \rev{$38$ of $45$ single-label and $41$ of $45$ multi-label} tests, confirming that the uncertainty distribution does move when imbalance is introduced. Across all use cases and perturbations, \rev{$198$ of $240$} tests reject the null\rev{, and $196$ after Benjamini--Hochberg correction over all $240$ tests at a false-discovery rate of $0.05$}.}

\chg{The KS statistic measures magnitude but not direction, and here direction is what matters: a UQ method is only useful if imbalance makes it \emph{more} uncertain. Comparing the median $\log_2$(SD) between settings, per class and per seed, \rev{BNN becomes more uncertain about the class the imbalance depleted, Erythroblast, in every seed ($+5.6$ on average at the predicted class, Figure~\ref{supp_fig:gex_sd_shift}b, Table~\ref{supp_tab:entropy_mi}), and it is the only method that shows the expected direction on both use cases; MCD does so on the RBP data but not on the single-cell data. ENS also becomes more uncertain about Erythroblast when read at the predicted class ($+1.2$), but on RBP it becomes} \emph{more} confident \rev{in every class and seed}, so its KS statistic is the largest of the three although it reflects imbalance least usefully. \rev{MCD does not respond: every class stays within $\pm0.8$ at the predicted class, and Erythroblast does not move at all}.}

\chg{The two use cases test this differently. The single-cell setting depletes its minority class, cutting Erythroblast from $2673$ to $587$ training cells ($-78\%$); the RBP setting leaves its smallest class untouched (QKI, $816$ to $825$) and creates the imbalance by enriching the majorities instead (MBNL1, \rev{$+916\%$}). Only the single-cell experiment therefore asks whether the depleted class becomes the most uncertain one. \rev{Reading the score at the predicted class, all three methods rank the classes in the expected order (median Spearman correlation between the change in a class's training-set size and the shift in its uncertainty $-0.8$ for BNN, $-1.0$ for ENS and $-0.8$ for MCD; Section~\ref{app_b:depletion}), but they differ in how far the depleted class moves: $+5.6$ $\log_2$ units for BNN against $+1.2$ for ENS and $0.0$ for MCD. With four classes per seed the correlation is descriptive and carries no significance; what supports the claim is the size of the shift and its sign in every seed. BNN is the method whose uncertainty responds most to which classes the training set under-represents.}}

The sensitivity result also transfers across modalities. \chg{As on the RBP data, all three methods respond to class imbalance in the cell type prediction case, but \rev{BNN does so most specifically to the depleted class}} (Figure~\ref{supp_fig:gex_sd_shift}). \chg{Among the two smallest classes, Erythroblast and CD4+ T activated, BNN shows a clear directed shift towards higher uncertainty, \rev{and the KS test confirms that all three distributions move (median KS for CD4+ T activated $0.29$ for MCD, $0.66$ for BNN and $0.93$ for ENS), but only BNN moves in the expected direction.} Measured by KS magnitude alone ENS responds most in all four classes (Figure~\ref{supp_fig:gex_ks}b), the same pattern as on the single-label RBP data. \rev{BNN is therefore the method whose response to imbalance is most specific to the depleted class, and it shows this on the single-cell data in every seed.}}

\begin{figure}
    \centering
    \includegraphics[width=\textwidth]{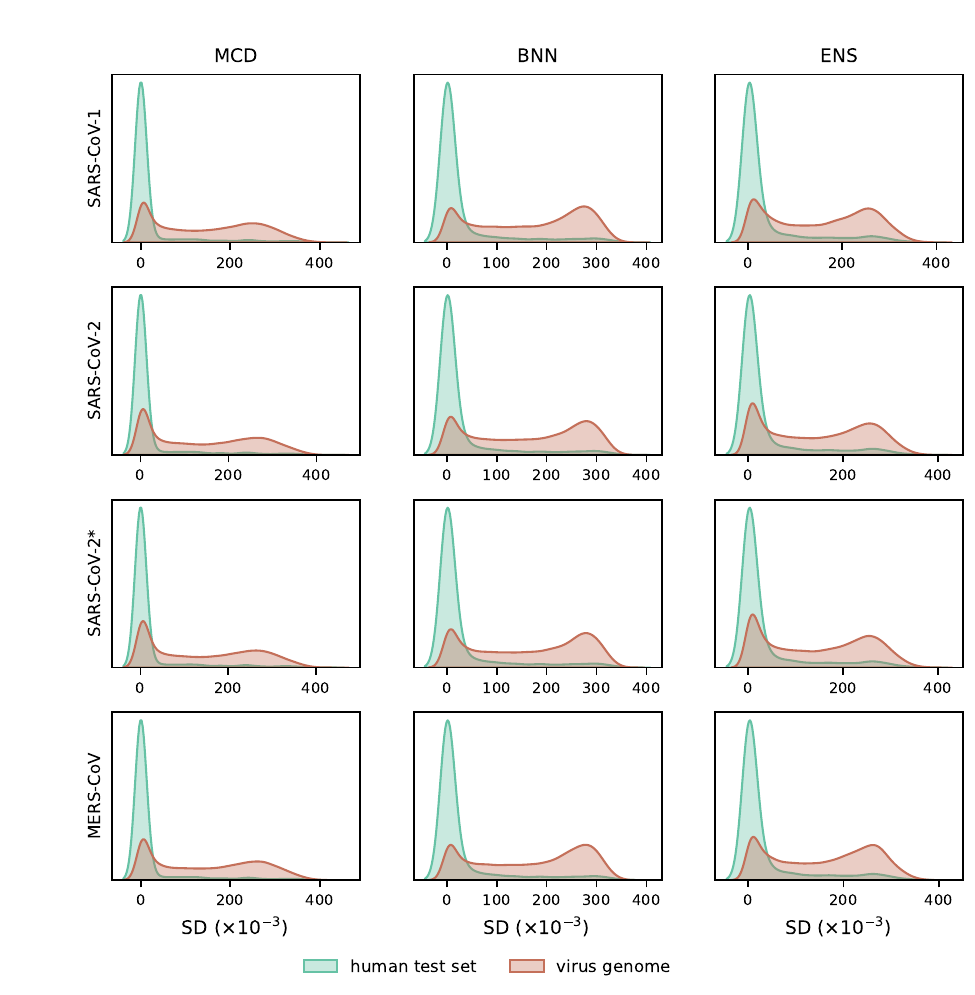}
    \caption{\textbf{Uncertainty increases on out-of-distribution sequences (real RBP use case).} Models trained on human single-label RBP data and tested on \chg{coronavirus genomes} as an out-of-distribution case. \chg{Green}: uncertainty distribution on the human (in-distribution) test set; \chg{orange}: on the out-of-distribution data. \chg{Rows are virus genomes and columns are UQ methods; SARS-CoV-2* denotes the second SARS-CoV-2 assembly (wuhCor1). \rev{All columns share the same horizontal scale.} \rev{The five random seeds are pooled; detection metrics per seed are in Table~\ref{supp_tab:ood}.}}}
        \label{fig:shift_virus}
\end{figure}

\chg{The same pattern extends to a genuinely out-of-distribution case. \rev{BNN again shows the largest shift on the probability scale when the model trained on the human genome is tested on coronavirus genomes: its median uncertainty on viral windows is $0.17$, against $0.15$ for ENS and $0.12$ for MCD, from in-distribution medians of $4\times10^{-4}$, $2\times10^{-3}$ and $1\times10^{-5}$ (Figure~\ref{fig:shift_virus}). Against a $95$th-percentile human threshold all three methods flag about a third of the viral windows as unusually uncertain ($35\%$ for BNN, $26\%$ for ENS, $41\%$ for MCD), and as a ranking score for detecting the viral windows all three work (AUROC $0.87$, $0.84$ and $0.89$; Table~\ref{supp_tab:ood}). The methods differ in how far the score moves, not in whether it puts the shifted inputs on top; on the ranking metrics MCD is slightly ahead of BNN (Table~\ref{supp_tab:ood}).} Section~\ref{app_b:ood} of the supplement elaborates on the out-of-distribution experiments.}

\rev{The results in Figures~\ref{fig:tf_shift}, \ref{fig:sd_shift_imbalance} and~\ref{fig:shift_virus} are the central empirical finding of this study: in the tested settings, BNN is the only method that gives a consistent, directionally specific uncertainty response to class imbalance and out-of-distribution data and marks the corrupted classes under label noise, with one configuration. MC-dropout matches it on the sequence data, on label noise, class imbalance and the ranking of out-of-distribution sequences, but on the single-cell data it is uncertain about every cell and does not respond to imbalance; with dropout active in a single layer its scores on the sequence data are near zero (Section~\ref{app_b:mcd_zero}). ENS responds, but less specifically, and under imbalance on the RBP data by becoming uniformly more confident.}

\clearpage

\subsection{Filtering predictions by uncertainty}\label{sec_4_4:filter}
\begin{figure}
     \centering
    \includegraphics[width=\textwidth]{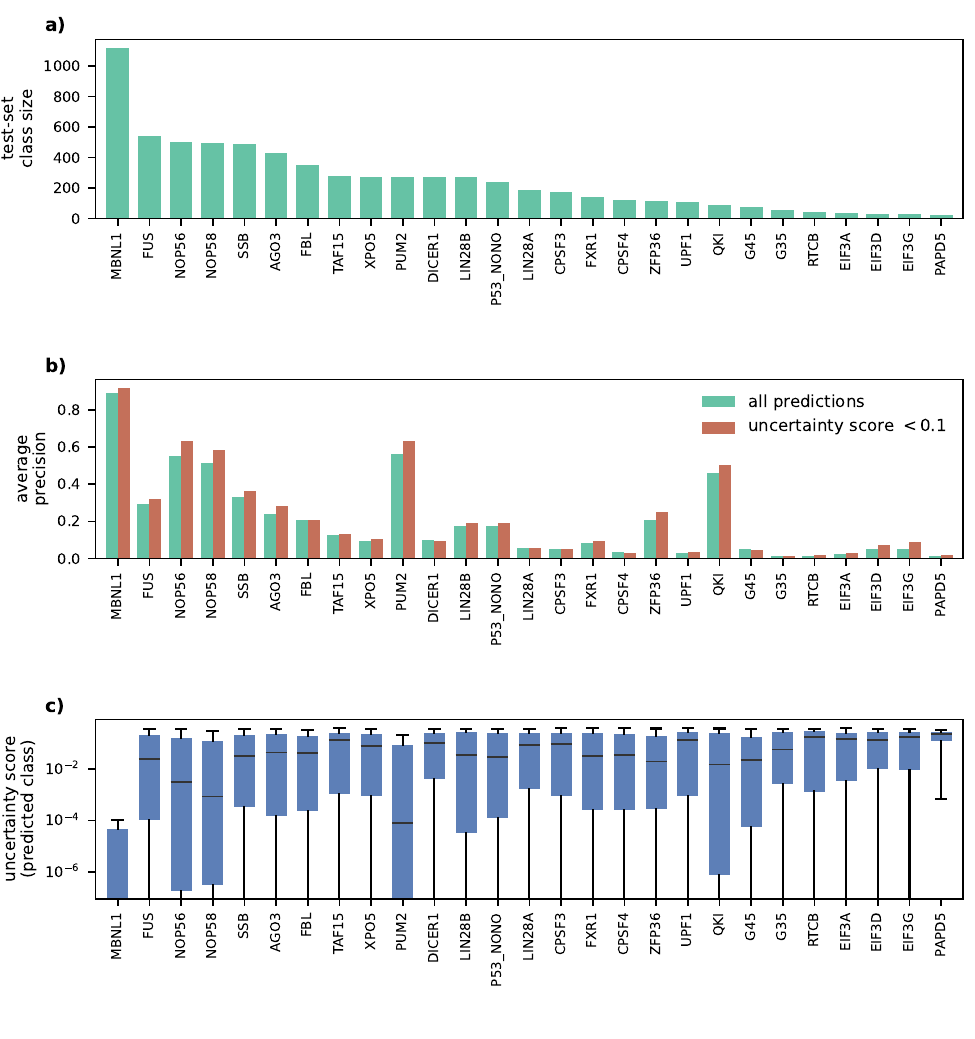}
    \caption{\chg{\textbf{Uncertainty filtering on the 27-class real RBP task.} a) Number of test samples per RBP class, ordered by class size. b) Average precision (AP) per class for a BNN trained and tested on this dataset, for all predictions \rev{(green)} and for the subset whose uncertainty score is below $0.1$ for the predicted class \rev{(orange)}. c) Uncertainty score of every test sequence, grouped by the predicted class \rev{(blue)}. The axis is logarithmic: the score spans seven orders of magnitude, and MBNL1's median is zero.}}
    \label{fig:27_rbp_class_imba}
    \label{supp_fig:ap_improvement}
\end{figure}

The controlled experiments above use small datasets for a fair and direct comparison. To test the practical impact of UQ scores on a larger, noisier task, we selected 27 RBPs of varying class sizes from PAR-CLIP experiments and trained a BNN to predict RBP binding\rev{. We use the BNN because it showed the most directionally specific response to class imbalance in Section~\ref{sec_4_3:seq_sens}, and this task is strongly imbalanced. This section is not part of the controlled comparison: with class imbalance, label noise and coverage differences all present at once, a difference between methods here could not be attributed to any one of them, so we illustrate what one method's score does in a realistic setting rather than compare the three.} Figure~\ref{fig:27_rbp_class_imba}a) shows the class sizes and imbalance for this dataset. Model architecture, hyper-parameters, and pre-processing are chosen as in the benchmark experiments above. We define a high-quality prediction as one with an uncertainty score below $0.1$ for the predicted class (note that we \chg{expect} this threshold to be application- and dataset-dependent). \rev{The threshold was fixed in advance as a round value on the probability scale and not tuned; Section~\ref{app_b:app27_extended} gives average precision against coverage for every threshold, with bootstrap intervals, so that other operating points can be read off.} Filtering on this criterion improves average precision per class (Figure~\ref{fig:27_rbp_class_imba}b))\chg{: it retains $65\%$ of the predictions and raises AP in $20$ of the $27$ classes, with the mean over classes rising from $0.20$ to $0.22$}, and the precision-recall curves for 5 individual RBPs show the corresponding gain (\chg{Figure~\ref{supp_fig:ap_curves}}). This demonstrates that uncertainty scores can be put to direct use in selecting reliable predictions for downstream analysis.
\chg{The uncertainty score also tracks how much data each protein contributed. Grouping the scores by predicted class (Figure~\ref{fig:27_rbp_class_imba}c), the median uncertainty rises steadily as the training-set class size falls: it is zero for MBNL1, the largest class with $7354$ training sequences, and $0.22$ for PAPD5, the smallest with $148$. Across all $27$ classes, the Spearman correlation between training-set size and median uncertainty is $-0.59$ ($p=0.001$), and the five smallest classes have a median $57$ times that of the five largest. This is the class-imbalance behaviour discussed in Section~\ref{sec_4_3:seq_sens}.} \rev{Because the score is highest for the rarest proteins, the filter removes proportionally more of their predictions: it retains $57\%$ of the positive sequences of the five smallest classes against $68\%$ of the five largest, with no class removed entirely (per-class coverage in Table~\ref{supp_tab:coverage}). In applications where a missed detection in a rare class is costly, the threshold should be set per class or the retained fraction reported alongside any aggregate gain.}
\chg{Section~\ref{app_b:app27_extended} of the supplement examines what drives the uncertainty on this task: incorrectly classified sequences carry no localized motif evidence, and the model's errors concentrate between RBPs that belong to the same protein complex, where the single-label ground truth is itself partly ambiguous.}

\section{Conclusion and Future Work}\label{sec_5:conclusion}

This work presents a structured framework for evaluating UQ methods in deep learning, one of the first of its kind for genomics applications. It provides initial guidelines for selecting the most appropriate among several popular methods. We systematically compared these methods across two genomics modalities - DNA/RNA sequence and (single-cell) gene expression -  focusing on dataset 
characteristics particularly relevant to genomics, such as strong class imbalance, out-of-distribution data, and label noise.
We introduce two complementary evaluation perspectives that do not rely on ground truth labels. First, we assess the agreement between UQ methods by identifying the most uncertain samples and measuring their overlap, an approach useful when working with noisy labels or applying pre-trained models to new unlabeled datasets. Second, we evaluate UQ quality through distributional shifts in uncertainty scores when a single source of uncertainty is introduced to otherwise clean data. To the best of our knowledge, no prior study has applied either of these perspectives to compare UQ methods in genomics.

\rev{From our observations, BNN produces stronger and more directionally specific responses to sources of uncertainty commonly encountered in biological applications, particularly class imbalance and out-of-distribution data, in the settings we tested. This holds when the BNN is trained to comparable accuracy, which required more training steps than the other methods and was not reached in every seed.}
\rev{On the sequence data the three methods largely agree on which samples are uncertain; on the single-cell data BNN and ENS agree with each other but not with MCD, whose scores there are large for every cell.} The practical cost of the three methods differs. BNNs need longer training and more computational resources in every scenario, and they were sensitive to the optimizer and the learning rate, with some runs in the multi-label setting not converging until we \rev{trained for more steps} and allowed longer patience for early stopping. MC-dropout was insensitive to these choices, so the better sensitivity we observe for BNN comes with a real cost in tuning effort (Section~\ref{app_a:impl}). \rev{MC-dropout's scores depend instead on where dropout is active at inference and at what rate (Section~\ref{app_b:mcd_zero}).} \rev{In the two data types and the applications we tested, we observe the following in practice. BNN responds to class imbalance on both data types and to label noise and out-of-distribution data on the sequence data, and is the method to choose when the question is which classes or inputs the training data under-represent; it costs longer training and a learning-rate search, and it did not converge in every seed. MC-dropout with all layers active covers label noise, class imbalance and out-of-distribution data on the sequence data at a fraction of the cost, but does not respond to class imbalance on the single-cell data, and its score depends on the dropout configuration. Deep ensembles are stable to train, with accuracy varying by less than $0.03$ across seeds and a ranking that is stable from five members, and give a usable score in every setting, but respond least specifically to the perturbed classes and become more confident, not less, under imbalance on the RBP data.} We further demonstrate that uncertainty scores can be used to filter predictions, improving average precision in a realistic $27$-class RBP binding task. These findings provide practical guidelines for computational biologists in choosing UQ methods based on their data characteristics, and ways to go about evaluating uncertainty estimates beyond \rev{its} relation to model accuracy. 

\rev{Both evaluation perspectives can be applied without a clean reference dataset. The agreement diagnostics need no reference at all: they compare methods on the same unlabeled data. The shift diagnostics need a perturbation, which a practitioner can construct on their own data by subsampling a class or permuting labels on a held-out slice and measuring how far the uncertainty scores move, as we did here.}

\rev{Biological data carry noise that no benchmark can remove: PAR-CLIP binding sites include false positives, and cell-type annotations are themselves model- or expert-derived. Our clean settings are therefore the optimistic case rather than a noise-free one. The mislabeling experiment is our controlled version of this noise: rather than assuming it away, we add a known amount and measure the response.}

\chg{Our empirical framework comes with limitations, which we set out here. Our comparison covers two genomics modalities and a fixed set of experimental scenarios, so which method performs best should be read as a property of these settings rather than as a general ranking.} \rev{Recent work on the evaluation of Bayesian deep learning methods makes the same point in general: method rankings are dataset-dependent and single-seed estimates of a metric are themselves random variables, with deep ensembles among the methods most affected~\citep{zhan2026unstable, zhan2026variances}, so we report every result over five seeds and state the direction of each effect per seed.} The present study evaluated UQ in a classification context, whereas many applications in genomics involve regression or unsupervised settings. \rev{The framework requires a task in which a single source of uncertainty can be introduced at a time, so we used classification on two data types; extending it to regression, to other sequence tasks such as predicting TF activity or classifying sequences by function or species, and to further tissues and species is left for future work.} Our study is also limited to neural network-based models, yet many actively used computational genomics solutions are still based on more traditional ML or statistical methods. On the other hand, future work should clearly evaluate how UQ methods play out on foundation models, e.g. when using the embeddings as input for smaller models for task-specific fine-tuning, expanding on the single-cell gene expression scenario covered here. \rev{The single-cell results are also conditional on the Liam embedding: a model trained on an embedding can only be uncertain about differences between cells that the embedding preserves, so embeddings that preserve the local geometry or topology of the data~\citep{zang2022dlme,li2021invml} may change which cells are flagged as uncertain, and comparing UQ methods across embeddings is a direction for future work.} We focused on aggregated predictive uncertainty, not differentiating between epistemic and aleatoric uncertainty\citep{huellermeier2021aleatoric, kendall2017uncertainties}, which remains an important direction for future work. \rev{We also did not vary the prior or the weighting of the KL term of the BNN, and MC-dropout's rate and placement only as far as Section~\ref{app_b:mcd_zero} goes; each method is compared in one configuration.} Finally, UQ methods produce different absolute ranges of uncertainty scores, and thresholds for filtering predictions will likely need to be determined in an application- and dataset-specific manner.

Taken together, our empirical analysis provides a novel framework for evaluating uncertainty estimation methods; we expect that its insights will contribute to an increased adoption of UQ within the practical context of genomics applications.

The accompanying code base is available \href{https://anonymous.4open.science/r/empirical_uq_genomics-1F4D}{here} for reviewers.

\begin{ack}
S. Saran and U. Ohler were supported by the Initiative and Networking Fund of the Helmholtz Association (Helmholtz AI project UNITY, ZT-I-PF-5-149) as well as by the Helmholtz-Einstein International Berlin School in Data Science (HEIBRiDS). M. Ghanbari and U. Ohler were supported by DFG research unit DeSBI (KI-FOR 5363).
\end{ack}

\bibliography{references}
\bibliographystyle{tmlr}

\clearpage
\appendix

\renewcommand{\thesection}{S\arabic{section}}
\renewcommand{\thefigure}{S\arabic{figure}}
\renewcommand{\thetable}{S\arabic{table}}
\setcounter{section}{0}
\setcounter{figure}{0}
\setcounter{table}{0}


\begin{center}
  {\LARGE\bf\sffamily Supplementary Materials for\\[0.35em]
   Uncertainty-Aware Deep Learning for Genomics Applications:\\[0.15em]
   Insights from an Empirical Study\par}
\end{center}
\vskip 0.25in


\section{Information for reproducibility}\label{app_a}
The accompanying code base is available \href{https://anonymous.4open.science/r/empirical_uq_genomics-1F4D}{here} for reviewers.
\subsection{Use case I: predicting protein binding from DNA/RNA sequence}\label{app_a:data}
This section provides additional reproducibility detail for the datasets introduced in Section~\ref{sec_2:background} of the main text. We focus here on pre-processing, the derivation of the dataset settings, and the train/validation/test splits rather than repeating the high-level description. Scripts for data generation and pre-processing are provided in the code repository.

\subsubsection{Real Dataset}\label{app_a:data:real}
The real dataset we used, was generated by PAR-CLIP experiments. These experiments can biochemically detect short sequences of RNA (i.e. binding sites), to which a specific known protein has bound. The data provided by these experiments is the list of binding site positions for a given RBP. In this work, we used a collection of existing PAR-CLIP datasets\citep{neel2019} in the HEK293 cell line for three RBPs that are known to have strong sequence motifs: MBNL1, PUM2, and QKI. We selected these RBPs because our classifier performs well for proteins where there is a strong sequence preference for binding to be learned by the model. This helps us eliminate the uncertainty caused by lack of existence of a motif and enables us to build our comparison scheme for UQ in a more controlled setting. This data is publicly available and does not contain personal information.  

We followed the pre-processing steps from \citep{DeepRipe} and used the human GRCh37/hg19 reference for retrieving RNA sequence as the input for the classification task. The final dataset had one-hot-encoded RNA sequence windows of size $150$ nucleotide as the input and vectors of size three for the RBP classes as labels. 
The original data, was naturally multi-labeled, meaning there were instances of RNA sequence sections, where several RBPs would bind to different locations along this section. Also, the datasets had different coverage for different RBPs, therefore we had a different number of binding sites detected which caused the dataset to be strongly imbalanced (see Figure~\ref{fig:class_dist} in the main text).
To generate single-label data settings, we simply removed samples where binding sites from several RBPs were present. To create the balanced data scenario, we sub-sampled larger classes to be in the size of the smallest class. As a result, we generated four dataset settings for the RBP-binding task: single-label balanced, single-label imbalanced, multi-label balanced, and multi-label imbalanced. \rev{For the imbalanced settings of the RBP and single-cell tasks and for the mislabeled setting of the TF task, the test set is the test set of the corresponding balanced or clean split, and the perturbed training and validation sets are drawn from the remaining samples only (duplicate sequences removed), so that no test sample is seen in training. The imbalanced training sets therefore contain $12{,}392$ RBP sequences ($9{,}063$ MBNL1, $2{,}504$ PUM2, $825$ QKI) and about $16{,}000$ cells ($7{,}444$ CD14+ Mono, $3{,}487$ CD4+ T activated, $587$ Erythroblast, $4{,}515$ NK for seed $7$); in the mislabeled TF setting the labels of half of the JUN and half of the MEF2A training and validation sequences are swapped.}
To repeat our experiments across different random seeds, we repeated splitting each dataset into training, validation, and test sets with $5$ random seeds \rev{($68\%$, $17\%$ and $15\%$ of the sequences for the TF and RBP data; $60\%$, $20\%$ and $20\%$ of the cells for the single-cell data)}.

\subsubsection{Simulated Dataset}\label{app_a:data:sim}

To account for further dataset characteristics, we generated simulated datasets for a similar classification task to the RBP-binding, but this time for protein-binding events for DNA sequences. These proteins are called Transcription Factors (TFs) and similar to RBP, can influence the gene regulation process in the cell in various ways. The classification task remains the same: given a sequence of $150$-nucleotide length, predict which proteins bind to it. The input data here is a section of DNA sequences and the target is a selection of Transcription Factors. We chose $6$ TFs with widely known sequence motifs, namely, CTCF, HNF4A, JUN, MEF2A, MYC, and TAL1 as our targets for this classification task. We aimed to generate input sequences to create scenarios where the dataset has missing sequence motifs or the data is partially mislabeled. Both cases are common when working with noisy biological data. 

\begin{figure}
     \centering
    \includegraphics[width=\textwidth]{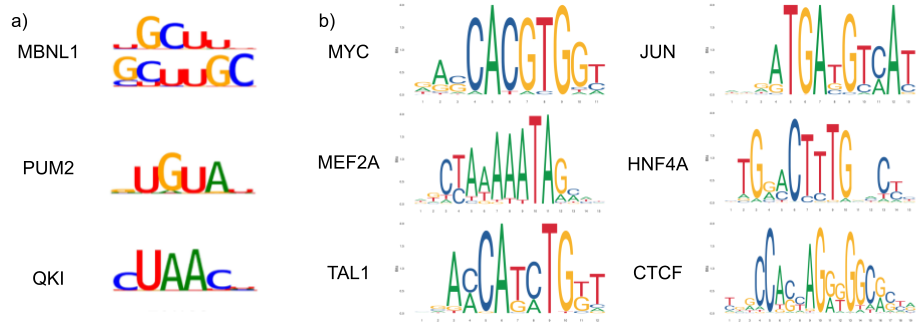}
    \caption{Example of sequence motifs for selected RBPs (a) and TFs (b). Some RBPs and TFs have preferences in the sequence patterns (i.e. motifs) onto which they bind. Identifying these motifs can help predict protein binding activities in RNA or DNA. a) RBP motifs are extracted using XAI on the DeepRiPe model \citep{DeepRipe}. b) TF sequence motifs are provided by JASPAR open-source database, version 2020 \citep{jaspar2020}. }
    \label{supp_fig:sample_motifs}
\end{figure}

To this end, we used SimDNA\citep{simdna} software to synthesize DNA sequences and plant sequence motifs in them. SimDNA is a tool developed in Python for generating Genomic regulatory sequences. It does so by first generating a background sequence and then embedding desired sequence motifs in the background. For our experiments, we have used the first-order Markov Model to generate DNA background sequences of length $150$ and embedded a single motif sequence for a TF class. We used the Jaspar open-source database\citep{jaspar2020} to obtain sequence motifs of our selected TFs. JASPAR provides manually curated TF binding profiles as position frequency matrices (PFMs). These PFMs are then used in SimDNA to embed a sequence motif of a TF into the background DNA sequence (see suppl. Figure~\ref{supp_fig:sample_motifs}). We generated three simulated datasets: a clean single-label balanced dataset as a noise-free reference, a noisy dataset where the sequences have no motifs, and a single-label balanced dataset where a portion of data is mislabeled. 

The simulated dataset consists of $2000$ samples per class\rev{, split into training, validation, and test sets as the real data (Section~\ref{app_a:data:real})} (the entire dataset size is $12000$). For the dataset with no motifs, we simply used the background sequences without implanting any motifs and for the mislabeled dataset, we swapped the labels between the two TF classes JUN and MEF2A for \rev{half of the samples of each ($16.6\%$ of all training samples)}. We repeat splitting each simulated dataset with $5$ random seeds, similar to the real datasets. 

\subsubsection{Modeling specification}\label{app_a:impl}
The base deterministic architecture is built based on DeepRiPe\citep{DeepRipe}. This network has two one-dimensional convolutional layers, each followed by a max-pooling layer and a drop-out rate of $0.25$. This is followed by two fully connected layers, the first one with $250$ nodes and a drop-out rate of $0.25$ and the last with $3$ or $6$ nodes, depending on the number of classes in the classification task. All layers except the last have ReLu for activation function. The deterministic model, as well as member models in ENS and MCD, have softmax activation for the output layer in the single-label dataset settings and the sigmoid function for multi-label. BNN doesn't have an activation function for the output layer, but the respective function will be applied to the output in a later stage. 

The ENS is a collection of $10$ deterministic models, each following the base model architecture. we initialized the member models with different random seeds and shuffled the training dataset with that seed. This is separate from the $5$ random seeds used to create different data splits. \rev{The ensemble therefore provides $10$ independent predictions per test sample. The member models are deterministic at inference, so repeating inference on a member returns the same prediction; we report the mean and standard deviation over the $10$ members.} 
\rev{For MCD, every drop-out layer of the trained network is active at inference, at its training rate: the three layers of rate $0.25$ on the sequence tasks and the two layers of rate $0.3$ on the single-cell classifier. Dropout is applied by calling the model in training mode for each of the $100$ passes. Section~\ref{app_b:mcd_zero} reports the variants with dropout in the last layer only and with a rate of $0.5$.}

In BNN implementation, for all network layers, we used FlipOut layers from TensorFlow Probability which assumes a normal distribution over weights and biases. For training the BNNs we used mean-field variational inference. The loss function for BNN is the sum of Kullback-Leibler (KL) Divergence loss and softmax cross-entropy loss with logits (for single-label) or sigmoid cross-entropy loss with logits (for multi-label) in TensorFlow. The loss function for deterministic, ENS, and MCD is just comprised of the cross-entropy loss.\citep{tensorflow2015-whitepaper}
BNN requires a longer training time and a larger number of epochs to train enough to converge and reach comparable accuracy to the other models. Training BNNs in the multi-label RBP setting was unstable across random seeds, with some runs not converging or getting stuck in local minima; this improved when we \rev{trained full-batch for more steps} and allowed longer patience for early stopping. We found BNNs sensitive to the choice of optimizer and hyper-parameters such as learning rate, whereas MC-dropout networks were robust to these choices, though the dropout rate and the choice of which layers to keep active at inference influence performance. Overall, the increased number of parameters and the approximate training of BNNs translate into a need for more computational resources and larger datasets. \rev{The BNN is trained full-batch, one optimizer step on the whole training set per epoch, for at most $1000$ epochs with early stopping on the validation loss (patience $50$ for single-label and $100$ for multi-label data) and a learning rate of $0.005$. Deterministic, ENS and MCD models are trained with mini-batches of $128$ for at most $50$ epochs with patience $5$, except for ENS on the simulated TF data ($100$ epochs, patience $10$), with a learning rate of $0.002$ (the submitted version stated $0.01$; $0.002$ is the rate the original runs used). All models use the Adam optimizer. All models of the controlled comparison were trained with this protocol; the 27-class model of Section~\ref{sec_4_4:filter} is the one of the submitted version.}

For generating distributions over predictions, the inference is repeated on the test set for $100$ iterations \rev{for BNN and MCD}, unless specified otherwise. The mean of the iterations is the prediction for the test samples and the standard deviation of the iterations is considered the uncertainty score. 

\rev{Unless stated otherwise, the uncertainty score of a test sample is the standard deviation, across inference passes, of the predicted probability of the class the model predicts. Where a figure reports the score for a specific class, we take the standard deviation of that class's probability instead, and the caption states which. For the filtering experiment in Section~}\ref{sec_4_4:filter}\rev{, the score is computed in probability space: the output activation is applied to each pass before the standard deviation is taken.}

\subsection{Evaluation Metrics}\label{app_a:metrics}
In this section we introduce the metrics we used in our analysis for evaluating the predictive performance and the quality of uncertainty scores. We denote the dataset as $D = \{x_n,y_n\}^{N}_{n=1}$ where $x_n$ is the one-hot-encoded input sequence of length $150$ (i.e. a $150$ by $4$ matrix) and $y_n$ is the one-hot encoded class label, $y \in \{1,...,K\}$, where $K=3$ for real dataset and $K=6$ for simulated dataset. $\theta$ is the model parameters and $p_{\theta}(y|x)$ is the predictive distribution that is produced by iterating inference on the model (\rev{stochastic forward passes}). $\mu_{\theta}$ is the mean of the predictive distribution, used for evaluating predictive performance and $\sigma_{\theta}$ is the standard deviation of this distribution used as the uncertainty score. 

\begin{enumerate}

        
    
    \item Negative Log Likelihood (NLL)

    NLL is a widely-used metric for evaluating predictive uncertainty\rev{. For a test sample $x_n$ with true class $y_n$ it is the negative log of the predictive probability of the true class, where the predictive probability $\bar{p}_\theta$ is the mean over the $T=100$ inference passes ($T=10$ members for ENS)}: 
    
    \begin{align}\label{eq:NLL}
    \begin{aligned}
    \color{blue}\mathrm{NLL} &= -\frac{1}{N}\sum_{n=1}^{N} \log \bar{p}_{\theta}(y_{n}|x_{n}), \qquad \bar{p}_{\theta}(y|x) = \frac{1}{T}\sum_{t=1}^{T} p_{\theta_{t}}(y|x) 
    \end{aligned}
    \end{align}
    \rev{For the multi-label settings, with independent sigmoid outputs, the NLL of a sample is the binary cross-entropy summed over the $K$ classes, $-\sum_{k=1}^{K}\left[y_{nk}\log \bar{p}_{\theta}(k|x_n) + (1-y_{nk})\log\left(1-\bar{p}_{\theta}(k|x_n)\right)\right]$, averaged over samples.}
    \\
    \item Brier Score
    
        The Brier score is a \rev{measure} of quality of the predictive uncertainty. It assigns a numerical score to a predictive distribution $p_{\theta}(y|x)$, rewarding better calibrated predictions with a \rev{lower} score \citep{NIPS2017_9ef2ed4b, proper_score_review, proper_score_2}. \rev{It is} equivalent to the squared error between the predictive probability of a label and one-hot encoding of the correct label in classification. 
        \begin{align}\label{eq:BS}
        \begin{aligned}
        \color{blue}BS &= \frac{1}{N}\sum_{n=1}^{N}\sum_{k=1}^{K} \bigl(t_{nk}-\bar{p}_{\theta}(y=k|x_{n})\bigr)^{2} & \text{with }
        t_{nk} = \begin{cases} 1 & \text{if \ } k = y_{n} \\ 0 & \text{otherwise} \end{cases}
        \end{aligned}
        \end{align}
    
    \item Kendall's Tau Test

    This test measures correspondence between two rankings. Values close to 1 indicate strong agreement, and values close to -1 indicate strong disagreement.\citep{2020SciPy-NMeth} The test provides $p-values$ as well as the tau statistics. 
    
    \item Kolmogorov–Smirnov (KS) Test 
    
    KS test is measure for comparing two samples in terms of goodness of fit\citep{ks_test}. It compares the  underlying distributions of two independent samples that can be of different sizes and it does not make assumptions about the underlying distribution being a normal. Here we used a two-sided KS test with the method set to exact for comparing the distribution statistics. The null hypothesis of this test is that the two distributions are identical and the alternative is that the statistics provided by the test is the absolute difference between the empirical distribution functions of the samples. \citep{2020SciPy-NMeth}
    
    We use this test to compare uncertainty scores over all samples in a test set, in different dataset settings. Meaning sample A is the list of $\sigma_{\theta}$ for test samples in scenario A and is compared against sample B for the same test set but in scenario B. An example is to compare a) distribution of uncertainty scores when a model is trained on balanced training data and is tested on balanced data against, b) the distribution of uncertainty scores when the same model is trained on imbalanced training data but tested on the same balanced test dataset. In such case, we hope that UQ model can show a shift in uncertainty scores of the smaller classes, reflecting on the imbalance of training data, therefore the KS test should reject the null hypothesis and measure the distribution shift in the uncertainty scores. Since this is a two-sided test, a large KS statistic show the distribution of uncertainty score has moved as a result of perturbation, but doesn't say anything about the direction of the move. 
 \end{enumerate}
 Moreover, we calculate the usual performance metrics for classification tasks, namely, accuracy, average precision, area under precision-recall curve (AUPR) and area under receiver operating characteristic curve (AUROC). Accuracy is computed per use case with the metric matching its labeling scheme: argmax agreement for single-label settings and element-wise agreement at a threshold of $0.5$ for multi-label settings. \chg{Average precision is the mean over classes of the one-vs-rest average precision.}
\subsection{Further Experimental Details}\label{app_a:add}
We ran our experiments on an NVIDIA A$40$  GPU node with CentOS Linux $7$ in our internal HPC cluster and are tracked using Neptune AI software \citep{neptune}. Python version is $3.10$ in Anaconda $24.3$. Cuda version is $12.3$ and cuDNN, $8.9$, to be compatible with \rev{TensorFlow $2.16$} and \rev{TensorFlow}-Probability $0.24.0$. For more information about packages and reproducibility please refer to the ReadMe file in the code repository. 

Each training experiment is repeated with $5$ different random seeds. The seed is used to split the training, validation, and test sets, and to fix the random seed in relevant Python libraries. The results presented in this paper are produced with random seeds: $7$, $5$, $6$, $9$ and $11$. When comparing experiments across different models (i.e. ENS, BNN, MCD), we compare results from the dataset splits on the same random seeds, to ensure that the comparison is done for models trained on the same training data and tested on the same test dataset. This is essential for comparing shifts in uncertainty scores or calculating correlations among predictions or among uncertainty scores for different models.  

\section{Additional Results}\label{app_b}

\subsection{XAI and the TF simulation Strategies}
\begin{figure}
     \centering
    \includegraphics[width=\textwidth]{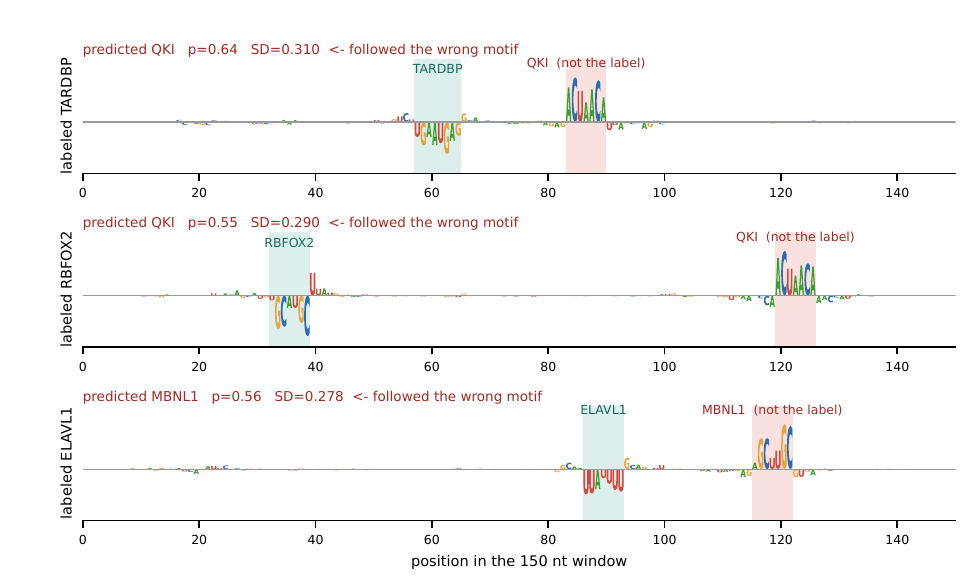}
    \caption{As our simulation strategy was not suitable for generating data for the RBP-binding task, we had several unwanted multi-labeled samples generated in the single-label setting. Having multiple class motifs is the reason for model confusion and higher uncertainty. This issue was resolved when we used the same simulation strategy for the TF-binding task.}
    \label{supp_fig:xai_uq}
\end{figure}

We initially used SimDNA to generate simulated datasets for the same RBP-binding tasks. We generated RNA backgrounds and embedded motifs from RBPs. Then we ran our experiments and inspected the samples that were identified as highly uncertain by BNN, even in the clean single-label balanced setting with no noise. For the most uncertain test samples, we ran the Integrated Gradient (IG)\citep{sundararajan2017axiomatic}, an explainable AI (XAI) method, and looked into the attribution maps produced by IG on BNN. Using XAI together with UQ identified the source of uncertainty in those samples, which was due to a data simulation flaw. The attribution maps revealed that while we planted only a single motif in each input sequence, there was by chance a sequence motif of another protein present in the background (see Figure~\ref{supp_fig:xai_uq}). This caused multiple class motifs to be present in the input which confused the model that was training in single-label setting. The reason for the simulation issue was that SimDNA is designed for synthesizing DNA sequences and implanting TF motifs, where TF motifs ($10$-$15$ nt) are usually longer than RBP motifs (typically $4$-$8$ nt), therefore by chance, the probability of creating a background sequence with an RBP motif is higher (see suppl. Figure~\ref{supp_fig:sample_motifs} for motif examples). This insight was the reason why we switched to using DNA sequences and TFs for the simulated data rather than RNA sequences and RBPs. This observation showed that XAI methods can be utilized for uncertainty-aware models to identify the source of uncertainty. 

\subsection{\chg{The 27-RBP application: extended analysis}}\label{app_b:app27}
\chg{This section collects the additional analyses of the 27-class RBP application introduced in Section~\ref{sec_4_4:filter}: the precision-recall curves behind the filtering result, and three analyses of what drives the uncertainty on this task.}

\subsubsection{\chg{Precision-recall curves for the 27-RBP filtering experiment}}\label{app_b:ap_curves}
\begin{figure}
     \centering
        \begin{subfigure}[b] {0.48\textwidth}
         \centering
         \includegraphics[width=\textwidth]{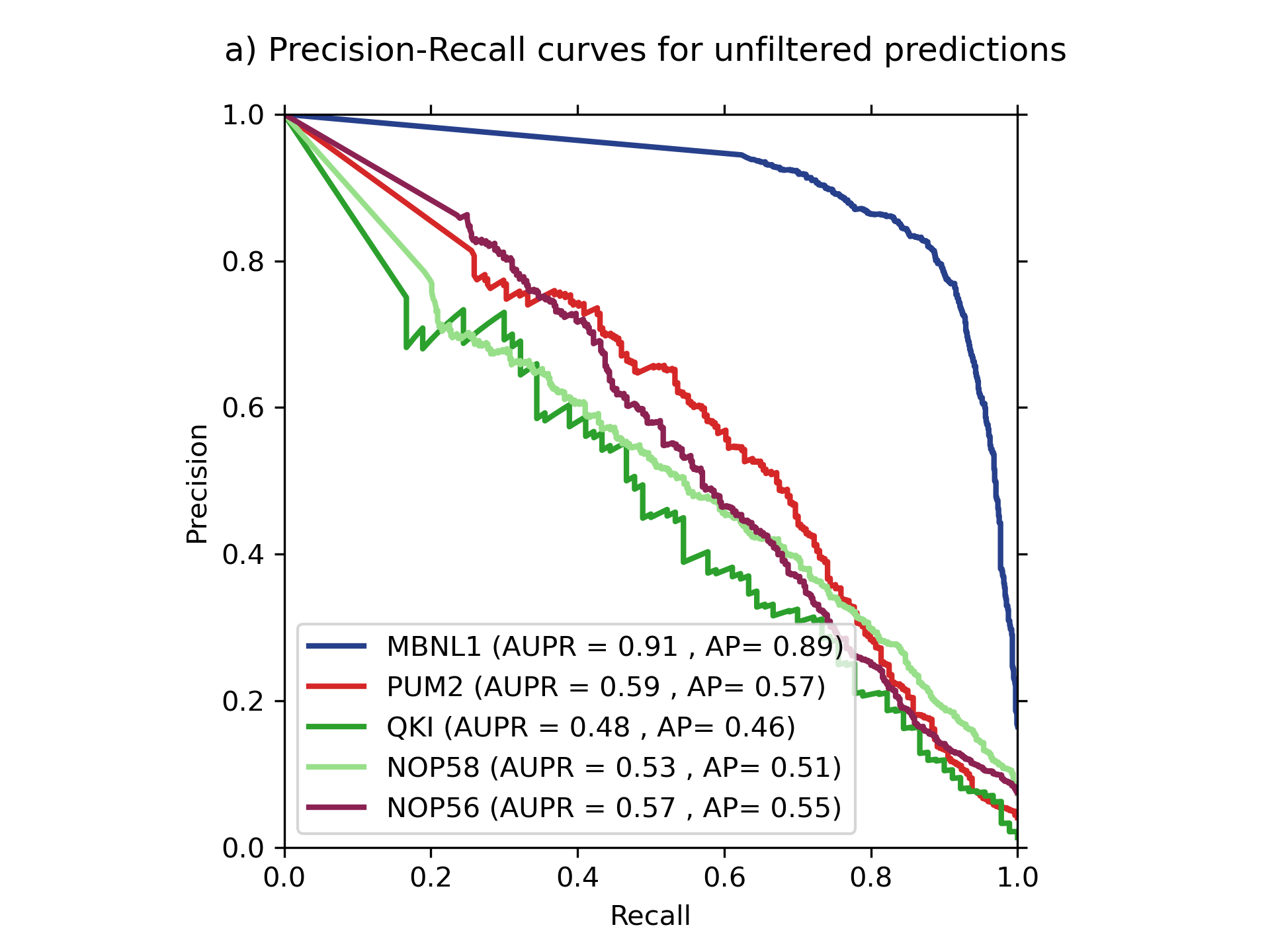}
     \end{subfigure}
     \hfill
     \begin{subfigure}[b]{0.48\textwidth}
         \centering
         \includegraphics[width=\textwidth]{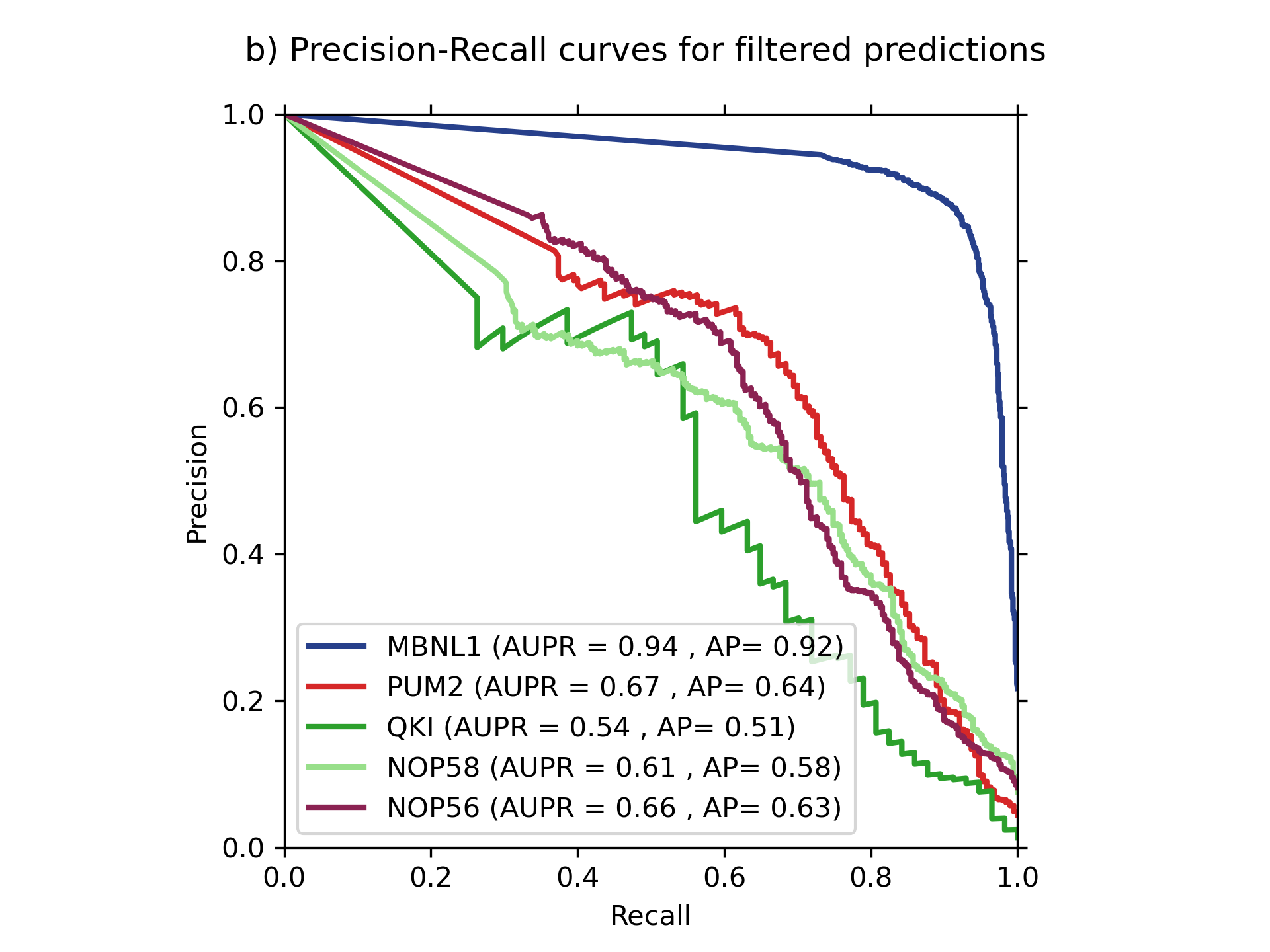}
     \end{subfigure}
     \hfill
        \caption{\textbf{Precision-recall curves before and after uncertainty filtering (27-RBP task).} Precision-recall curves for 5 of the 27 classes for a BNN: a) unfiltered predictions; b) predictions filtered to an uncertainty score below $0.1$ for the predicted class.}
        \label{supp_fig:ap_curves}
\end{figure}

\chg{The main text reports that filtering predictions on an uncertainty score below $0.1$ raises average precision on the 27-class RBP task. Figure~\ref{supp_fig:ap_curves} shows the underlying precision-recall curves for $5$ of the $27$ classes, before and after filtering.}

\subsubsection{\chg{What drives the uncertainty on this task}}\label{app_b:app27_extended}
\chg{The filtering experiment of Section~\ref{sec_4_4:filter} shows that the uncertainty score is useful, but not what drives it. We therefore examined the 27-class predictions further, using integrated gradients~\citep{sundararajan2017axiomatic} on the trained BNN and the structure of the model's errors. Three findings are reported here. The accompanying code repository contains the complete set of analyses, including further tests of additional candidate factors, which showed no other conclusive effect.}

\paragraph{\chg{Incorrect predictions carry no sequence-motif evidence}}
\chg{For each test sequence we computed integrated-gradient attribution maps with respect to the predicted class, averaged over $20$ draws from the BNN posterior, and summarized each map by the fraction of the total attribution falling inside the most informative $8$-nucleotide window. RBP motifs are typically $4$--$8$ nucleotides long~\citep{Gerstberger1}, so this window is matched to the scale of the feature being detected. As a reference for ``no motif present'' we repeated the procedure on dinucleotide-shuffled copies of the same sequences, which preserve base and dinucleotide composition while destroying any motif.}

\chg{Correctly classified sequences carry a clear motif signal: for MBNL1, PUM2 and QKI the attribution is concentrated $+0.101$, $+0.070$ and $+0.026$ above the shuffled reference ($p<10^{-4}$ in all three). Incorrectly classified sequences do not. Their attribution maps are statistically indistinguishable from, or below, the dinucleotide-shuffled reference in all three RBPs, and differ from the correctly classified ones with $p$ between $10^{-22}$ and $10^{-34}$. The model does not attend to a competing motif when it errs; it has no localized sequence evidence at all.}

\begin{figure}
    \centering
    \includegraphics[width=\textwidth]{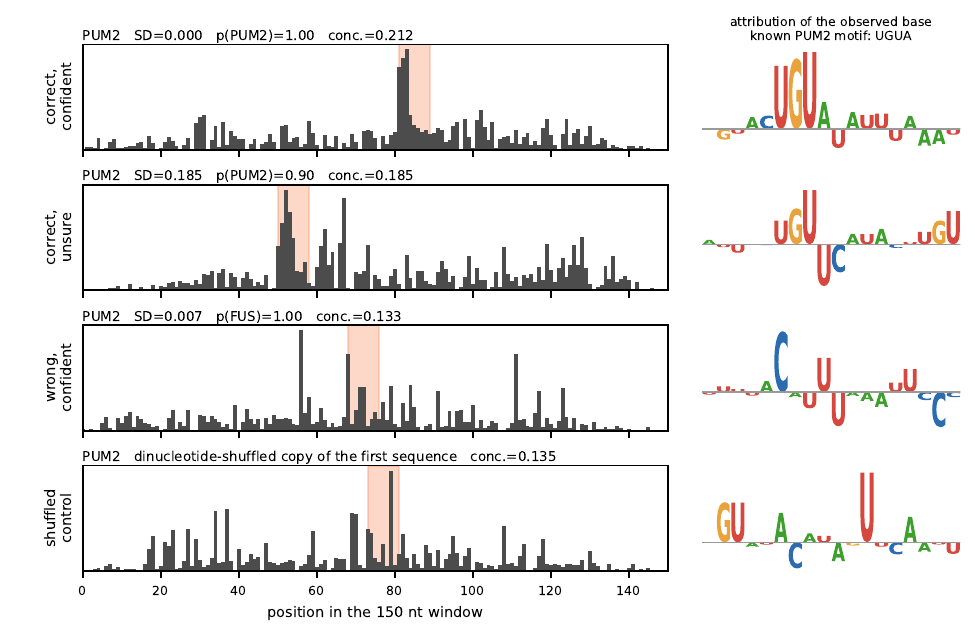}
    \caption{\chg{\textbf{Attribution maps for individual PUM2 sequences.} Rows: a confidently correct prediction, a correct but uncertain one, a confidently incorrect one, and a dinucleotide-shuffled copy of the first sequence. Left: attribution summed over the four bases at each position, with the most informative $8$-nucleotide window shaded. Right: that window as a logo, letter height giving the signed attribution of the base present. The confidently correct sequence carries the UGUA motif; the confidently incorrect one carries no motif and is predicted as FUS. Sequences are the median of each group by attribution concentration.}}
    \label{supp_fig:xai_examples}
\end{figure}

\begin{figure}
    \centering
    \includegraphics[width=0.92\textwidth]{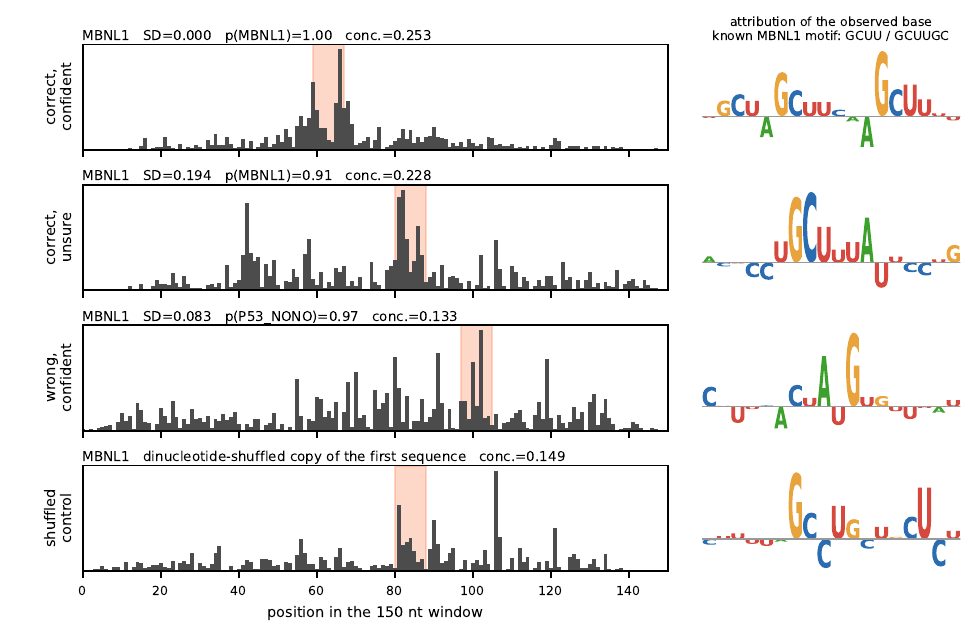}

    \vspace{0.6em}
    \includegraphics[width=0.92\textwidth]{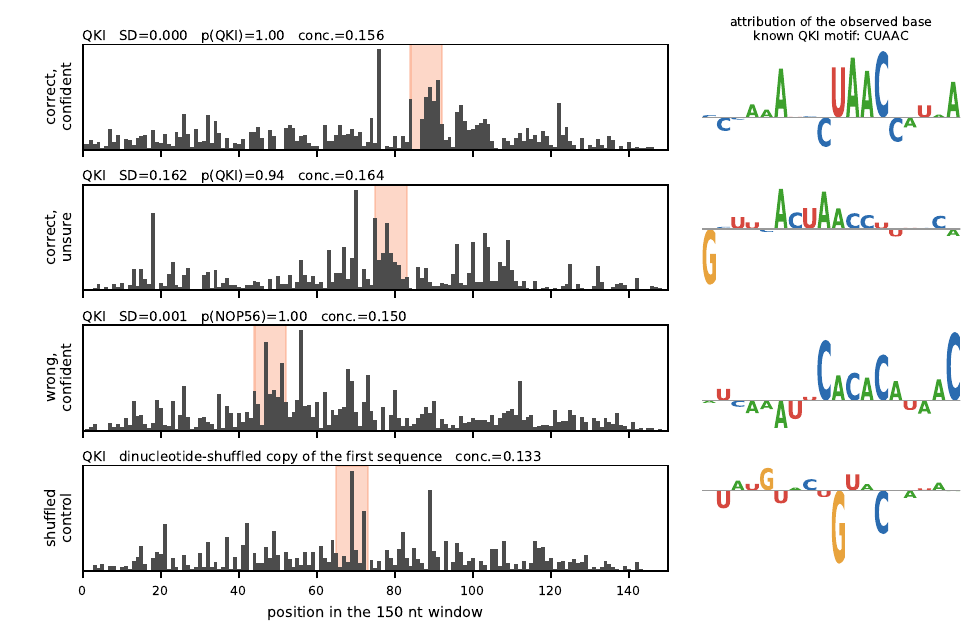}
    \caption{\chg{\textbf{Attribution maps for individual MBNL1 and QKI sequences.} As Figure~\ref{supp_fig:xai_examples}, for the other two RBPs. The confidently correct MBNL1 sequence carries three copies of GCUU; the confidently correct QKI sequence carries UAAC. The confidently incorrect predictions are assigned to P53\_NONO and to NOP56 respectively, the latter being the class QKI is most often confused with (Figure~\ref{supp_fig:complex}).}}
    \label{supp_fig:xai_examples_2}
\end{figure}

\chg{Figures~\ref{supp_fig:xai_examples} and~\ref{supp_fig:xai_examples_2} show this for individual sequences. For each RBP we plot the attribution profile of one confidently correct, one correct but uncertain, and one confidently incorrect prediction, together with a dinucleotide-shuffled copy of the first. Beside each profile the most informative window is drawn as a logo in which letter height is the signed attribution of the base actually present, so the nucleotides the model used can be read directly. In the confidently correct sequences these are the published binding motifs: GCUU repeated three times for MBNL1, UGUA for PUM2 and UAAC for QKI, matching the motif panel in Figure~\ref{supp_fig:sample_motifs}a. In the confidently incorrect sequences no such motif is present and the attribution is spread across the window, at the level of the shuffled control.}

\chg{The sequences shown are the median of each group by attribution concentration rather than the sharpest examples available, so they are representative rather than best-case; their indices are given in the released code repository. The margin over the shuffled control tracks how well each class was learned: it is $0.104$ for MBNL1, $0.070$ for PUM2 and only $0.023$ for QKI, the smallest of the three classes, whose motif is correspondingly the least distinct in the figure.}

\paragraph{\chg{Uncertainty is structured by protein complex membership}}
\chg{The errors themselves are not arbitrary. We measured, for each ordered pair of RBPs, how often the model assigns a sequence of one class to the other, relative to how often that second class is predicted overall. This normalization matters: without it a class that is simply predicted frequently would appear to attract errors specifically.}

\begin{figure}
    \centering
    \includegraphics[width=\textwidth]{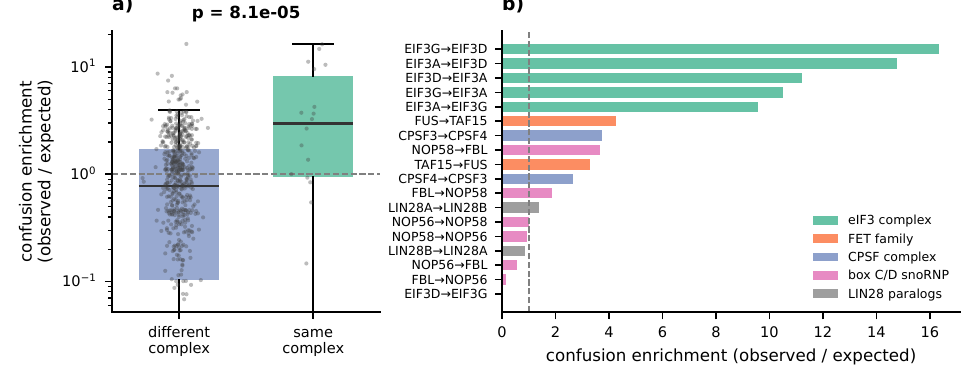}
    \caption{\chg{\textbf{Misclassifications concentrate within protein complexes.} a) Confusion enrichment, defined as the observed rate of predicting class $b$ for sequences of class $a$ divided by the rate expected from how often $b$ is predicted overall. Points are ordered RBP pairs; a value of $1$ (dashed line) is the rate expected by chance. b) The individual pairs within each complex or family. Enrichment is a ratio, so panel a) uses a logarithmic axis.}}
    \label{supp_fig:complex}
\end{figure}

\chg{Pairs of RBPs belonging to the same protein complex or family are confused $2.97$ times more often than expected, against $0.78$ for pairs from different complexes (Mann--Whitney, $p=8\times10^{-5}$; Figure~\ref{supp_fig:complex}). The effect is reciprocal in both directions for the subunits of eIF3 (up to $16$-fold), for the FET-family proteins FUS and TAF15 ($3$--$4$-fold), and for the CPSF subunits ($3$--$4$-fold). These groupings are established independently of our data: the eIF3 subunits A, D and G bind an overlapping program of messenger RNAs~\citep{lee2015eif3}, the FET proteins show closely similar binding profiles in PAR-CLIP~\citep{hoell2011}, and CPSF3 and CPSF4 are subunits of one pre-mRNA $3'$ processing complex~\citep{shi2009}.}

\chg{The interpretation is that the model is least certain where the underlying biology is genuinely ambiguous. A transcript bound by one member of a complex is frequently bound by the others, so the single-label ground truth is partly arbitrary for those sequences, and a well-behaved uncertainty score should be elevated on them. We tested if we can associate high uncertainty score to presence of the wrong/swapped motifs for these misclassifications but couldn't establish that connection. This is expected, since these proteins are general initiation factors that are not known to bind to a specific sequence motif, and will target different messenger RNAs. Therefore XAI on the model doesn't pick any motifs for these protein complexes.}

\paragraph{\chg{Class size drives the uncertainty score}}
\chg{The third factor is not a property of individual sequences but of the classes themselves, and is reported in Section~\ref{sec_4_4:filter}. The effect depends on how the scores are grouped. Grouping by the class the model predicted, which is what the filtering threshold acts on, the Spearman correlation between training-set size and median uncertainty is $\rho=-0.59$ ($p=0.001$, $n=27$) and the five smallest classes have a median $57$ times that of the five largest.
}

\chg{In the controlled experiments we construct the imbalance deliberately; here it is simply a property of how many binding sites each protein has in the PAR-CLIP data, and the BNN's uncertainty tracks it. It is also worth mentioning that the mechanism behind the filtering result of Section~\ref{sec_4_4:filter}: discarding high-uncertainty predictions removes disproportionately from the classes the model had least opportunity to learn, which is why average precision improves most for the rarest RBPs.}

\begin{figure}[h]
\centering
\includegraphics[width=0.6\textwidth]{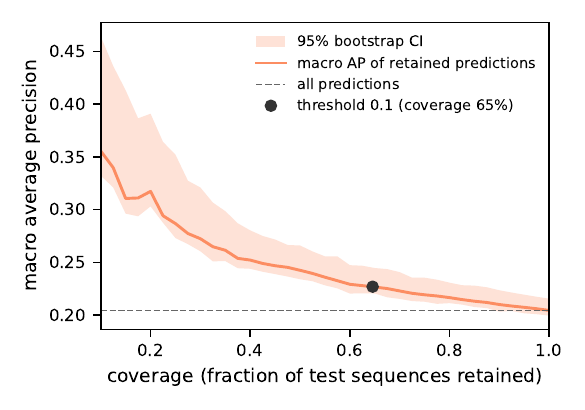}
\caption{\rev{\textbf{Average precision against coverage on the 27-class RBP task (BNN).} Macro average precision of the retained predictions when the test sequences with the lowest uncertainty score are kept, as a function of the fraction kept. Shaded: $95\%$ bootstrap interval over test sequences. Dashed: all predictions. The dot marks the threshold of $0.1$ used in Figure~\ref{fig:27_rbp_class_imba}.}}
\label{supp_fig:ap_coverage}
\end{figure}

\paragraph{\rev{The filter as selective prediction.}}
\rev{Filtering predictions by an uncertainty threshold is selective prediction, and the standard way to show it is the performance of the retained predictions against the fraction retained (coverage) as the threshold is swept. Figure~\ref{supp_fig:ap_coverage} gives the macro average precision over the $27$ classes against coverage for the BNN, with a $95\%$ bootstrap interval over test sequences ($200$ resamples). The curve is monotone: macro AP rises from $0.20$ with every prediction kept to $0.36$ when only the $10\%$ most confident predictions are kept, so any threshold trades coverage for precision and there is no single correct value. The threshold of $0.1$ used in the main text was fixed before the experiment as a round value on the probability scale (a standard deviation of ten percentage points) rather than tuned; it retains $65\%$ of the test sequences and raises macro AP from $0.20$ to $0.22$. A data-driven alternative is the $95$th percentile of the score among correctly classified sequences, $0.29$, which retains more and gains less. In practice the curve, not the threshold, is the object to report, so that the operating point can be chosen for the coverage an application needs.}

\paragraph{\rev{Per-class coverage of the filter.}}
\rev{Filtering by an uncertainty threshold keeps a different fraction of each class, and a filter that discards rare classes first would amplify missed-detection bias for exactly the proteins with the fewest binding sites. Table~\ref{supp_tab:coverage} gives, for each of the 27 RBPs, the number of positive test sequences, the fraction retained by the threshold of $0.1$ on the predicted class, and the average precision before and after filtering. The filter retains 65\% of all positive sequences; per class the retained fraction ranges from 50\% to 85\%, with a mean of 57\% for the five smallest training classes against 68\% for the five largest. No class is removed entirely, and the rank correlation between training-set size and retained fraction is $\rho = +0.31$ ($p = 0.11$, $n = 27$). The filter therefore does discard proportionally more of the rare classes, but mildly: the difference between the extreme groups is 11 percentage points of coverage. In applications where missed detections in rare classes are costly, we recommend setting the threshold per class, or reporting the retained fraction per class next to any aggregate gain, as done here.}

\begin{table}[h]
\centering\small
\caption{\rev{Per-class coverage of the uncertainty filter on the 27-class RBP task (BNN, seed 9). Classes are ordered by training-set size. Coverage is the fraction of positive test sequences whose uncertainty score at the predicted class is below $0.1$; AP is the average precision before and after filtering.}}
\label{supp_tab:coverage}
\begin{tabular}{lrrrrr}
\toprule
RBP & train & test positives & coverage & AP all & AP filtered \\
\midrule
MBNL1 & 7354 & 1119 & 85\% & 0.89 & 0.92 \\
FUS & 3740 & 540 & 60\% & 0.30 & 0.32 \\
NOP56 & 3629 & 505 & 70\% & 0.55 & 0.63 \\
SSB & 3608 & 490 & 60\% & 0.33 & 0.37 \\
NOP58 & 3446 & 497 & 67\% & 0.51 & 0.58 \\
AGO3 & 3129 & 432 & 61\% & 0.24 & 0.28 \\
FBL & 2601 & 348 & 57\% & 0.21 & 0.21 \\
XPO5 & 2010 & 275 & 60\% & 0.10 & 0.11 \\
LIN28B & 1965 & 270 & 57\% & 0.18 & 0.19 \\
DICER1 & 1875 & 272 & 54\% & 0.10 & 0.10 \\
TAF15 & 1780 & 279 & 56\% & 0.13 & 0.13 \\
PUM2 & 1734 & 274 & 68\% & 0.57 & 0.64 \\
P53\_NONO & 1724 & 241 & 61\% & 0.18 & 0.19 \\
CPSF3 & 1364 & 177 & 54\% & 0.05 & 0.05 \\
LIN28A & 1229 & 187 & 51\% & 0.06 & 0.05 \\
FXR1 & 1025 & 141 & 56\% & 0.09 & 0.09 \\
CPSF4 & 942 & 124 & 60\% & 0.03 & 0.03 \\
UPF1 & 729 & 108 & 61\% & 0.03 & 0.03 \\
ZFP36 & 709 & 118 & 54\% & 0.21 & 0.25 \\
QKI & 703 & 90 & 63\% & 0.46 & 0.50 \\
G45 & 495 & 78 & 60\% & 0.05 & 0.05 \\
G35 & 363 & 54 & 61\% & 0.01 & 0.01 \\
EIF3D & 360 & 31 & 52\% & 0.05 & 0.07 \\
RTCB & 352 & 44 & 52\% & 0.02 & 0.02 \\
EIF3A & 314 & 37 & 73\% & 0.03 & 0.03 \\
EIF3G & 194 & 30 & 60\% & 0.05 & 0.09 \\
PAPD5 & 148 & 24 & 50\% & 0.01 & 0.02 \\
\bottomrule
\end{tabular}
\end{table}

\subsection{\chg{Missing-motif noise on the simulated TF data}}\label{app_b:tf_no_motif}
\begin{figure}
     \centering
     \includegraphics[width=\textwidth]{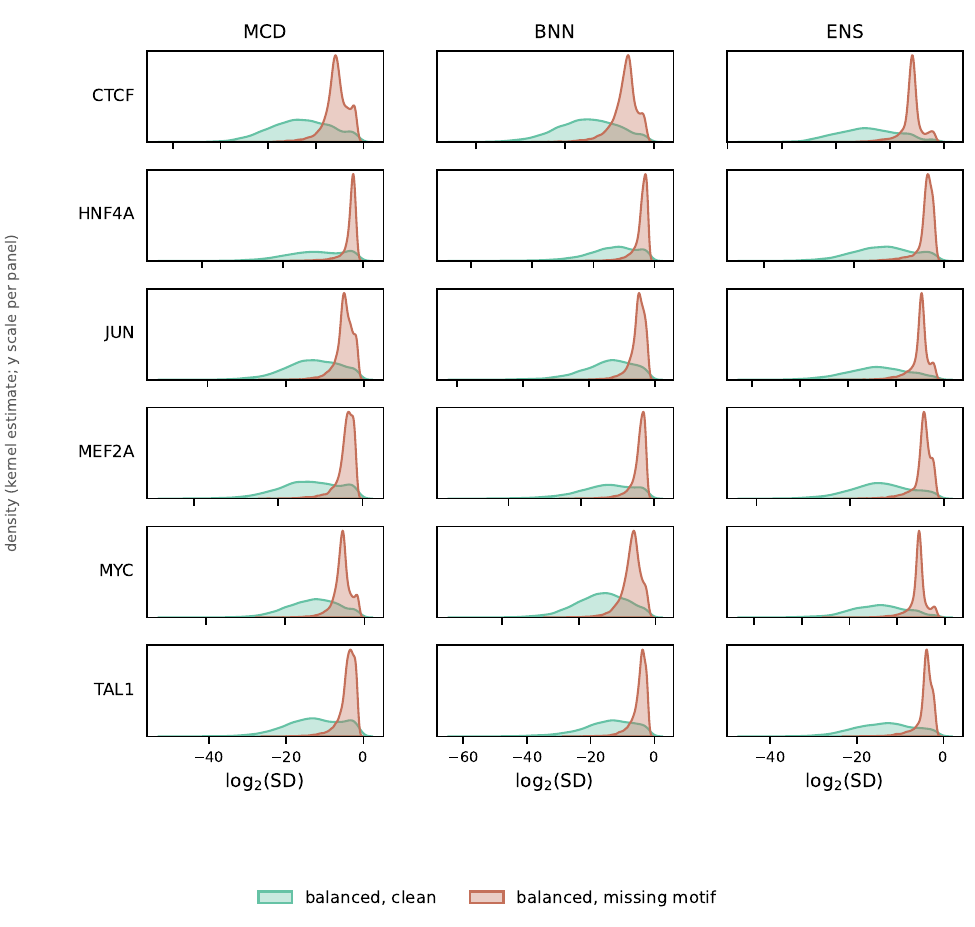}
     \caption{\chg{\textbf{Sensitivity of uncertainty scores to missing motifs (simulated TF data).} Distribution of uncertainty scores when the model is trained and tested on clean balanced single-label data (green) versus background sequences with no planted motif (orange). Rows are the six transcription factors; columns are MCD, BNN, and ENS. Scores are on a $\log_2$ scale\rev{, pooled over the five random seeds}.}}
     \label{supp_fig:tf_shift_no_motif}
\end{figure}

\chg{The main text reports the mislabeling perturbation, where the three UQ methods differ in how specifically the uncertainty increase is confined to the corrupted classes. The second perturbation we applied to the simulated TF data replaces the input with background sequences carrying no planted motif, removing the predictive signal altogether. Figure~\ref{supp_fig:tf_shift_no_motif} shows the result. Every class shifts by a large and roughly equal amount, \rev{between $7$ and $10$ $\log_2$ units}, and this is true for all three methods, so the experiment does not separate them. We report it as a positive control: it confirms that the shift measurement responds as expected when the signal is unambiguously destroyed, which is what makes the more selective response to mislabeling in the main text interpretable.}

\subsection{\chg{Uncertainty scores on the raw scale}}\label{app_b:tf_raw}
\begin{figure}
     \centering
     \includegraphics[width=\textwidth,height=0.4\textheight,keepaspectratio]{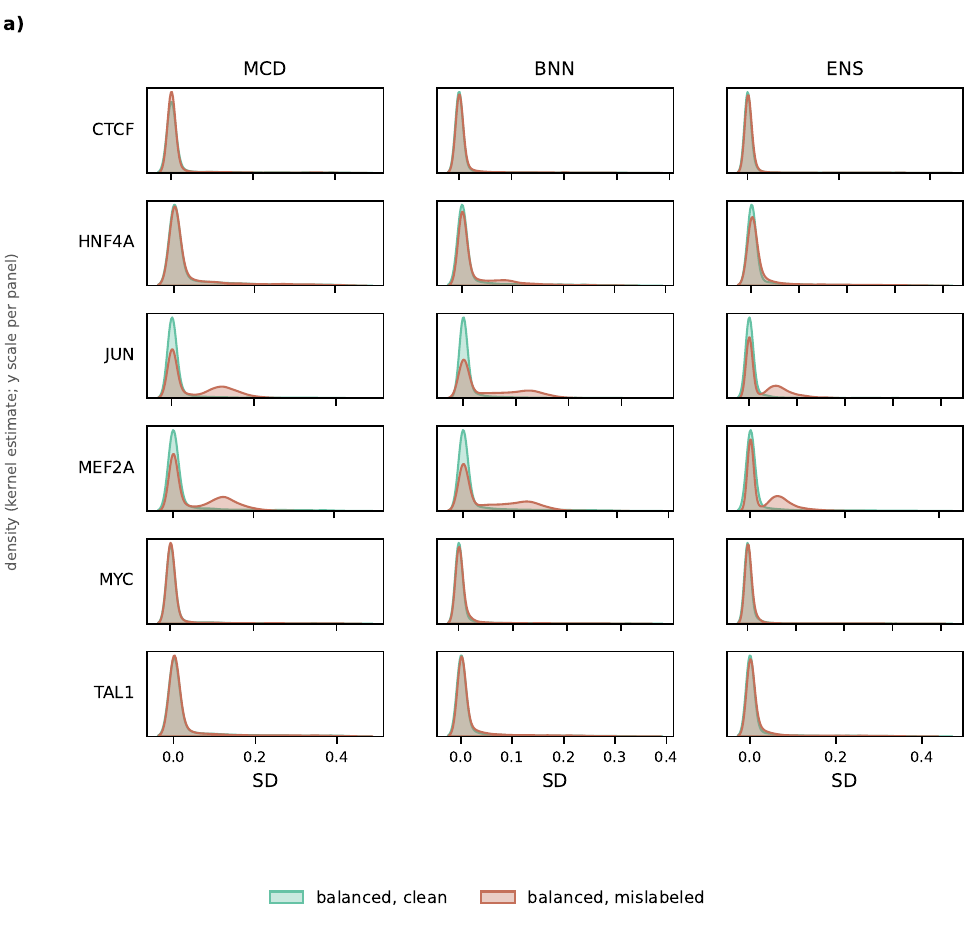}

     \vspace{0.5em}
     \includegraphics[width=\textwidth,height=0.4\textheight,keepaspectratio]{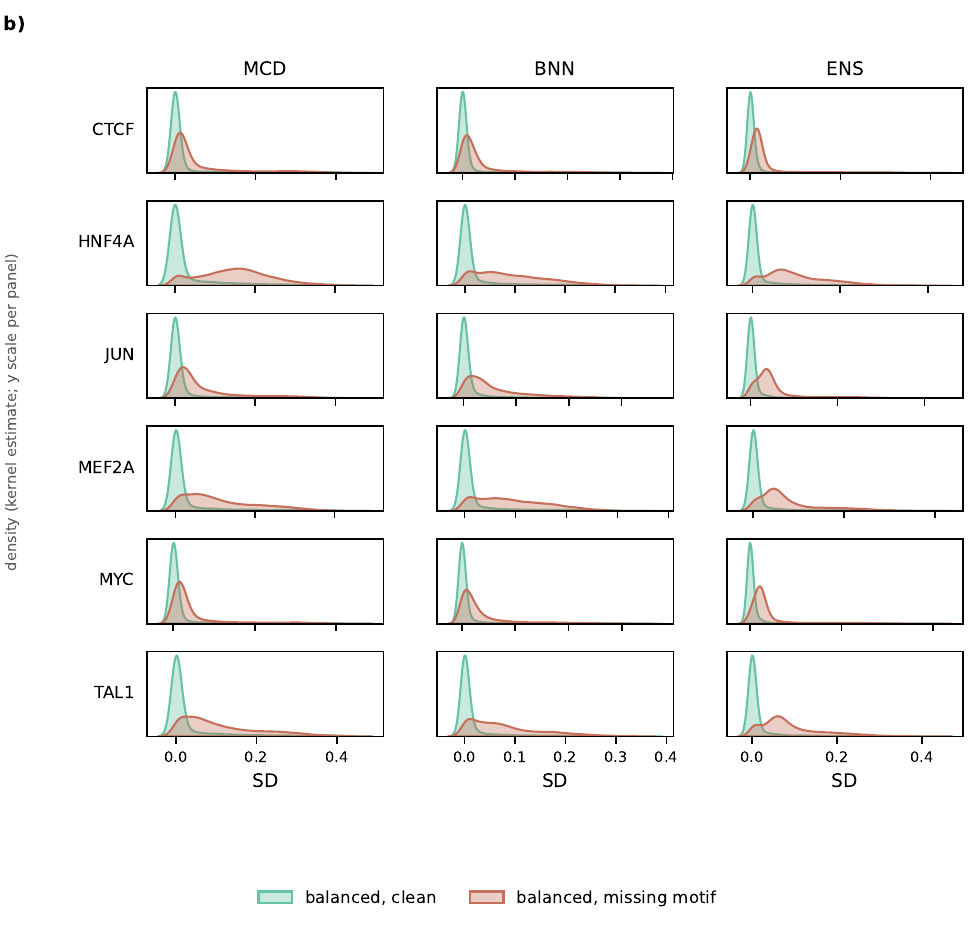}
     \caption{\chg{\textbf{Sensitivity of uncertainty scores on the raw scale (simulated TF data).} Predictive standard deviation on a linear axis, for a) partially mislabeled noise and b) missing-motif noise\rev{, pooled over the five seeds. Green is the model trained on clean data, orange the model trained on the perturbed data}.}}
     \label{supp_fig:tf_shift_raw}
\end{figure}

\begin{figure}
     \centering
     \includegraphics[width=\textwidth]{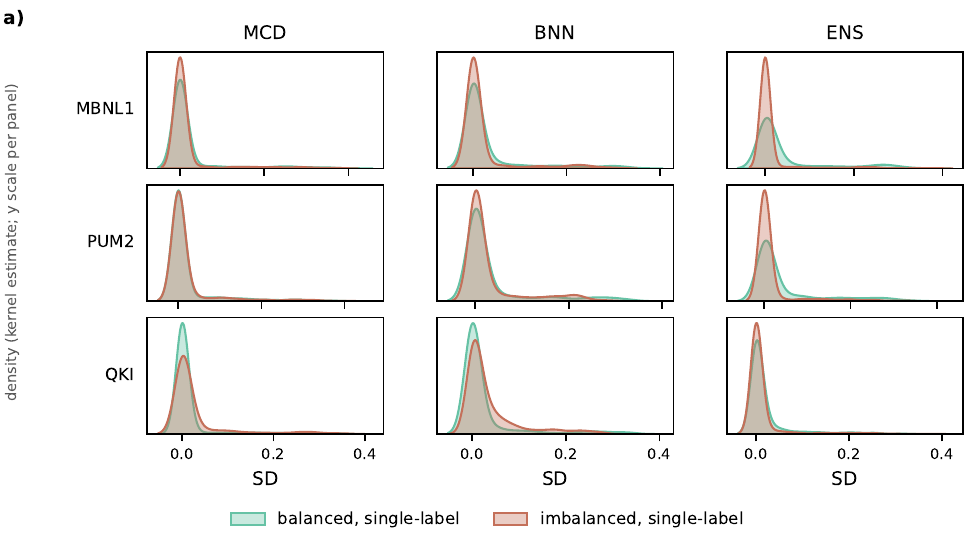}

     \vspace{0.5em}
     \includegraphics[width=\textwidth]{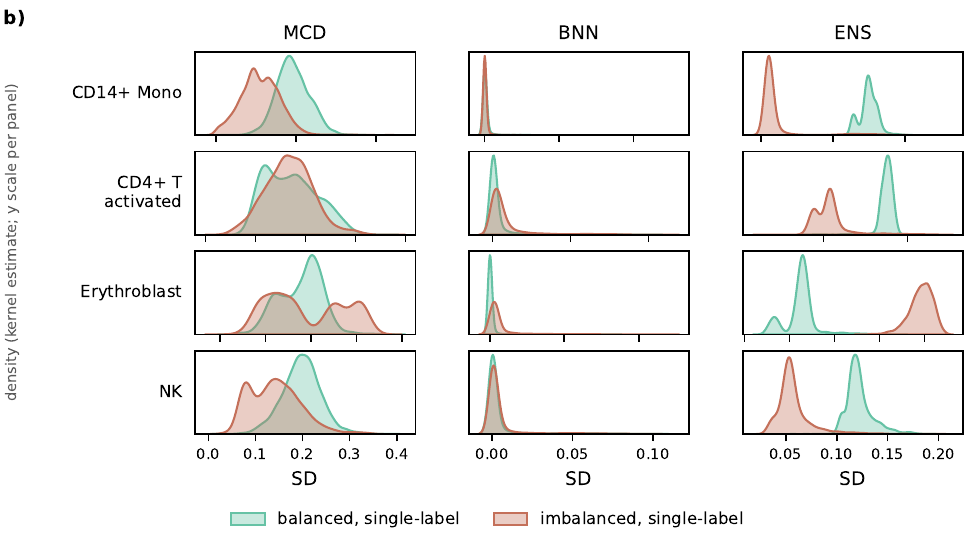}
     \caption{\chg{\textbf{Sensitivity of uncertainty scores to class imbalance on the raw scale.} Predictive standard deviation on a linear axis, for a) the real RBP use case and b) the single-cell use case. Green: balanced training set; orange: highly imbalanced training set. \rev{Score at the predicted class, pooled over the five seeds.}}}
     \label{supp_fig:sd_shift_raw_real}
\end{figure}

\chg{The sensitivity figures in the main text plot uncertainty scores on a $\log_2$ scale. Figures~\ref{supp_fig:tf_shift_raw} and~\ref{supp_fig:sd_shift_raw_real} give the same data on the raw scale, for reference. They illustrate why the log scale is necessary rather than cosmetic: \rev{most predictive standard deviations lie below $10^{-2}$ while a few reach $0.4$, so on a linear axis every class in every panel collapses to a spike at the origin and neither the size nor the direction of the shift can be read off.}}

\subsection{\chg{Uncertainty shift in class imbalance}}\label{app_b:depletion}
\begin{figure}
    \centering
    \includegraphics[width=\textwidth]{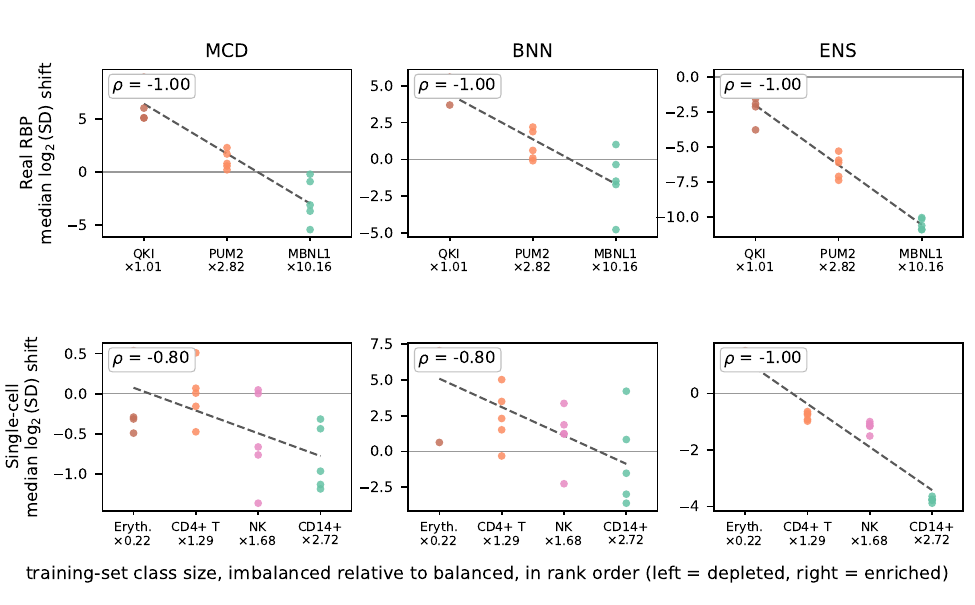}
    \caption{\chg{\textbf{Change in training-set class size against the shift in uncertainty.} Each point is one class in one of $5$ random-seed repeats. Classes are ordered by how their training-set size changed when imbalance was introduced, given below each label as a multiplier; left-most is the most depleted. Top row: real RBP use case, evaluated on a class-balanced test set. Bottom row: single-cell use case; abbreviated cell types are Erythroblast, CD4+ T activated, NK and CD14+ Mono. The dashed line is a least-squares fit and $\rho$ is the median Spearman correlation across seeds. A negative $\rho$ means the classes that lost training data became the more uncertain ones. \rev{The score is the SD at the predicted class.}}}
    \label{supp_fig:depletion}
\end{figure}

\chg{The sensitivity figures in the main text show\rev{, pooled over the five random seeds,} that the distribution of uncertainty scores moves when the training set is made imbalanced.
Figure~\ref{supp_fig:depletion} \rev{shows} the change in training-set size against the median $\log_2$(SD) shift for each class separately, with one point per random seed. There are two different questions to be asked about the class imbalance: does the model produce higher uncertainty scores overall for all classes when imbalanced is introduced to data? and does the model produce higher uncertainty score for the smallest class compared to the other classes in the imbalanced setting? Answering the first question will be affected by how we built the balance/imbalanced versions of the datasets for our experiments. For commenting on usability of UQ methods in working with highly imbalanced genomics datasets, we care about the answer to the second question.}

\chg{The two use cases construct imbalance differently: in the single-cell data the minority class is depleted from the balanced version: Erythroblast falls from $2673$ to $587$ training cells, a reduction of $78\%$, while the other three classes grow. In the RBP data, the imbalanced version is the real starting point and the smallest class keeps essentially all of its data (QKI, $816$ to $825$ sequences) and the balanced version is a subset of this data. Only the single-cell panels therefore test whether the harshly sub-sampled class becomes the most uncertain one; the RBP panels test the complementary manipulation.
}

\rev{The score is the SD at the predicted class, as in Figure~\ref{fig:sd_shift_imbalance}. On the single-cell data all three methods give a negative Spearman correlation between the change in class size and the shift in uncertainty (median over seeds $-0.8$ for BNN, $-1.0$ for ENS, $-0.8$ for MCD), so the depleted class is the one that becomes relatively most uncertain for every method. The methods differ in the size of that shift: Erythroblast moves by $+5.6$ $\log_2$ units for BNN in every seed (Table~\ref{supp_tab:entropy_mi}), by $+1.2$ for ENS, and not at all for MCD, whose four classes all stay within $\pm0.8$. On the RBP data, where QKI is the class that received no additional data, the correlation is $-1.0$ for all three methods, with QKI moving by $+4.6$ for BNN and $+6.7$ for MCD (each setting's own predicted class, Table~\ref{supp_tab:entropy_mi}), and ENS becoming more confident in every class.}

\subsection{\rev{MC-dropout configuration}}\label{app_b:mcd_zero}
\rev{MC-dropout's uncertainty score depends on which drop-out layers are active at inference and at what rate. All MCD results in this paper use every drop-out layer of the trained network at its training rate (Section~\ref{app_a:impl}). Table~\ref{supp_tab:mcd_config} compares this with three other configurations on each use case, using the same trained models and $100$ passes each: on the sequence tasks, dropout in the last layer only and a rate of $0.5$; on the single-cell classifier, whose two layers are trained at $0.3$, the last layer only and a rate of $0.15$. The configuration changes the scale of the score and the fraction of samples with no score at all, not the accuracy of the mean prediction. On the sequence data a single active layer gives a near-degenerate score (median SD $2\times10^{-5}$ on clean TF data with $10\%$ of test samples at exactly zero, $3\times10^{-6}$ on balanced RBP data); all layers raise it by one to two orders of magnitude, and a rate of $0.5$ in all layers to $0.16$, where the model is uncertain about every sequence. The shift of the perturbed classes is present in every configuration. On the single-cell data the score stays large under every configuration, between $0.02$ and $0.18$ at an accuracy of $0.98$--$0.99$, and no configuration makes Erythroblast, the depleted class, more uncertain in a consistent direction (for these rows single seeds dominate the mean shift, and the sign count is the informative quantity).}

\begin{table}[h]
\centering\small
\caption{\rev{MC-dropout under four inference configurations per use case, same trained models. Accuracy of the mean prediction, median SD at the predicted class (mean over five seeds) and the fraction of test samples with an SD of exactly zero, in the reference (ref.) and the perturbed (pert.) setting; last column, the shift of the perturbed class from the reference to the perturbed setting (median $\log_2$ SD at the predicted class, mean over seeds, seeds with a positive shift).}}
\label{supp_tab:mcd_config}
\begin{tabular}{lcccccc}
\toprule
dropout at inference & \multicolumn{2}{c}{accuracy} & \multicolumn{2}{c}{median SD at predicted class} & zero SD & shift of the perturbed class \\
 & ref. & pert. & ref. & pert. & ref. / pert. & $\log_2$ (seeds up) \\
\midrule
\multicolumn{7}{l}{\textit{simulated TF, clean $\to$ mislabeled (JUN, MEF2A)}} \\
last layer, 0.25 & $0.950$ & $0.791$ & $1.8\times10^{-5}$ & $6.8\times10^{-4}$ & $9.6\%$ / $5.0\%$ & $+11.5$ (5/5) \\
all layers, 0.25 (used) & $0.949$ & $0.789$ & $3.5\times10^{-3}$ & $0.09$ & $1.0\%$ / $0.3\%$ & $+5.2$ (5/5) \\
last layer, 0.5 & $0.950$ & $0.790$ & $1.8\times10^{-4}$ & $3.2\times10^{-3}$ & $2.0\%$ / $0.8\%$ & $+9.4$ (5/5) \\
all layers, 0.5 & $0.944$ & $0.787$ & $0.16$ & $0.22$ & $0.0\%$ / $0.0\%$ & $+0.6$ (5/5) \\
\midrule
\multicolumn{7}{l}{\textit{real RBP, balanced $\to$ imbalanced (QKI)}} \\
last layer, 0.25 & $0.973$ & $0.969$ & $2.9\times10^{-6}$ & $1.1\times10^{-5}$ & $10.3\%$ / $12.3\%$ & $+6.1$ (5/5) \\
all layers, 0.25 (used) & $0.973$ & $0.969$ & $1.5\times10^{-5}$ & $6.9\times10^{-5}$ & $6.9\%$ / $5.7\%$ & $+6.7$ (5/5) \\
last layer, 0.5 & $0.973$ & $0.969$ & $2.9\times10^{-5}$ & $1.1\times10^{-4}$ & $2.1\%$ / $3.3\%$ & $+7.0$ (5/5) \\
all layers, 0.5 & $0.970$ & $0.967$ & $1.9\times10^{-3}$ & $3.5\times10^{-3}$ & $1.0\%$ / $0.8\%$ & $+5.0$ (5/5) \\
\midrule
\multicolumn{7}{l}{\textit{single-cell, balanced $\to$ imbalanced (Erythroblast)}} \\
last layer, 0.3 & $0.990$ & $0.984$ & $0.12$ & $0.09$ & $0.0\%$ / $0.0\%$ & $-3.8$ (3/5) \\
both layers, 0.3 (used) & $0.990$ & $0.983$ & $0.18$ & $0.15$ & $0.0\%$ / $0.0\%$ & $-0.0$ (2/5) \\
last layer, 0.15 & $0.990$ & $0.984$ & $0.05$ & $0.02$ & $0.0\%$ / $0.0\%$ & $-2.7$ (4/5) \\
both layers, 0.15 & $0.990$ & $0.984$ & $0.08$ & $0.05$ & $0.0\%$ / $0.0\%$ & $+0.7$ (4/5) \\
\bottomrule
\end{tabular}
\end{table}

\subsection{\rev{Shift of the uncertainty scores across random seeds}}\label{app_b:seed_shift}
\rev{The sensitivity figures in the main text pool the five random seeds. Tables~\ref{supp_tab:seed_shift} and~\ref{supp_tab:paired} give the shift for all five seeds. For each class and seed we take the median $\log_2$(SD) of that class's probability over all test samples with non-zero SD, in the reference and in the perturbed setting, and report the difference, averaged over seeds with its standard deviation; the number of seeds in which the shift is positive is given in parentheses. With five seeds a sign test cannot reach $p<0.05$ (its smallest attainable two-sided $p$ is $0.0625$), so we report the consistency of the sign rather than a $p$-value. For the RBP data the table gives the shift both over all test samples and restricted to the samples the reference model predicts as the class, as a paired per-sample comparison on the shared 574-sequence test set.}

\begin{table}[h]
\centering\small
\caption{\rev{Median $\log_2$(SD) shift per class from the reference to the perturbed setting, mean $\pm$ sd over five seeds (seeds with a positive shift). Top: simulated TF, clean to mislabeled, over all test samples; JUN and MEF2A are the swapped classes. Middle: single-cell, balanced to imbalanced, over all test samples; Erythroblast is the depleted class. Bottom: real RBP, balanced to imbalanced on the shared balanced test set, over all test samples (difference of the class medians) and over the samples the reference model predicts as the class (median of the paired per-sample differences).}}
\label{supp_tab:seed_shift}
\begin{tabular}{lccc}
\toprule
 & MCD & BNN & ENS \\
\midrule
\multicolumn{4}{l}{\textit{Simulated TF, label noise}} \\
JUN & $+6.4 \pm 1.1$ (5/5) & $+6.8 \pm 2.1$ (5/5) & $+4.9 \pm 0.5$ (5/5) \\
MEF2A & $+6.1 \pm 1.6$ (5/5) & $+6.2 \pm 1.9$ (5/5) & $+4.7 \pm 0.3$ (5/5) \\
CTCF & $+1.5 \pm 0.6$ (5/5) & $+3.2 \pm 2.7$ (5/5) & $+3.7 \pm 0.3$ (5/5) \\
HNF4A & $+1.8 \pm 1.1$ (5/5) & $+3.7 \pm 3.4$ (5/5) & $+3.7 \pm 0.3$ (5/5) \\
MYC & $+1.5 \pm 0.8$ (5/5) & $+2.4 \pm 3.6$ (4/5) & $+3.3 \pm 0.3$ (5/5) \\
TAL1 & $+2.0 \pm 0.7$ (5/5) & $+3.4 \pm 1.4$ (5/5) & $+3.7 \pm 0.4$ (5/5) \\
\midrule
\multicolumn{4}{l}{\textit{Single-cell, class imbalance}} \\
Erythroblast & $-0.4 \pm 0.6$ (1/5) & $+1.5 \pm 2.4$ (4/5) & $-1.6 \pm 0.1$ (0/5) \\
CD4$+$ T activated & $-0.7 \pm 0.6$ (1/5) & $+3.6 \pm 1.1$ (5/5) & $+0.1 \pm 0.2$ (4/5) \\
CD14$+$ Mono & $-0.0 \pm 0.4$ (1/5) & $+0.8 \pm 3.8$ (2/5) & $-0.8 \pm 0.2$ (0/5) \\
NK & $-0.2 \pm 0.5$ (3/5) & $+1.6 \pm 2.5$ (4/5) & $-0.0 \pm 0.1$ (3/5) \\
\midrule
\multicolumn{4}{l}{\textit{Real RBP, class imbalance, shared test set, all samples}} \\
MBNL1 & $+2.8 \pm 0.6$ (5/5) & $+2.8 \pm 1.6$ (5/5) & $-5.2 \pm 0.7$ (0/5) \\
PUM2 & $+2.4 \pm 0.9$ (5/5) & $+2.1 \pm 1.4$ (4/5) & $-5.6 \pm 0.3$ (0/5) \\
QKI & $+1.0 \pm 0.8$ (4/5) & $+1.1 \pm 1.4$ (3/5) & $-6.1 \pm 0.6$ (0/5) \\
\midrule
\multicolumn{4}{l}{\textit{Real RBP, class imbalance, shared test set, predicted class}} \\
MBNL1 & $-3.9 \pm 2.1$ (0/5) & $-1.4 \pm 2.0$ (1/5) & $-13.6 \pm 0.9$ (0/5) \\
PUM2 & $+0.8 \pm 1.1$ (3/5) & $+1.4 \pm 1.0$ (4/5) & $-7.0 \pm 0.7$ (0/5) \\
QKI & $+8.2 \pm 1.6$ (5/5) & $+4.5 \pm 0.8$ (5/5) & $-1.8 \pm 0.9$ (0/5) \\
\bottomrule
\end{tabular}
\end{table}

\begin{table}[h]
\centering\small
\caption{\rev{Paired comparison on the settings that share a test set. For each test sample, the $\log_2$(SD) of the class's probability is compared between the reference and the perturbed setting. Cells give the median paired difference averaged over seeds, the percentage of samples whose uncertainty increased, and the number of seeds in which a Wilcoxon signed-rank test on the paired differences rejects the null at $p<0.05$. With 1,800 and 3,600 paired samples the test is sensitive to small shifts, so the direction and the percentage are the informative quantities. The median paired difference can differ in sign from the difference of medians in Table~\ref{supp_tab:seed_shift} when the shift is small.}}
\label{supp_tab:paired}
\begin{tabular}{lccc}
\toprule
 & MCD & BNN & ENS \\
\midrule
\multicolumn{4}{l}{\textit{Simulated TF, clean to mislabeled (1,800 paired samples)}} \\
JUN & $+3.4$ / 77\% (5/5) & $+4.9$ / 80\% (5/5) & $+2.1$ / 83\% (5/5) \\
MEF2A & $+2.9$ / 72\% (5/5) & $+3.9$ / 75\% (5/5) & $+1.6$ / 77\% (5/5) \\
CTCF & $+0.7$ / 58\% (4/5) & $+2.5$ / 67\% (5/5) & $+0.5$ / 63\% (5/5) \\
HNF4A & $+0.9$ / 60\% (4/5) & $+3.0$ / 69\% (5/5) & $+0.6$ / 63\% (5/5) \\
MYC & $+0.7$ / 58\% (4/5) & $+1.9$ / 61\% (5/5) & $+0.4$ / 59\% (5/5) \\
TAL1 & $+0.9$ / 60\% (4/5) & $+2.7$ / 70\% (5/5) & $+0.5$ / 61\% (5/5) \\
\midrule
\multicolumn{4}{l}{\textit{Single-cell, balanced to imbalanced (3,600 paired samples)}} \\
Erythroblast & $-0.4$ / 37\% (5/5) & $+1.9$ / 56\% (5/5) & $-1.8$ / 24\% (5/5) \\
CD4$+$ T activated & $-0.5$ / 32\% (5/5) & $+2.6$ / 76\% (5/5) & $-0.4$ / 30\% (5/5) \\
CD14$+$ Mono & $-0.3$ / 41\% (5/5) & $+0.5$ / 61\% (5/5) & $-0.9$ / 25\% (5/5) \\
NK & $-0.2$ / 44\% (5/5) & $+1.5$ / 65\% (5/5) & $-0.5$ / 29\% (5/5) \\
\bottomrule
\end{tabular}
\end{table}

\subsection{\rev{Number of ensemble members and inference passes}}\label{app_b:ens_size}
\rev{The ensemble provides $10$ independent predictions per sample, against $100$ correlated inference passes for BNN and MCD. To see how many members or passes the uncertainty ranking needs, we subsampled the raw per-pass predictions of the RBP models (five seeds, balanced and imbalanced single-label settings), recomputed the standard deviation of the predicted class from $k$ members or passes, and measured Kendall's tau between that ranking and the one obtained from the full set. Table~\ref{supp_tab:ens_size} gives the mean over settings and seeds, and in parentheses the minimum. \rev{With five members the ENS ranking is at $\tau = 0.90$ against the full ensemble; BNN and MCD are at $\tau = 0.81$ and $0.78$ with five passes and at $0.93$ and $0.91$ with $50$.} The standard deviation over the ten ENS members is identical to that over the $100$ stored rows, so the correction of the ENS description in Section~\ref{app_a:impl} changes no reported number.}

\begin{table}[h]
\centering\small
\caption{\rev{Kendall's tau between the uncertainty ranking from $k$ ensemble members (ENS) or $k$ inference passes (BNN, MCD) and the ranking from the full set; mean over five seeds and two RBP settings, minimum in parentheses.}}
\label{supp_tab:ens_size}
\begin{tabular}{lccccc}
\toprule
 & $k=2$ & $k=3$ & $k=5$ & $k=10$ & \\
\midrule
ENS (members) & 0.77 (0.70) & 0.83 (0.76) & 0.90 (0.84) & 1.00 (1.00) & \\
\midrule
 & $k=5$ & $k=10$ & $k=20$ & $k=50$ & $k=100$ \\
\midrule
BNN (passes) & 0.81 (0.78) & 0.85 (0.82) & 0.88 (0.85) & 0.93 (0.91) & 1.00 (1.00) \\
MCD (passes) & 0.78 (0.76) & 0.81 (0.79) & 0.85 (0.83) & 0.91 (0.90) & 1.00 (1.00) \\
\bottomrule
\end{tabular}
\end{table}

\subsection{\rev{Evaluation per donor and per site (single-cell data)}}\label{app_b:donor}
\rev{The train, validation and test splits of the single-cell data are random over cells, so cells from the same donor and site appear on both sides of a split. The embedding is computed by Liam, which is trained to remove the site and donor batch effect while retaining biological variation~\citep{liam}, and the comparison of UQ methods is conditional on that embedding and on how well it preserves the geometry and topology of the data~\citep{zang2022dlme,li2021invml}. To check whether the random split inflates the reported accuracies, we traced every test cell back to its sample of origin (Table~\ref{supp_tab:donor_counts}; the 28,214 cells of the four selected types come from 13 samples, 10 donors at 4 sites) and recomputed the accuracy of each method per sample (Table~\ref{supp_tab:donor_acc}). Accuracy is uniform across samples: between $0.95$ and $1.00$ for every method on the balanced data, and $0.93$--$1.00$ (BNN), $0.92$--$0.99$ (ENS) and $0.94$--$1.00$ (MCD) on the imbalanced data, with no sample or site standing out. The stricter test is to hold a whole sample out of training. Table~\ref{supp_tab:ldo} gives the accuracy of each method on five held-out samples, one per fold and covering all four sites, with the models trained on the other twelve samples (class-balanced, $4{,}100$--$4{,}500$ cells per type). Accuracy on a never-seen donor is $0.977$--$0.998$ for every method and fold, the same as on the random split ($0.986$--$0.991$), and the median uncertainty scores on the held-out cells are at the level of the random-split test cells (BNN $0.000$--$0.002$ against $0.000$; ENS $0.12$--$0.13$ against $0.13$--$0.16$; MCD $0.11$--$0.21$ against $0.14$--$0.22$). We therefore find no donor effect in the embedding that the classifiers could pick up, and the random split does not inflate the reported accuracies.}

\begin{table}[h]
\centering\small
\caption{\rev{Leave-donor-out evaluation on the single-cell data. Each row holds one sample (site and donor) out of training entirely; the models are trained on the other twelve samples and tested on all cells of the four selected types from the held-out sample. Accuracy on the held-out sample, with the accuracy of the same seed's random-split model on its own test set in parentheses.}}
\label{supp_tab:ldo}
\begin{tabular}{lrccc}
\toprule
held-out sample & test cells & MCD & BNN & ENS \\
\midrule
s1d1 & 3145 & 0.996 (0.987) & 0.989 (0.988) & 0.995 (0.987) \\
s2d1 & 2501 & 0.983 (0.991) & 0.977 (0.989) & 0.982 (0.989) \\
s4d1 & 3231 & 0.998 (0.991) & 0.996 (0.990) & 0.997 (0.989) \\
s3d10 & 2746 & 0.991 (0.991) & 0.989 (0.989) & 0.988 (0.991) \\
s3d3 & 1806 & 0.988 (0.989) & 0.981 (0.986) & 0.982 (0.989) \\
\bottomrule
\end{tabular}
\end{table}

\begin{table}[h]
\centering\small
\caption{\rev{Number of cells of the four selected cell types per sample (site $\times$ donor) in the single-cell dataset.}}
\label{supp_tab:donor_counts}
{\footnotesize\setlength{\tabcolsep}{3.5pt}
\begin{tabular}{lrrrrrrrrrrrrrr}
\toprule
cell type & s1d1 & s1d2 & s1d3 & s2d1 & s2d4 & s2d5 & s3d10 & s3d3 & s3d6 & s3d7 & s4d1 & s4d8 & s4d9 & All \\
\midrule
CD14$+$ Mono & 921 & 704 & 242 & 1308 & 1140 & 1716 & 1603 & 855 & 208 & 674 & 1014 & 458 & 0 & 10843 \\
CD4$+$ T activated & 836 & 709 & 392 & 446 & 385 & 493 & 207 & 228 & 276 & 102 & 619 & 684 & 149 & 5526 \\
Erythroblast & 797 & 136 & 374 & 478 & 1485 & 64 & 391 & 246 & 84 & 31 & 474 & 266 & 90 & 4916 \\
NK & 591 & 982 & 696 & 269 & 136 & 848 & 545 & 477 & 126 & 256 & 1124 & 421 & 458 & 6929 \\
\midrule
All & 3145 & 2531 & 1704 & 2501 & 3146 & 3121 & 2746 & 1806 & 694 & 1063 & 3231 & 1829 & 697 & 28214 \\
\bottomrule
\end{tabular}}
\end{table}

\begin{table}[h]
\centering\small
\caption{\rev{Accuracy of each UQ method per sample on the single-cell test set, mean over the five random seeds, for the balanced and the imbalanced training setting. Both settings are evaluated on the same balanced test set.}}
\label{supp_tab:donor_acc}
\begin{tabular}{lcccccc}
\toprule
 & \multicolumn{3}{c}{balanced} & \multicolumn{3}{c}{imbalanced} \\
sample & MCD & BNN & ENS & MCD & BNN & ENS \\
\midrule
s1d1 & 0.995 & 0.993 & 0.995 & 0.995 & 0.992 & 0.972 \\
s1d2 & 0.955 & 0.946 & 0.950 & 0.936 & 0.930 & 0.927 \\
s1d3 & 0.998 & 0.998 & 1.000 & 0.994 & 0.993 & 0.978 \\
s2d1 & 0.990 & 0.989 & 0.978 & 0.960 & 0.956 & 0.918 \\
s2d4 & 1.000 & 0.998 & 1.000 & 0.990 & 0.991 & 0.920 \\
s2d5 & 1.000 & 0.998 & 0.999 & 0.997 & 0.997 & 0.995 \\
s3d10 & 0.985 & 0.985 & 0.986 & 0.987 & 0.988 & 0.972 \\
s3d3 & 0.985 & 0.986 & 0.984 & 0.985 & 0.986 & 0.963 \\
s3d6 & 0.994 & 0.990 & 0.998 & 0.980 & 0.985 & 0.967 \\
s3d7 & 0.972 & 0.978 & 0.970 & 0.974 & 0.968 & 0.964 \\
s4d1 & 0.996 & 0.997 & 0.998 & 0.995 & 0.995 & 0.977 \\
s4d8 & 0.988 & 0.988 & 0.989 & 0.987 & 0.982 & 0.968 \\
s4d9 & 0.991 & 0.991 & 0.992 & 0.983 & 0.984 & 0.963 \\
\bottomrule
\end{tabular}
\end{table}

\subsection{\rev{Where the uncertain cells are}}\label{app_b:umap_unc}
\begin{figure}[h]
    \centering
    \includegraphics[width=\textwidth]{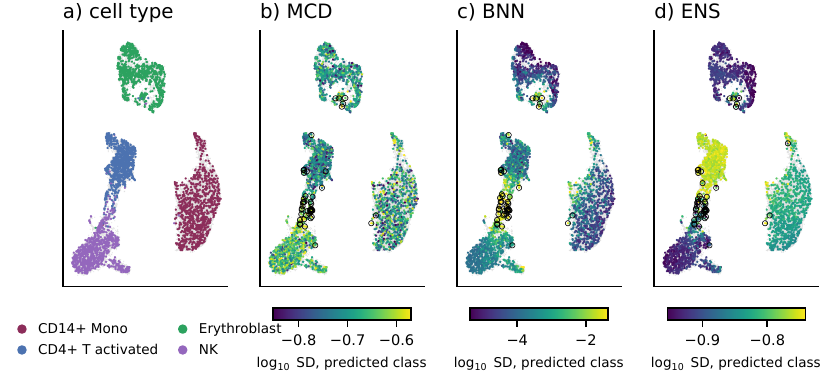}
    \caption{\rev{\textbf{Uncertainty of the single-cell test cells in the embedding.} UMAP of the Liam embedding of the $28{,}214$ selected cells (gray), with the $3{,}600$ test cells of seed $7$ colored by cell type (a) and by each method's $\log_{10}$ SD at the predicted class on the balanced setting (b--d). Misclassified cells have a black edge. Color ranges are the 2nd to 98th percentile of each method's scores.}}
    \label{supp_fig:umap_unc}
\end{figure}

\rev{Figure~\ref{supp_fig:umap_unc} places the test cells of one seed in the UMAP of the Liam embedding and colors them by each method's uncertainty score on the balanced setting. BNN's uncertainty concentrates where CD4+ T activated and NK cells meet, which is also where the misclassified cells are: over the five seeds, $11\%$ of the cells in BNN's most uncertain decile are misclassified, against $1.2\%$ of all test cells. For MCD the decile contains $6\%$ misclassified cells against $1\%$ overall. ENS's most uncertain decile is $93\%$ CD4+ T activated cells and contains fewer misclassified cells than the test set as a whole ($0.6\%$ against $1.1\%$): its score separates cell types rather than cells.}

\subsection{Detecting out-of-distribution sequences}\label{app_b:ood}
A common situation in genomics is applying a model trained on human data to a novel, out-of-distribution domain, such as a newly sequenced organism. \chg{Understanding the binding interactions of RNA-binding proteins in viruses such as SARS-CoV-2 can provide valuable insights into the infection mechanism. RBP binding datasets generated by experimental assays are widely available for the human transcriptome, yet similarly rich datasets are missing for novel viral genomes. At the same time, the differences in composition between the viral and the human genome limit the effectiveness of applying a model trained on one genome's data to the other: the viral sequence composition is biased towards AT relative to the coding and non-coding human transcriptome~\citep{horlacher2023}. More generally, novel or rapidly emerging organisms are precisely the case where labeled data are expensive or impossible to generate quickly, as during the SARS-CoV-2 pandemic, and where the underlying biology is not yet fully understood, so the extent to which a model trained on human data can be expected to transfer is not known in advance, even between datasets from the same experimental protocol and tissue. We therefore treat the viral data as an out-of-distribution case for models trained on human data and ask whether the uncertainty scores flag it as such. No labels exist for the viral windows, so the evaluation rests entirely on the uncertainty scores --- precisely the label-free setting this framework is designed for.}

We trained on human single-label RBP data and tested on \chg{eight human pathogenic coronavirus genomes, tiled into windows of the same length as the training sequences. The models are those of Section~\ref{app_a:data:real}, applied without retraining.}

All UQ methods show a shift towards higher uncertainty on the out-of-distribution data, with BNN \rev{moving furthest on the probability scale} (Figure~\ref{fig:shift_virus}), indicating that the uncertainty score can flag inputs that fall outside the training distribution even without labels. \chg{Quantifying the shift with the two-sided KS statistic between the human and viral score distributions, over $5$ random seeds and all eight genomes, gives a median of \rev{$0.63$ for BNN, $0.56$ for ENS and $0.69$ for MCD}. A more practical framing is to set a flagging threshold at the $95$th percentile of the human scores, as a practitioner without viral labels would have to: BNN then flags \rev{$35\%$} of viral windows as unusually uncertain\rev{, ENS $26\%$ and MCD $41\%$. The size of the response differs: the median score on viral windows is $0.17$ for BNN, $0.15$ for ENS and $0.12$ for MCD, so BNN's score moves furthest on the probability scale, while all three put the viral windows above most human sequences.}}

\paragraph{\rev{Out-of-distribution detection metrics.}}
\rev{To separate the size of the response from its usefulness, we treat the viral windows as the positive class and the human test sequences as the negative class and use the uncertainty score as the detection score. Table~\ref{supp_tab:ood} gives the area under the ROC curve and the true-positive rate at $5\%$ false-positive rate, pooled over all eight genomes and averaged over the five seeds, with the minimum and maximum across seeds. All three methods separate viral from human sequences (AUROC $0.87$ for BNN, $0.84$ for ENS and $0.89$ for MCD); at $5\%$ false positives BNN recovers $34\%$ of viral windows against $26\%$ for ENS and $40\%$ for MCD. The ranges over seeds of BNN and MCD overlap. BNN's response is the largest on the probability scale (median SD $0.17$ on viral windows against $0.15$ and $0.12$), as Figure~\ref{fig:shift_virus} shows; the ranking metrics here ask only whether the shifted inputs come out on top, which they do for every method. The score for the viral windows is the maximum SD over classes, since no prediction is stored for them; using it for the human side as well, or the SD at the predicted class as in Figure~\ref{fig:shift_virus}, changes the numbers by at most $0.01$.}

\begin{table}[h]
\centering\small
\caption{\rev{Out-of-distribution detection on the coronavirus genomes: AUROC, true-positive rate at $5\%$ false-positive rate, and the fraction of viral windows above the $95$th percentile of the human test scores. Mean over five seeds, range in parentheses; all eight genomes pooled.}}
\label{supp_tab:ood}
\begin{tabular}{lccc}
\toprule
 & AUROC & TPR at 5\% FPR & flagged at 95th pct. \\
\midrule
MCD & 0.89 (0.87--0.91) & 0.40 (0.28--0.46) & 0.41 (0.29--0.47) \\
BNN & 0.87 (0.85--0.90) & 0.34 (0.27--0.45) & 0.35 (0.28--0.46) \\
ENS & 0.84 (0.81--0.86) & 0.26 (0.19--0.29) & 0.26 (0.20--0.29) \\
\midrule
\multicolumn{4}{l}{\textit{AUROC per genome (mean over seeds)}} \\
 & MCD & BNN & ENS \\
229E & 0.91 & 0.88 & 0.86 \\
HKU1 & 0.85 & 0.84 & 0.79 \\
MERS & 0.91 & 0.89 & 0.86 \\
SARS-CoV-2 (NC\_045512) & 0.90 & 0.88 & 0.84 \\
NL63 & 0.88 & 0.86 & 0.83 \\
OC43 & 0.89 & 0.86 & 0.82 \\
SARS-CoV-1 & 0.91 & 0.89 & 0.86 \\
SARS-CoV-2 (wuhCor1) & 0.90 & 0.88 & 0.84 \\
\bottomrule
\end{tabular}
\end{table}

\subsection{\rev{Calibration}}\label{app_b:calibration}
\begin{figure}[h]
\centering
\includegraphics[width=\textwidth]{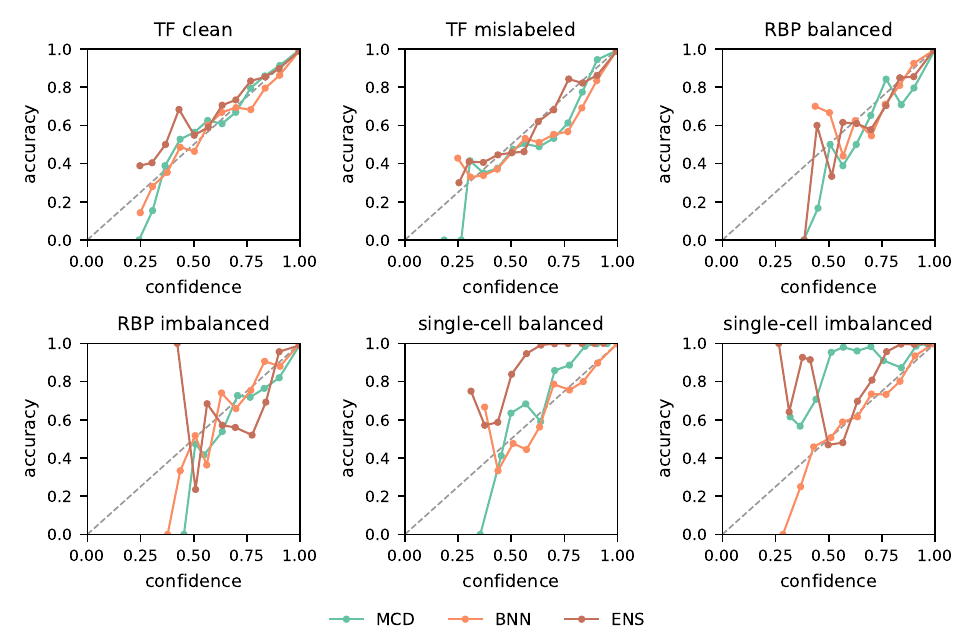}
\caption{\rev{\textbf{Reliability diagrams.} Accuracy against confidence of the predicted class, 15 bins, pooled over five seeds, for the six single-label settings. The dashed line is perfect calibration\rev{; green is MCD, orange BNN and brown ENS}. ECE values are given in Table~\ref{supp_tab:ece}.}}
\label{supp_fig:reliability}
\end{figure}

\rev{NLL and the Brier score are proper scoring rules and confound calibration with accuracy. Table~\ref{supp_tab:ece} reports the expected calibration error (ECE, 15 equal-width bins on the confidence of the predicted class) for the single-label settings, and Figure~\ref{supp_fig:reliability} the reliability diagrams. On the sequence data all three methods are well calibrated (ECE $0.01$--$0.05$). On the single-cell data the deep ensemble is strongly under-confident: its mean confidence is 0.75 at an accuracy of 0.99 (ECE 0.24), because its ten members disagree on cells they all classify correctly. \rev{MCD is under-confident there as well (ECE $0.09$ and $0.13$), for the same reason: its passes disagree on cells the mean prediction classifies correctly. BNN is the best calibrated method on the single-cell data (ECE $0.004$ and $0.011$).} This is the same effect that gives ENS the highest NLL and Brier score on that task in Figure~\ref{fig:pred_acc}, and it is the starting point from which ENS ``becomes more confident'' when trained on imbalanced data. }

\begin{table}[h]
\centering\small
\caption{\rev{Expected calibration error, mean $\pm$ sd over five seeds, with mean confidence / accuracy in parentheses. Single-label settings; 15 bins.}}
\label{supp_tab:ece}
\begin{tabular}{lccc}
\toprule
setting & MCD & BNN & ENS \\
\midrule
TF clean & 0.013 $\pm$ 0.003 (0.95 / 0.95) & 0.017 $\pm$ 0.003 (0.94 / 0.93) & 0.019 $\pm$ 0.005 (0.95 / 0.95) \\
TF mislabeled & 0.046 $\pm$ 0.009 (0.83 / 0.79) & 0.051 $\pm$ 0.013 (0.80 / 0.76) & 0.034 $\pm$ 0.004 (0.80 / 0.78) \\
RBP balanced & 0.015 $\pm$ 0.004 (0.98 / 0.97) & 0.013 $\pm$ 0.002 (0.97 / 0.97) & 0.016 $\pm$ 0.004 (0.96 / 0.96) \\
RBP imbalanced & 0.013 $\pm$ 0.005 (0.98 / 0.97) & 0.012 $\pm$ 0.003 (0.97 / 0.97) & 0.018 $\pm$ 0.006 (0.98 / 0.97) \\
single-cell balanced & 0.090 $\pm$ 0.024 (0.90 / 0.99) & 0.004 $\pm$ 0.001 (0.99 / 0.99) & 0.237 $\pm$ 0.005 (0.75 / 0.99) \\
single-cell imbalanced & 0.128 $\pm$ 0.025 (0.86 / 0.98) & 0.011 $\pm$ 0.005 (0.97 / 0.98) & 0.188 $\pm$ 0.006 (0.77 / 0.96) \\
\bottomrule
\end{tabular}
\end{table}

\subsection{\rev{Predictive entropy and mutual information}}\label{app_b:entropy_mi}
\rev{All analyses in the main text use the standard deviation of the predicted probability across passes as the uncertainty score. From the same passes one can also compute the predictive entropy $H[\bar p]$ (total uncertainty), the expected entropy $\mathbb{E}_t H[p_t]$ (aleatoric part) and their difference, the mutual information between prediction and parameters (epistemic part, i.e.\ the disagreement between passes). Table~\ref{supp_tab:entropy_mi} compares them with the SD score on the raw per-pass predictions. For BNN and ENS, mutual information and the SD rank the test samples nearly identically (Kendall $\tau$ between 0.86 and 0.98) and give the same reference-to-perturbed shift per class, so none of the conclusions depends on which of the three scores is used. \rev{For MCD the same holds ($\tau$ between SD and MI $0.85$ to $0.98$).}}

\begin{table}[h]
\centering\small
\caption{\rev{Top: rank agreement between the SD score and mutual information (MI) or predictive entropy (Kendall $\tau$, mean over seeds and settings), and the fraction of test samples with zero MI. Bottom: median shift ($\log_2$) at the predicted class from the reference to the perturbed setting, for the SD score and for MI; mean over five seeds (seeds with a positive shift). Here each setting's own model selects the samples it predicts as the class and the shift is the ratio of the two medians, so the RBP values differ from the paired comparison in Table~\ref{supp_tab:seed_shift}.}}
\label{supp_tab:entropy_mi}
\begin{tabular}{llccc}
\toprule
task & method & $\tau$(SD, MI) & $\tau$(SD, entropy) & MI $=0$ \\
\midrule
single-cell, class imbalance & MCD & 0.85 & 0.40 & 0\% \\
single-cell, class imbalance & BNN & 0.92 & 0.92 & 0\% \\
single-cell, class imbalance & ENS & 0.86 & 0.69 & 0\% \\
RBP, class imbalance & MCD & 0.98 & 0.95 & 3\% \\
RBP, class imbalance & BNN & 0.98 & 0.95 & 0\% \\
RBP, class imbalance & ENS & 0.93 & 0.94 & 7\% \\
TF, label noise & MCD & 0.94 & 0.84 & 0\% \\
TF, label noise & BNN & 0.93 & 0.85 & 1\% \\
TF, label noise & ENS & 0.94 & 0.86 & 3\% \\
\bottomrule
\end{tabular}

\vspace{0.8em}
\begin{tabular}{lcccccc}
\toprule
 & \multicolumn{3}{c}{SD score} & \multicolumn{3}{c}{mutual information} \\
class & MCD & BNN & ENS & MCD & BNN & ENS \\
\midrule
\multicolumn{7}{l}{\textit{single-cell, class imbalance}} \\
CD14$+$ Mono & $-0.8$ (0/5) & $-0.6$ (2/5) & $-3.8$ (0/5) & $-1.2$ (0/5) & $-0.2$ (2/5) & $-3.9$ (0/5) \\
CD4$+$ T activated & $-0.0$ (3/5) & $+2.4$ (4/5) & $-0.8$ (0/5) & $-0.0$ (3/5) & $+2.5$ (4/5) & $-0.8$ (0/5) \\
Erythroblast & $-0.0$ (2/5) & $+5.6$ (5/5) & $+1.2$ (5/5) & $-0.3$ (3/5) & $+5.1$ (5/5) & $+1.8$ (5/5) \\
NK & $-0.5$ (2/5) & $+1.1$ (4/5) & $-1.2$ (0/5) & $-0.7$ (1/5) & $+1.2$ (4/5) & $-1.3$ (0/5) \\
\multicolumn{7}{l}{\textit{RBP, class imbalance}} \\
MBNL1 & $-2.7$ (0/5) & $-1.5$ (1/5) & $-10.5$ (0/5) & $-4.2$ (0/5) & $-1.6$ (1/5) & $-13.3$ (0/5) \\
PUM2 & $+1.1$ (5/5) & $+0.9$ (4/5) & $-6.4$ (0/5) & $+1.1$ (5/5) & $+0.9$ (3/5) & $-5.7$ (0/5) \\
QKI & $+6.7$ (5/5) & $+4.6$ (5/5) & $-2.0$ (0/5) & $+8.5$ (5/5) & $+4.5$ (5/5) & $-2.0$ (0/5) \\
\multicolumn{7}{l}{\textit{TF, label noise}} \\
CTCF & $+1.0$ (4/5) & $+1.6$ (4/5) & $+0.2$ (4/5) & $+1.1$ (5/5) & $+2.0$ (4/5) & $+0.6$ (5/5) \\
HNF4A & $+0.0$ (3/5) & $+2.2$ (5/5) & $+1.0$ (5/5) & $+0.1$ (3/5) & $+2.1$ (5/5) & $+0.9$ (4/5) \\
JUN & $+5.7$ (5/5) & $+6.1$ (5/5) & $+7.1$ (5/5) & $+5.3$ (5/5) & $+5.6$ (5/5) & $+6.2$ (5/5) \\
MEF2A & $+4.7$ (5/5) & $+4.5$ (5/5) & $+5.8$ (5/5) & $+4.3$ (5/5) & $+4.0$ (5/5) & $+4.9$ (5/5) \\
MYC & $+0.3$ (3/5) & $+1.9$ (4/5) & $+0.5$ (4/5) & $+0.4$ (3/5) & $+2.1$ (4/5) & $+0.4$ (5/5) \\
TAL1 & $+0.8$ (4/5) & $+2.2$ (5/5) & $-0.4$ (1/5) & $+0.8$ (4/5) & $+2.1$ (4/5) & $-0.5$ (1/5) \\
\bottomrule
\end{tabular}
\end{table}
\subsection{MNIST as a benchmark domain}\label{app_b:mnist}
\begin{figure}[h]
    \centering
    \includegraphics[width=0.62\textwidth]{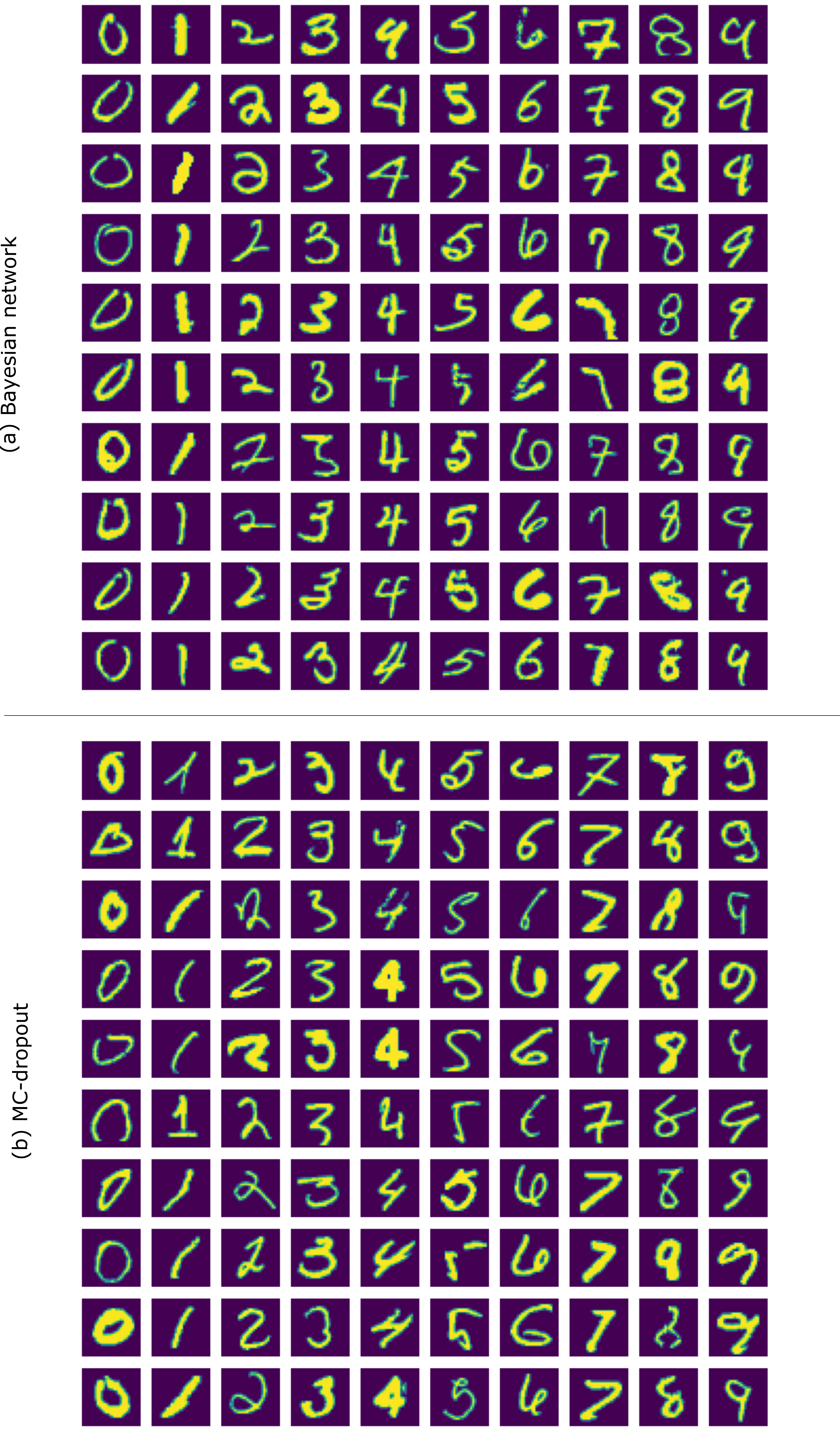}
    \caption{\chg{\textbf{MNIST samples identified as most uncertain.} Columns are digit classes $0$--$9$. Top block: samples among the most uncertain for the BNN; bottom block: samples among the most uncertain for MC-dropout.}}
    \label{supp_fig:mnist_digits}
\end{figure}

\rev{As a qualitative cross-check outside genomics, we repeated the agreement analysis for BNN and MC-dropout on MNIST~\citep{mnist_lecun}, on the original balanced dataset ($70{,}000$ images) and on an imbalanced version of about $42{,}000$ images obtained by sub-sampling the ten digit classes with fractions between $0.2$ (class $7$) and $0.9$ (class $8$). The picture matches the genomics results: the overlap between the two methods' most uncertain samples is well above chance but far from complete, it varies between digit classes, and the spread widens under imbalance, with some digits agreeing on more than $70$ of the top $10\%$ and others on fewer than $20$. Figure~\ref{supp_fig:mnist_digits} shows the samples themselves; by inspection, the samples MC-dropout flags look more like what a human would call ambiguous. Disagreement between the two methods therefore does not isolate a subset of ``genuinely'' hard images; it reflects their different posterior approximations, as argued for the genomics tasks in Section~\ref{sec_4_2:seq_agree}.}

\subsection{\rev{Label-noise rate}}\label{app_b:noise_rate}
\rev{The label-noise experiment of the main text swaps the labels of half of the JUN and MEF2A sequences. Table~\ref{supp_tab:noise_rate} repeats it with $15\%$ and $30\%$ of the two classes swapped, with the same protocol, test set and seeds. All three methods respond at every rate: the two swapped classes shift by $+4$ to $+6$ $\log_2$ units already at $15\%$, in every seed, and the shift grows little with the rate. The untouched classes shift less at every rate; for MCD their shift stays at about $+2$, for BNN and ENS it grows with the rate ($+1.8$ to $+3.2$ and $+2.3$ to $+3.6$), so the gap between the swapped and the untouched classes is largest for MCD, followed by BNN and ENS. At $15\%$ and $30\%$ the clean-test accuracy stays at $0.87$--$0.95$, because the majority label of each class is still the correct one; at $50\%$ the two classes cannot be told apart and the mean accuracy over seeds falls to $0.76$--$0.79$ (per-seed values in Section~\ref{sec_4_1:perf}). One interpretation is that the BNN's weight posterior widens when the data become harder to fit, and the extra uncertainty reaches every class through the shared convolutional filters, whereas MC-dropout perturbs each prediction separately; our experiments cannot confirm this.}

\begin{table}[h]
\centering\small
\caption{\rev{Label-noise rate on the simulated TF data. For $15\%$, $30\%$ and $50\%$ of the JUN and MEF2A labels swapped in the training and validation sets: accuracy on the clean test set, and the shift of the median $\log_2$(SD) from the clean setting, over all test samples, averaged over the two swapped classes and over the four untouched classes (mean over five seeds; seeds with a positive shift).}}
\label{supp_tab:noise_rate}
\begin{tabular}{lcccc}
\toprule
swapped & method & accuracy & shift, swapped classes & shift, untouched classes \\
\midrule
15\% & MCD & $0.95$ & $+5.7$ (5/5) & $+1.9$ (5/5) \\
 & BNN & $0.91$ & $+5.5$ (5/5) & $+1.8$ (4/5) \\
 & ENS & $0.95$ & $+4.3$ (5/5) & $+2.3$ (5/5) \\
\midrule
30\% & MCD & $0.93$ & $+5.5$ (5/5) & $+1.7$ (5/5) \\
 & BNN & $0.87$ & $+6.3$ (5/5) & $+3.1$ (5/5) \\
 & ENS & $0.94$ & $+4.8$ (5/5) & $+3.2$ (5/5) \\
\midrule
50\% & MCD & $0.79$ & $+6.2$ (5/5) & $+1.7$ (5/5) \\
 & BNN & $0.76$ & $+6.5$ (5/5) & $+3.2$ (5/5) \\
 & ENS & $0.78$ & $+4.8$ (5/5) & $+3.6$ (5/5) \\
\bottomrule
\end{tabular}
\end{table}

\end{document}